\documentclass[12pt, letterpaper]{article}
\usepackage[margin=1in]{geometry}
\usepackage{authblk}
\usepackage{array} % for defining a new column type
\usepackage{enumitem} % for customizing the list
\usepackage{graphicx}
\usepackage{subcaption}
\usepackage{hyperref}
\usepackage{multirow}
\usepackage{tabularx}
\usepackage{url}
\usepackage{hyperref}

\usepackage{comment}

\newcolumntype{M}[1]{>{\centering\arraybackslash}m{#1}}

\title{Shapley Values-Powered Framework for Fair Reward Split in Content Produced by GenAI}
\author[1]{Alex Glinsky\thanks{alex.glinsky@dataxide.ai}}
\author[1]{Alexey Sokolsky\thanks{al@dataxide.ai}}

\affil[1]{Dataxide.ai}

\date{} 

\begin{document}

\maketitle

\begin{abstract}
It is evident that, currently, generative models are surpassed in quality by human professionals. However, with the advancements in Artificial Intelligence, this gap will narrow, leading to scenarios where individuals who have dedicated years of their lives to mastering a skill become obsolete due to their high costs, which are inherently linked to the time they require to complete a task—a task that AI could accomplish in minutes or seconds. To avoid future social upheavals, we must, even now, contemplate how to fairly assess the contributions of such individuals in training generative models and how to compensate them for the reduction or complete loss of their incomes. In this work, we propose a method to structure collaboration between model developers and data providers. To achieve this, we employ Shapley Values to quantify the contribution of artist(s) in an image generated by the Stable Diffusion-v1.5 model and to equitably allocate the reward among them.
\end{abstract}

\section{Introduction}
Today, training deep models, particularly those based on the Transformers \cite{vaswani2023attention} architecture, increasingly demands larger volumes of data. Generally, utilizing vast quantities of data sourced from the internet would not pose a problem. For instance, training the CLIP network required millions of image-caption pairs. However, the critical issue is the function the model performs. So far, CLIP simply correlated images with the most appropriate text, streamlined the process of graphical search, and facilitated enhancements in one-shot classification pipelines. It did not pose a significant threat to the job market and appeared as an iterative improvement over existing approaches. However, everything changed when StabilityAI and their generative model, Stable Diffusion, made a quantum leap in quality, reaching a level where the works created with its participation could be sold (albeit with certain reservations) at the level of real digiral artists. Ironically, the same CLIP, responsible for precise style matching in Stable Diffusion, existed several years before the advent of SD but didn't attract any attention outside the data science community.

The same holds for text models. As long as transformers were used for text classification, automatic translation, and text gap filling, no concerns were raised. Even GPT-2 \cite{radford2019language}, capable of generating unique text on any given topic, did not arouse apprehension, as it was evidently below the average human level. However, with the growth and development of technology, it has become clear that over time, generative AI will continue to seize a larger market share from humans as it provides comparable quality in significantly less time, thereby exponentially accelerating various processes.

That is, the combination of two factors— the ability to generate new specified content and near-human level competencies— is what allows AI to compete with humans in fields that were deemed impervious to AI due to its lack of creativity. However, it’s crucial to understand that behind the knowledge and skills possessed by generative AI are the experience and efforts of real people who have dedicated themselves to their craft. AI has learned to absorb and transform their experience but, concurrently, has the potential to render these very people redundant in the job market.

This leads us to the question of how to support individuals who find themselves at risk, not only because it is humane but also because if AI can replace specialist humans, their number and quality will diminish over time. We might then find ourselves in a situation where the entry threshold into certain domains where AI excels most conspicuously (e.g., visual arts, poetry, prose, music) might become so elevated that new specialists simply cease to emerge. This will not only have a detrimental effect on the job market but could also lead to the degradation of AI itself, as the source of new, real data for training and further development of models will be exhausted.

To avert this potentially adverse scenario, we need to redistribute the revenues acquired by AI developers in favor of data providers. However, this must be done equitably and justly to ensure that every participant retains incentives for development and refinement of their tool. It is evident that the advancement of AI is only feasible in synergy with real people (data providers), and we must balance the allocation of rewards in such a way that neither side outweighs and consequently stifles the development of the other, leading to the cessation of advancement on both ends in the long run. In other words, a third entity is required to mediate between the model provider and the data provider for their equitable development.

The second point to be highlighted is the reward process itself. Assuming that we purchase data and pay for it only once is not a viable solution. At the model training stage, we do not know which data will be in demand in the future. It might turn out that 1\% of the data accounts for 100\% of the generated content. Moreover, we are unaware of which data will yield the most benefit during model training. It has long been known that the importance lies not in the quantity of data, but in its quality and diversity. Therefore, we propose an approach where income is distributed each time new content is generated, and is directly dependent on the content itself. For instance, if images generated by StableDiffusion are sold on a photo stock, we must determine for each image sold, say for \$1.00, what portion goes to the model developers, what portion to the prompt engineer who generated the image, what portion to the photo stock that managed to bring together buyer and seller, and most importantly for our work, what portion goes to the artist whose works had the most substantial influence on the final generation result.

It is well known that Stable Diffusion and similar generative models can replicate the styles of existing artists in a fairly analogous manner, provided that the model has been adequately trained on their works. However, complications arise when attributing the style of a generated image unequivocally to a specific artist and, especially, when attempting to accurately assess the degree of an artist's style contribution to the final result. To address this, we introduce our meticulously developed framework. This framework aims to elucidate the following questions:

\begin{itemize}
    \item Is the model familiar with a specific artist and their style, or in other words, were this artist’s works included in the training dataset?
    \item To what extent can it precisely reproduce the artist's style?
    \item How to assess the artist's contribution to the final generation and, consequently, their share of the reward?
    \item What measures should be taken in cases of mixed style, stemming from multiple artists?
\end{itemize}

To answer the first two questions, we employ heuristics developed by our team. They rely on available open data, collected by enthusiasts, as well as on fundamental properties of deep neural networks, such as embeddings and gradients.

To respond to the remaining queries and to quantify the contributions of the artist(s) and the model to the final output, we employ Shapley Values \cite{shapley1953value}, a well-established and robust method renowned for evaluating the individual contributions of participants in cooperative game scenarios. This method offers precision and reliability in attributing distinct contributions, ensuring a fair and balanced assessment of input from both artists and models. Within the realm of data contributors, employing Shapley Values is extraordinarily pertinent; it not only ascertains the relevance of each contributor’s input to model training but also acts as an equitable mechanism for apportioning remuneration or indemnification amongst all stakeholders. This equitable apportionment can potentially invigorate involvement and cooperation in the development and refinement of machine learning models.

Shapley Values, despite their precision and objectivity, encounter severe scalability issues when dealing with large volumes of data. The computational complexity of the Shapley Values method grows factorially with the increase in the number of participants, rendering it practically unfeasible for analyzing the contributions of a multitude of data providers. Even when employing approximations and optimizations, such as, for example, Truncated Monte Carlo Shapley \cite{ghorbani2019data}, calculating Shapley Values can demand substantial computational resources and time. This limits their applicability in big data scenarios where swift and scalable contribution assessment methods are critically important. Furthermore, we lack access to the data and the training processes of large generative models.

Therefore, we are rethinking the process of computing Shapley Values, drawing inspiration from how it is implemented in the \textbf{SHAP} \cite{lundberg2017unified} package for transformer-like models, such as \textbf{BERT} \cite{kokalj-etal-2021-bert} . We attempt to calculate the contribution of training data, having access only to a black-box model and the prompt - a string inputted by the user to obtain a generated image (e.g., \textit{an image of a space shuttle, by Van Gogh}).

For this purpose, we propose the following model: each generated image will be considered as a combination of content and style. Content pertains to what is depicted in the image (e.g., a shuttle, a man, an ancient castle, a dragon, etc.), while style refers to how it is depicted (e.g., in the style of Van Gogh, in anime style, etc.). Now, we would like to delineate what we reward the artist for and what we reward the generative model for (refer to Table~\ref{tab:table_1}).

\begin{table}[h!]
    \centering
    \begin{tabular}{|M{0.5\textwidth}|M{0.5\textwidth}|}
        \hline
        \textbf{Artist(s)} & \textbf{Generative Model} \\
        \hline
        \begin{enumerate}[left=0pt]
            \item The artist earns more compensation the closer the generation aligns with their style.
            \item Artist receives extra compensation if the generated content is similar to content they have previously created.
        \end{enumerate} &
        \begin{enumerate}[left=0pt]
            \item Transferring the style to new content not inherent to the given artist.
            \item Combining the styles of multiple artists.
        \end{enumerate} \\
        \hline
    \end{tabular}
    \caption{Reasons to reward participant}
    \label{tab:table_1}
\end{table}

\begin{figure}[h!]
    \centering
    
    \begin{subfigure}{0.32\textwidth}
        % \centering
        \includegraphics[width=\linewidth]{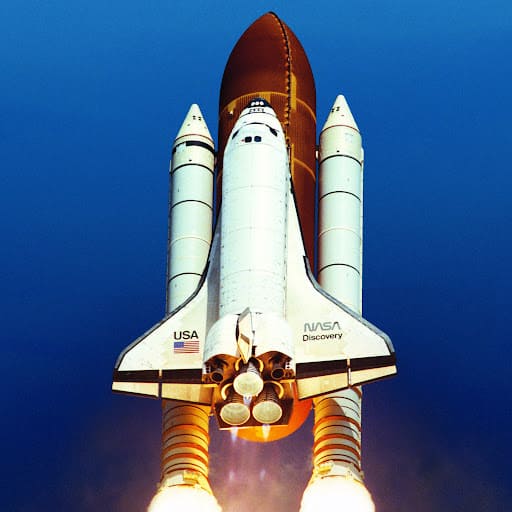}
        \caption{Content image (0.8\$)}
    \end{subfigure}
    \hfill
    \begin{subfigure}{0.32\textwidth}
        % \centering
        \includegraphics[width=\linewidth]{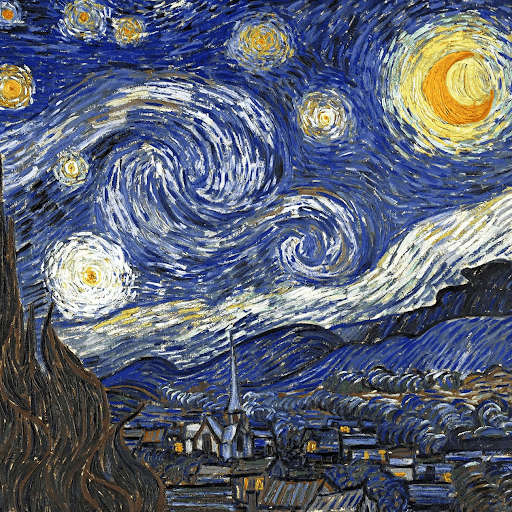}
        \caption{Style image (0.2\$)}
    \end{subfigure}
    \hfill
    \begin{subfigure}{0.32\textwidth}
        % \centering
        \includegraphics[width=\linewidth]{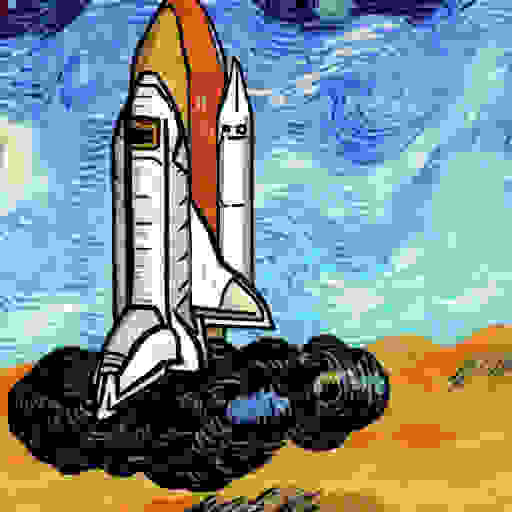}
        \caption{Generated image (1.0\$)}
    \end{subfigure}
    
    \centering
    \caption{An example of disereable reward split for an image generated by the prompt: \textit{an image of space shuttle, by Van Gogh}}
    \label{fig:figure_1}
\end{figure}

\begin{figure}[h!]
    \centering
    
    \begin{subfigure}{0.32\textwidth}
        % \centering
        \includegraphics[width=\linewidth]{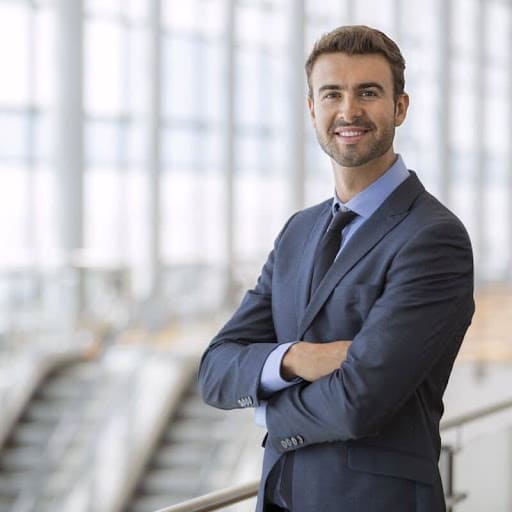}
        \caption{Content image (0.0\$)}
    \end{subfigure}
    \hfill
    \begin{subfigure}{0.32\textwidth}
        % \centering
        \includegraphics[width=\linewidth]{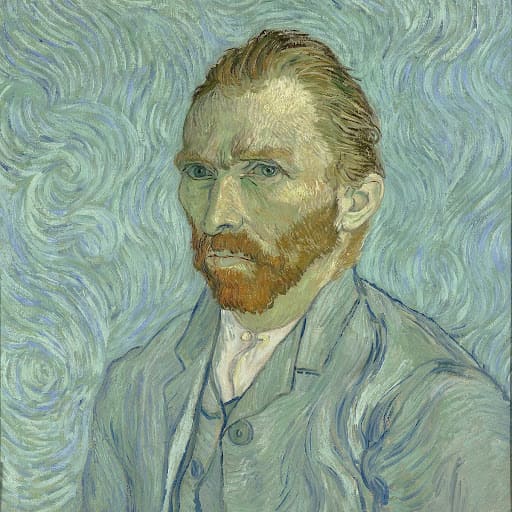}
        \caption{Style image (1.0\$)}
    \end{subfigure}
    \hfill
    \begin{subfigure}{0.32\textwidth}
        % \centering
        \includegraphics[width=\linewidth]{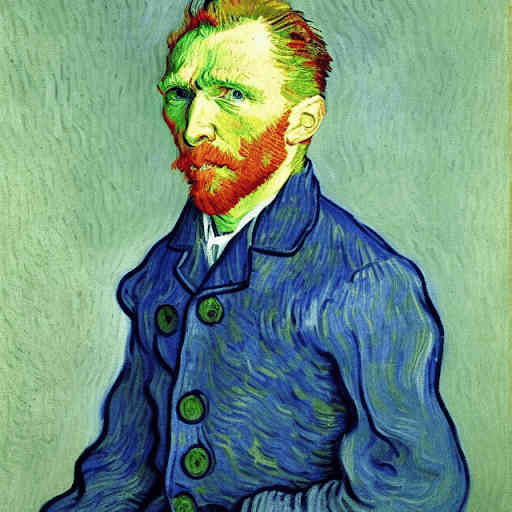}
        \caption{Generated image (1.0\$)}
    \end{subfigure}
    
    \centering
    \caption{An example of disereable reward split for an image generated by the prompt: \textit{an image of a man, by Van Gogh}}
    \label{fig:figure_2}
\end{figure}

\begin{figure}[h!]
    \centering
    
    \begin{subfigure}{0.24\textwidth}
        % \centering
        \includegraphics[width=\linewidth]{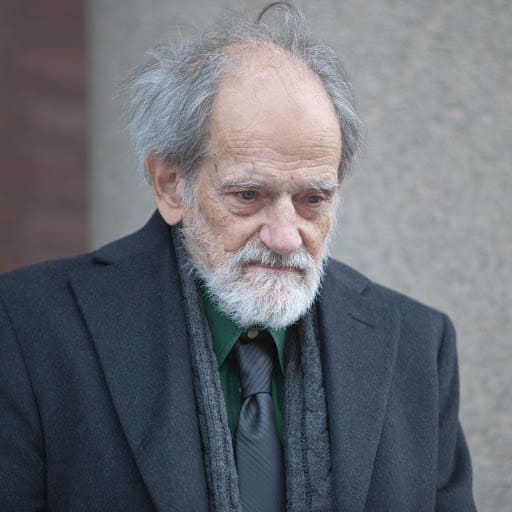}
        \caption{Content}
    \end{subfigure}
    \hfill
    \begin{subfigure}{0.24\textwidth}
        % \centering
        \includegraphics[width=\linewidth]{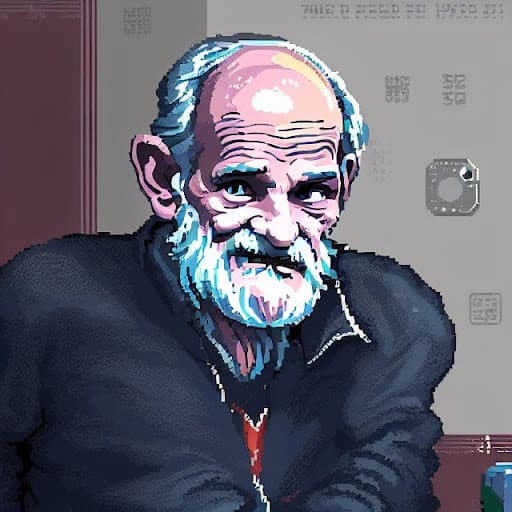}
        \caption{Style A}
    \end{subfigure}
    \hfill
    \begin{subfigure}{0.24\textwidth}
        % \centering
        \includegraphics[width=\linewidth]{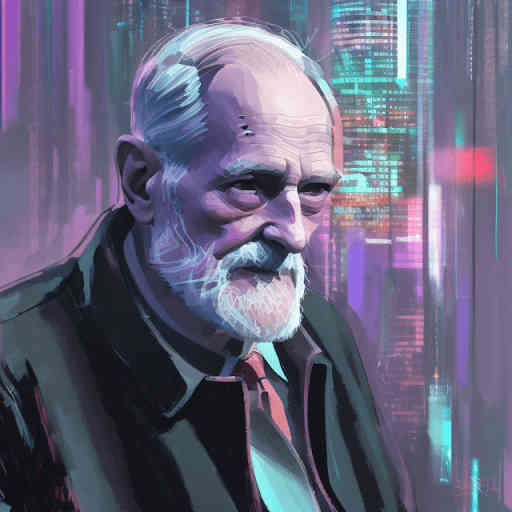}
        \caption{Style B}
    \end{subfigure}
    \hfill
    \begin{subfigure}{0.24\textwidth}
        % \centering
        \includegraphics[width=\linewidth]{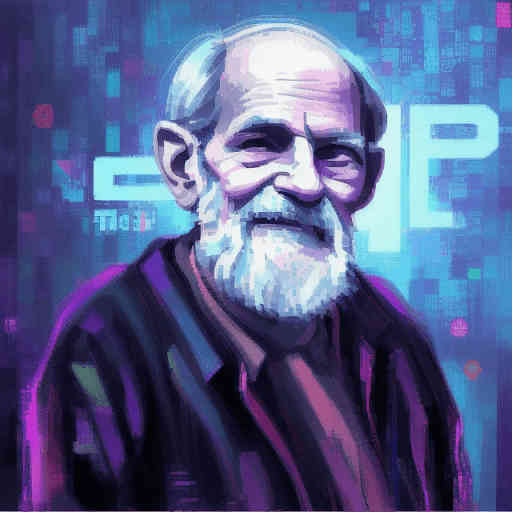}
        \caption{Generated image}
    \end{subfigure}
    
    \centering
    \caption{An image generated by the prompt: \textit{an image of a Lloyd Shapley}}
    \label{fig:figure_3}
\end{figure}

We will further elucidate this concept with several illustrative examples. Initially, in Figure~\ref{fig:figure_1}, we observe an instance where the 'content' receives a substantial portion of the reward, while a smaller proportion is allocated to 'style'. It’s evident that among Van Gogh's works, there are no depictions of spacecraft, hence, a significant portion of the reward is attributed to Stable Diffusion for its innovative transposition of style onto a novel subject. In Figure~\ref{fig:figure_2}, we present a more complex scenario where the generated image is nearly identical to the images contained in the training dataset. Such instances are not uncommon, as indicated in the work of \cite{carlini2023extracting}, and in such cases, it would be just to allocate the entire reward to the author of the original work. Finally, we illustrate one of the most intricate scenarios, where multiple styles are blended, particularly on content that is not inherent to any of the artists whose styles were utilized in generating the image, as exemplified in Figure~\ref{fig:figure_3}.

In summary, our study is framed around a more formal objective: we aim to devise a mechanism to equitably allocate the prospective profits derived from the use of generative models between AI developers and data providers, leveraging the principles of Shapley Values. The methodological approach involves utilizing prompts, generated images, and the black-box configurations of models such as StableDiffusion and CLIP to compute these values.

\section{Related Work}

\subsection{Image Generation Models}

The CLIP model (Contrastive Language–Image Pre-training) by OpenAI \cite{radford2021learning} was trained on a vast number of pairs of images and text descriptions. The training data for CLIP consisted of billions of images from the internet along with their corresponding text descriptions. The primary goal was for the model to learn to associate textual phrases with images where those phrases are best depicted. This enabled the CLIP model to perform various tasks, such as searching for images using text queries, classifying images, and more, without requiring additional task-specific training data.

DALL-E, a subsequent development by OpenAI, builds directly upon the foundational work established by the CLIP model. While CLIP was groundbreaking in its ability to create associations between textual and visual data, DALL-E \cite{ramesh2021zeroshot} extended this capability from mere recognition and categorization to actual generation of images. Utilizing a modified version of the GPT-3 \cite{brown2020language} architecture, DALL-E integrates the conceptual framework of CLIP to interpret and visualize textual descriptions as images. This progression from understanding to creation marks a significant advancement in the field of AI-driven image synthesis. The synthesis ability of DALL-E is in many ways a direct expansion of the associative learning demonstrated by CLIP, showcasing a novel application of multimodal AI models in bridging the realms of natural language processing and computer vision.

DALL-E utilizes VQ-VAE-2 \cite{razavi2019generating} to generate high-resolution images from latent representations, which are provided by a variant of the GPT-3 model. This integration allows DALL-E to efficiently transform detailed textual descriptions into corresponding visual outputs, ensuring both high fidelity and coherence in the generated images.

The VQ-VAE-2 (Vector Quantized Variational AutoEncoder 2) is a powerful generative model that stands out for its ability to create high-fidelity images. It works by first encoding images into a compressed, lower-dimensional latent space using a deep neural network. This latent space is quantized, meaning it represents the image data as a series of discrete tokens, much like words in a sentence. This quantization is key to VQ-VAE-2's effectiveness, as it allows the model to handle complex image data more efficiently.

A significant aspect of VQ-VAE-2 is its hierarchical structure, which encodes and decodes images at multiple resolutions. This multi-scale approach enables the model to capture both high-level abstract features and fine-grained details of the images. 

CLIP-Guided Diffusion \cite{crowson2021clip}, introduced in mid-2021, marked a significant advancement following CLIP's development. It employs diffusion models \cite{nichol2021improved} for synthesizing high-quality images up to 256x256 resolution (and even further, up to 512x512 \cite{crowson2021clip512}), using a U-Net architecture with ResNet blocks. Crucially, the CLIP model guides this process, providing semantic understanding to ensure the generated images align closely with textual descriptions. This method not only yields detailed visuals but also maintains semantic coherence, effectively merging natural language with visual representation.

GLIDE (Guided Language to Image Diffusion for Generation and Editing) \cite{nichol2022glide} represents an iterative advancement in the realm of AI-driven image generation, building upon the foundations laid by DALL-E and CLIP-Guided Diffusion. Developed by OpenAI, GLIDE integrates the textual understanding capabilities of models like CLIP with the generative power of diffusion-based methods, similar to those used in CLIP-Guided Diffusion, to produce highly detailed and contextually accurate images from text descriptions

DiscoDiffusion \cite{alembics2022disco} has emerged as a prominent tool in the realm of AI-driven image generation, particularly gaining popularity among digital artists for its creative potential. While it is capable of generating high-quality images, what sets DiscoDiffusion apart is its appeal to artistic creativity. It leverages diffusion models for image synthesis, guided by the CLIP model to interpret textual descriptions. However, the defining feature of DiscoDiffusion is its flexibility and the extensive customization options it offers, allowing artists to experiment with a wide range of styles, textures, and compositions. This flexibility has made DiscoDiffusion a favored tool for those looking to explore new dimensions of digital art, combining the precision of AI with the imaginative flair of human creativity.

This model marks a refinement in the approach to text-to-image synthesis. While retaining the core concept of generating images from textual input, as seen in DALL-E, GLIDE enhances the quality and relevance of the output by leveraging the iterative refinement process intrinsic to diffusion models.

DALL-E 2 \cite{ramesh2022hierarchical}, developed by OpenAI, marks a substantial technical evolution from its predecessor, DALL-E. One of the most significant technical changes in DALL-E 2 is the adoption of a new approach to image synthesis, shifting from the transformer-based methodology used in the original DALL-E to a diffusion-based generative model. This shift enables DALL-E 2 to produce images with remarkably higher resolution and fidelity. The diffusion model approach, which involves iteratively refining an image from a pattern of noise, allows for a more nuanced and detailed rendering of visual elements, leading to images that are not only clearer but also more complex and lifelike.

Another key technical enhancement in DALL-E 2 is its improved handling of the relationships between textual descriptions and visual elements. While the original DALL-E demonstrated the ability to generate images from text, DALL-E 2 refines this process, offering a more sophisticated interpretation of textual nuances. This is achieved through advancements in the underlying language models and more effective training methodologies, which enable a deeper understanding of the text and its intended visual representation.

Furthermore, DALL-E 2 incorporates a more refined training dataset, ensuring a broader and more diverse range of styles and subjects in the generated images. This improved dataset, combined with the advanced capabilities of the diffusion model, positions DALL-E 2 as a more versatile and powerful tool for image generation, capable of producing a wide array of images that better align with the variety and complexity of human imagination and linguistic expression.

Imagen \cite{saharia2022photorealistic}, Google's advanced text-to-image model, represents a leap forward in AI image generation, boasting several key technical advancements over its predecessors like DALL-E and GLIDE. At its core, Imagen employs a cascade of diffusion models, each designed to enhance the resolution of generated images. This multi-stage approach begins with creating a basic image, which is then progressively refined through subsequent models, each adding finer details. This method allows Imagen to achieve a level of detail and clarity that is a significant improvement over the single-stage processes used in earlier models.

In terms of processing textual inputs, Imagen integrates Google's T5 (Text-To-Text Transfer Transformer) model \cite{raffel2023exploring}, renowned for its versatility in language tasks. The T5 model in Imagen is fine-tuned to excel in interpreting and mapping complex, nuanced text descriptions to visual representations. This fine-tuning enables a more accurate translation of written descriptions into images, particularly effective with abstract or creatively challenging texts.

It's worth mentioning that working in pixel space allows models like Imagen and GLIDE to directly manipulate and refine the image at the pixel level, often resulting in higher fidelity and more detailed images. In contrast, models using latent space representations, like DALL-E (both versions), Stable Diffusion, and CLIP-Guided approaches, initially work with a more abstract, compressed representation of the image, which is then transformed into the final image. This approach can offer computational efficiencies and sometimes more control over the generative process.

This is possible thanks to the integration of latent space representation with diffusion processes in latent diffusion models \cite{rombach2021highresolution}. Rather than operating directly in pixel space as Imagen does, these models first encode images into a compressed, lower-dimensional latent space. The diffusion process is then applied within this latent space, iteratively refining the encoded representations into detailed images. Models like Stable Diffusion exemplify this approach, capitalizing on the efficiency of working in latent space while still achieving high-quality outputs.

\subsection{Training Data}

In this study, our interest extends beyond just the implementation details of generative models; we also focus on the datasets on which these models were trained. Given that CLIP is an integral component of many such models, it is logical to start by examining the dataset on which it was trained. However, this is a proprietary dataset, with only certain details disclosed:
"CLIP was trained on a dataset comprising 400 million (400M) image-text pairs. This expansive dataset is pivotal for the model's ability to learn a broad spectrum of visual concepts and their correlations with textual descriptions." 

But there exists an alternative to OpenAI's CLIP model, which was trained on private data. OpenCLIP \cite{cherti2022reproducible} is an open-source endeavor aimed at recreating OpenAI's CLIP model using publicly available data. This project stands out as it is driven by the community, involving contributions from a diverse group of researchers and AI enthusiasts. The goal is to achieve similar performance to the original CLIP but through the use of open datasets, enhancing transparency and accessibility in AI research. OpenCLIP not only makes this technology more accessible for experimentation and further development but also places a strong emphasis on ethical AI development. By focusing on the use of ethically sourced data and being mindful of biases, OpenCLIP contributes to the broader movement towards responsible AI innovation.

One of the main datasets used for training OpenCLIP is the LAION dataset. The LAION dataset is a large-scale, openly accessible collection of image-text pairs, making it suitable for training models that learn associations between visual content and textual descriptions, similar to CLIP. The LAION dataset is significant for its size and diversity, containing millions of image-text pairs sourced from the web.

Various versions and architectures of the CLIP model exist. For instance, Stable Diffusion v1.5 employed the ViT-L/14 variant. OpenCLIP offers several alternative variations of this model, including ViT-L/14, ViT-H/14, and ViT-G/14, all of which demonstrate similar metric results and quality. For training these models, open datasets such as LAION-2B \cite{schuhmann2022laion5b} and DataComp-1B \cite{gadre2023datacomp} were utilized.

LAION-5B comprises an impressive 5.85 billion image-text pairs, with 2.32 billion pairs in English, all filtered using CLIP. The creation of LAION-5B is a response to the need for large, publicly available datasets for training sophisticated language-vision models like CLIP, GLIDE, and Stable Diffusion. LAION-5B marks a significant step in democratizing AI research, providing the broader research community access to a dataset of this magnitude for the first time. The dataset not only enables successful replication and fine-tuning of foundational models but also opens up opportunities for new experiments in the field. The paper also highlights additional resources provided alongside LAION-5B, including tools for dataset exploration, subset generation, and detection of watermarks, NSFW, and toxic content, further enhancing its utility for AI research.

DATACOMP introduces a new candidate pool of 12.8 billion image-text pairs sourced from Common Crawl, which is broader than the collection in LAION-5B. This extensive pool offers a more diverse base for dataset creation. Unlike LAION-5B, which provided a pre-curated dataset, DATACOMP is designed as a testbed for conducting dataset experiments. The DATACOMP framework led to the creation of DATACOMP-1B, which, through optimized dataset curation and filtering, enabled the training of a CLIP ViT-L/14 model to achieve higher zero-shot accuracy on ImageNet compared to the original CLIP model.

In the Stable Diffusion 2.0 version, researchers at StabilityAI attempted to enhance the quality of generation over Stable Diffusion 1.5 by replacing OpenAI's CLIP (ViT-L/14) with OpenCLIP (ViT-H/14) trained on LAION-2B (a cleared subset of LAION-5B). OpenCLIP demonstrated superior metric performance in zero-shot validation and was considered more ethically suitable. However, this change led to a decline in generation quality and prompt understanding. Therefore, in our work, we adhere to the Stable Diffusion version 1.5 and conduct experiments on it, as it is more popular within the community and offers greater artistic variations and capabilities.

Large companies are highly protective of their data, leading to the majority of models discussed in Section 2.1 being trained on private datasets. Moreover, ordinary users do not have access to the Imagen model, and Midjourney has not provided any technical description of its model, making it effectively a black box. Therefore, it can be inferred that most publicly available online data are included in the private training datasets of generative models. LAION, due to its openness, is compelled to adopt a more stringent approach to data filtering.

Anticipating the unavailability of information regarding the data contained in the training dataset, we propose a method to assess the likelihood of specific images being present within a generative model. We also evaluate the model's degree of understanding of these data and its ability to reproduce them.

\subsection{Influence evaluation methods}

Numerous methodologies employing Shapley value computations mandate direct access to the dataset and the training process. Consequently, a predominant category of these methodologies strives to expedite the conventional Shapley value computation through approximation techniques or by segmenting the dataset into more substantial, interconnected clusters. Nevertheless, despite these modifications aimed at computational efficiency, these methods still require iterative retraining of the model, although the frequency and extent of this retraining are substantially reduced compared to the classical approach.

In contrast to the methods reliant on Shapley values, the influence functions \cite{koh2020understanding} approach and its derivatives offer a distinct methodology. This approach is applied to a model that has already been trained, assuming complete access to it, including the capability to compute gradients. The procedure involves the sequential input of training samples into the model, followed by an analysis of how their gradients align with those of the test samples. This alignment of gradients is then used to quantify the influence of each training example on the model's predictions for test data, thereby enabling a precise evaluation of the training sample's impact on the predictive performance of the model.

However, the application of influence function methodologies presents significant challenges in interpretability, and more critically, the precision of their measurements degrades as the depth of the neural network increases. This case is comprehensively detailed in the study 'Influence Functions in Deep Learning Are Fragile' by Samyadeep Basu, Philip Pope, and Soheil Feizi \cite{basu2021influence}. An alternative and more robust method, specifically designed for deep networks, was developed by Garima Pruthi and Frederic Liu in their work 'Estimating Training Data Influence by Tracing Gradient Descent'\cite{pruthi2020estimating}. This approach necessitates access to intermediate checkpoints during the training process. For instance, in a notable experiment, checkpoints at 10\%, 50\%, and 70\% of the training iterations were required for a more precise assessment of the influence of training samples on test outcomes. This requirement poses a limitation even in the context of open-source models like stable-diffusion, where the community typically only has access to the final trained model, lacking these intermediate developmental stages.

Hence, we introduce an innovative methodology to assess the impact of a data subset characterized by a shared attribute (here, the attribute in question is the authorship of artworks) on the overall efficacy of image generation. This process is oriented by the said attribute. More precisely, in the context of Stable-Diffusion, a frozen CLIP model plays a pivotal role in steering image generation towards the desired attribute, which in this instance is the stylistic signature of a specific artist. This approach offers a nuanced understanding of how particular data characteristics influence the generative capabilities of the model in producing artistically styled images.

\section{Methodology}

Before proceeding, it is imperative to address a foundational question: Why should we endeavor to calculate the contribution of each participant in the final output? Why not simply divide the reward equally among all contributors? Several rationales substantiate the need for a more nuanced approach. First and foremost, the model has not been trained on the works of every artist in existence but rather on a select subset. Consequently, including the name of a specific artist in the prompt does not necessarily guarantee influence over the output. Secondly, even for those artists on whose works the model has been definitively trained, it does not always exhibit a robust ability to recall and comprehend their unique styles.

Thirdly, the model may not be adept at transferring the style of certain artists to unfamiliar subjects. We have observed that not every subject can be depicted by the model in a given style, as some artists might have an affinity for specific types of content. For instance, if an artist primarily created female portraits, the model would likely find it easier to transfer their style to another female portrait. Conversely, if an artist predominantly painted landscapes, applying such a style to a female portrait may result in subtle or even indiscernible stylistic influences.

This leads to the ensuing inquiry: How can we verify whether the model was trained on the works of a particular artist and, if so, ascertain the extent to which it can reproduce their unique style and creatively adapt it to new content?

Currently, the procedure to confirm a model's capability to reproduce a specific style relies solely on enthusiasts and manual assessment of the results. Is it possible to automate this procedure? The internet offers extensive lists of artists whose styles various generative models, including Stable Diffusion, can reproduce. We will use this data as a validation set for our method. Through crowdsource efforts, a list of 1789 artist names has been compiled \cite{image-style-study-2023-part1}. Each name was added to the prompt, several images were generated, and the results were evaluated by humans. Subsequently, it was noted whether adding each name to the prompt influenced the final generation. As the outcome of such research, 1649 names were recognized by Stable Diffusion, while the remaining 140 were unknown. This data served as our evaluation set. Upon applying our method to this data, we achieved an accuracy score of approximately 92.86\% in the stratified sampling validation procedure.

Our methodology was influenced by a comparable compilation of artist names sourced from a different provider, which contained a reduced number of artists as detailed in \cite{artiststudy2022}. In this dataset, researchers constructed a concise set of validation prompts, which required execution under consistent parameters (W/H: 512, cfg\_scale, optimizer, resolution, number of steps, etc.). Consequently, four distinct images were generated for each of the predetermined prompts (A beach, A Mansion, A Mech, A portrait of a woman). These images were then meticulously assessed, juxtaposed with original artworks, and subsequently categorized with labels—either 'Yes' or 'No'—based on their resemblance with original artworks.

We have augmented the existing methodology with a particular logical framework. When generating images utilizing solely the Stable Diffusion model, and focusing merely on context words (e.g., "a castle"), the results predominantly hinge on the random seed, often yielding highly arbitrary outcomes. In such cases, we do not anticipate any consistent resemblance among the generated images. Conversely, when generating content that also includes an artist's name, we navigate to a specific region within the model's latent space. Here, the generated images consistently adopt a particular style, enhancing their similarity. Thus, the intra-similarity of such images increases, while their resemblance to purely content-driven images (those generated without appending artist names) diminishes. As the model's capability to recognize styles improves, the latent space occupied by the generated images becomes more constrained, amplifying the divergence between purely content-driven and style-influenced images. This differential can serve as a metric for the model's style memorization.

In our study, we encountered scenarios where the application of style was contingent upon the content. For certain prompts, the style was successfully implemented, while for others, it failed to materialize. This observation led us to delineate varying degrees of style comprehension by the generative model. For specific artists, there were clear indications of overfitting. For instance, when introducing the name "Agnes Cecile" into the prompt, a consistent generation of styled female faces was observed, irrespective of the original prompt's intent (refer to Figure X). Notably, only the prompts labeled "a car" achieved the desired outcome. And among the generated images denoting "castles", a singular image accurately portrayed a castle, while the majority of various prompts resembled a female face. Based on these findings, we classified this phenomenon into a category termed "overfitting."

Upon reflection, we recognized the critical importance of well-crafted prompts. Consequently, we began enhancing our set of prompts, systematically evaluating each for its effectiveness in determining style adaptability. In this context, we prepared a set of prompts covering 20 different contexts. Subsequently, we compiled all possible combinations from $n$ number of prompts, where {$n \in \{1, 2, \ldots, 20\}$}. We found that the accuracy of our method plateaued after a subgroup of 5 prompts, indicating that the quality of each individual prompt is more crucial than the quantity of variations in the prompts. The final list of these prompts can be found in appendix \ref{app:example}. The variation of the prompt set presented there achieved an accuracy of 92.66\%. Notably, we also managed to associate specific numerical values with each style, facilitating the quantification of the style's significance, or equivalently, the model's capacity to replicate it. Thus, we established a numerical threshold, enabling us to discern which styles the model has effectively internalized and which it has not.

In our rigorous analysis, we identified several categories beyond the "overfitting" designation.
The subsequent category we discerned pertains to styles that are seamlessly applicable across a majority of contexts from our prompt list, exemplified in Figure 4-b. Following this, we observed instances where the style, although aesthetically pleasing or acceptable, could not be applied consistently across desired content, as illustrated in Figure 4-c. While certain prompts from our list faltered, others prevailed. The final category encompasses styles that remain entirely unrecognized by the model. In such scenarios, no discernible distinction is evident when juxtaposed with a pure content prompt devoid of any stylistic influence. For clarity, we constructed a graphical representation delineating the ratings and manually demarcated boundaries between these categories and provide some examples in Table~\ref{tab:table_2}

\begin{table}[h!]
    \centering
    \begin{tabularx}{\textwidth}{cccc}
        \textbf{'a beach'} & \textbf{'a car'} & \textbf{'a human'} & \textbf{'a machine'} \\
        \includegraphics[width=0.24\textwidth]{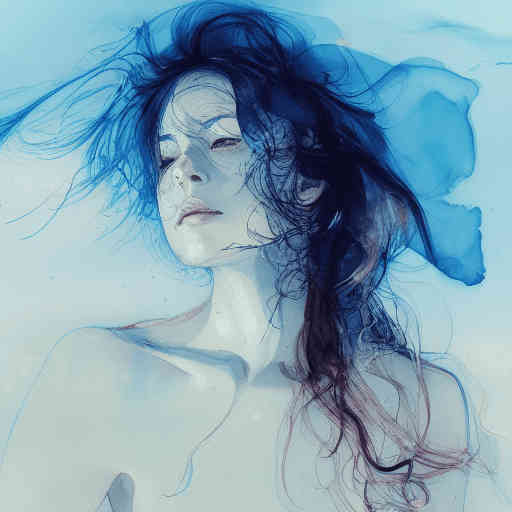} & \includegraphics[width=0.24\textwidth]{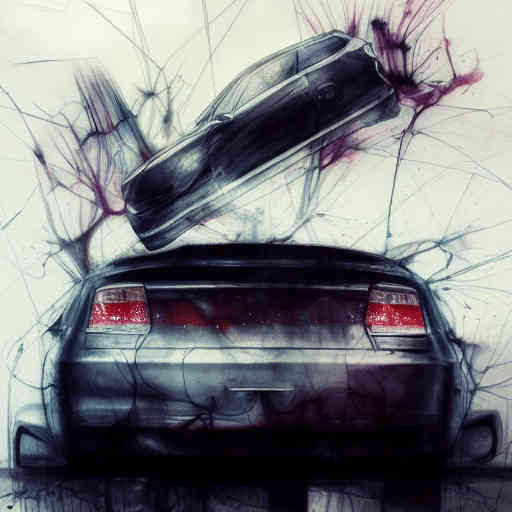} & \includegraphics[width=0.24\textwidth]{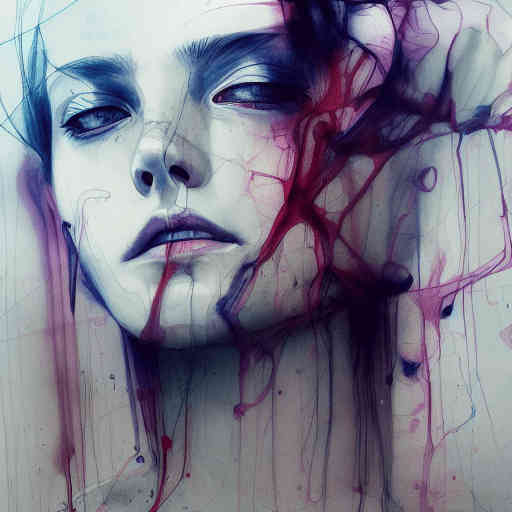} & \includegraphics[width=0.24\textwidth]{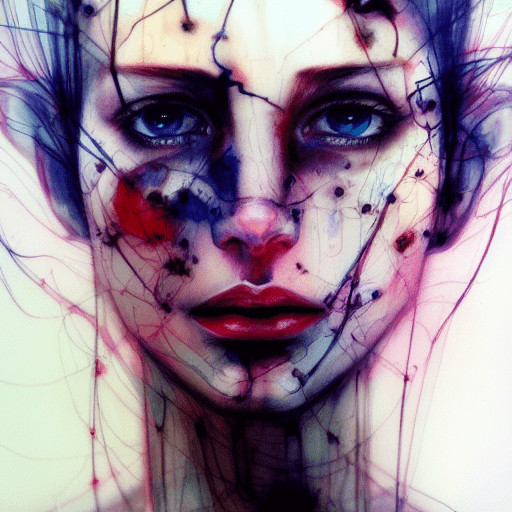} \\
        \multicolumn{4}{c}{'by Agnes Cecile' [ Category: overfitted ]} \\
        \includegraphics[width=0.24\textwidth]{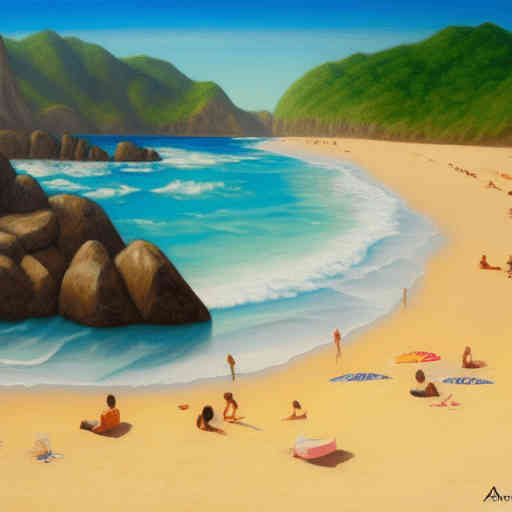} & \includegraphics[width=0.24\textwidth]{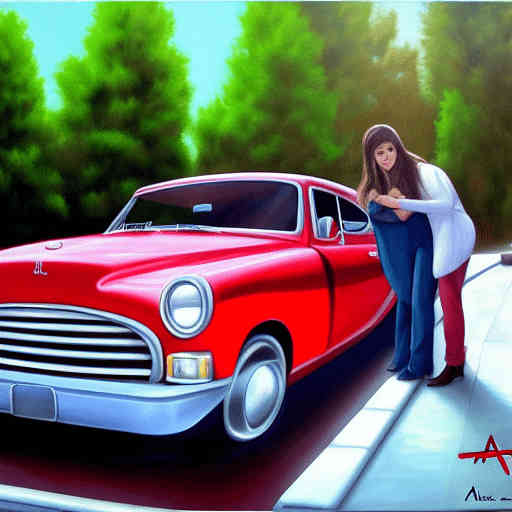} & \includegraphics[width=0.24\textwidth]{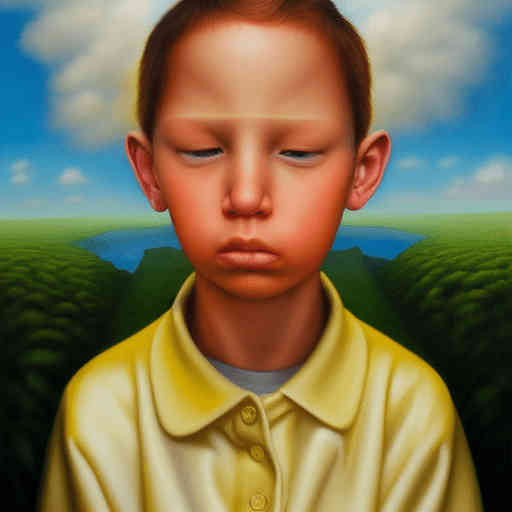} & \includegraphics[width=0.24\textwidth]{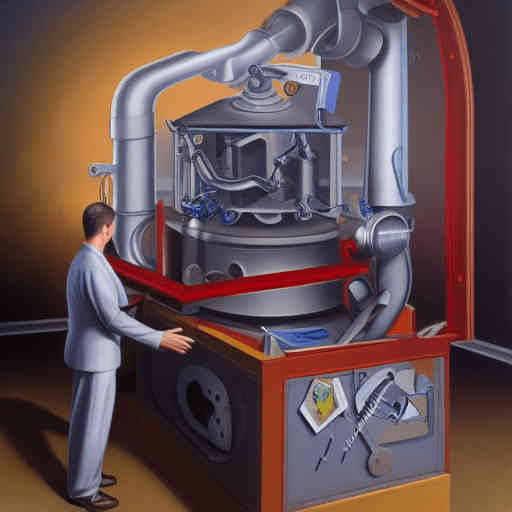} \\
        \multicolumn{4}{c}{'by Alex Alemany' [ Category: balanced ]} \\
        \includegraphics[width=0.24\textwidth]{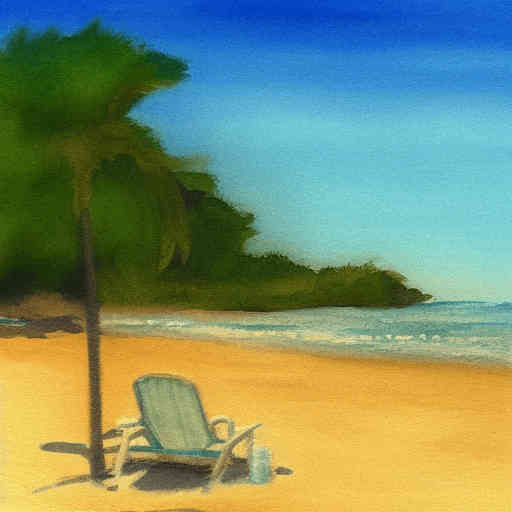} & \includegraphics[width=0.24\textwidth]{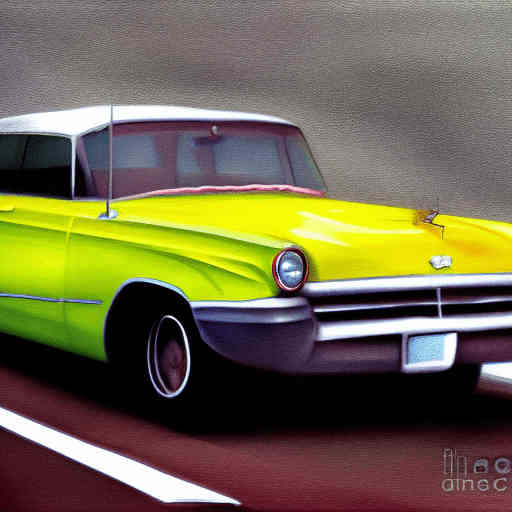} & \includegraphics[width=0.24\textwidth]{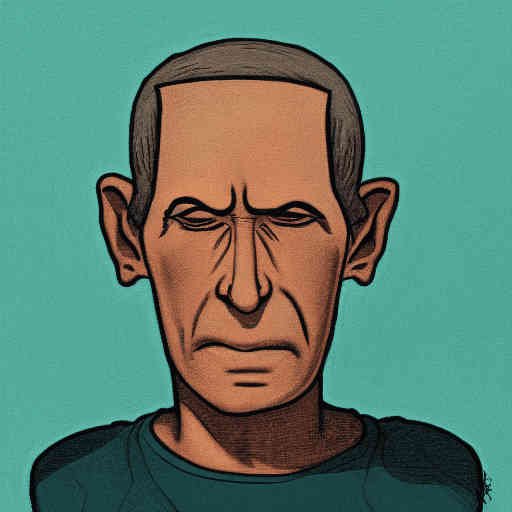} & \includegraphics[width=0.24\textwidth]{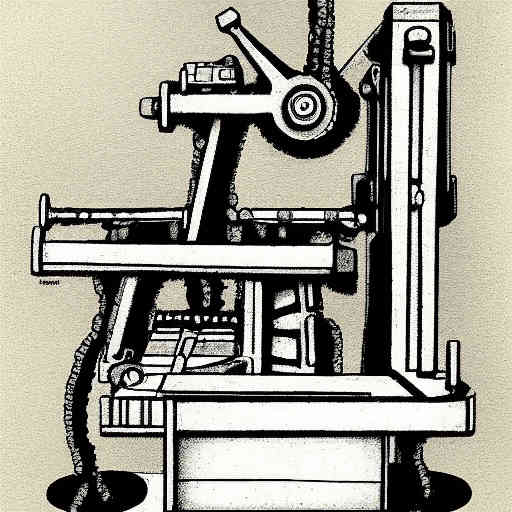} \\
        \multicolumn{4}{c}{'by Rick Guidice' [ Category: underfitted ]} \\
        \includegraphics[width=0.24\textwidth]{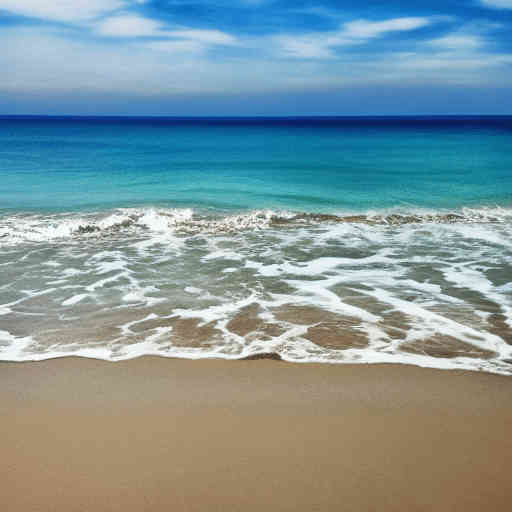} & \includegraphics[width=0.24\textwidth]{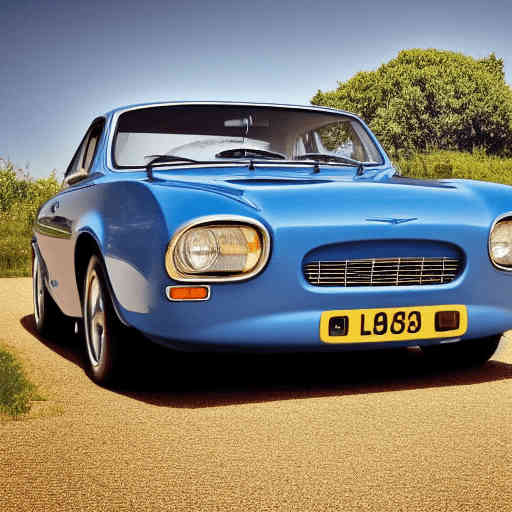} & \includegraphics[width=0.24\textwidth]{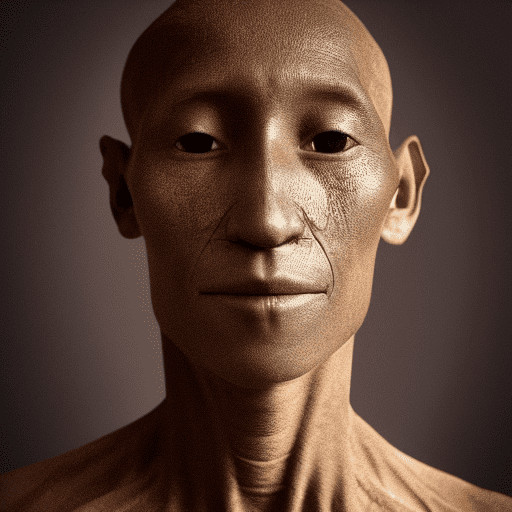} & \includegraphics[width=0.24\textwidth]{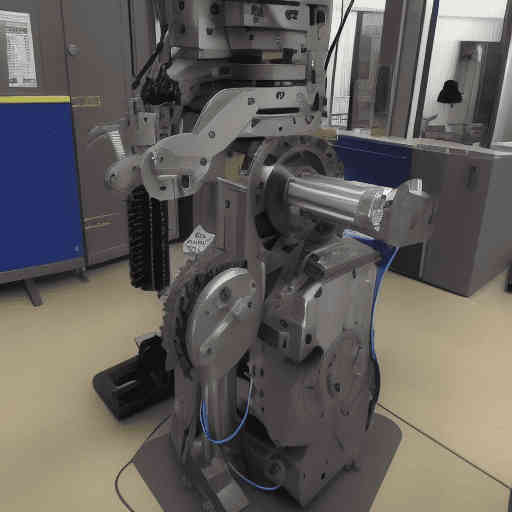} \\
        \multicolumn{4}{c}{pure SD-v1-5 [ Category: untrained ]} \\
    \end{tabularx}
    \caption{Example of 1 style from each category with four different prompts.}
    \label{tab:table_2}
\end{table}

\begin{itemize}
    \item Overfitting to the style (as previously exemplified)
    \item Equilibrium between style comprehension and its transferability to novel content
    \item Style transfer is restricted to objects present in the original artworks
    \item The style is non-reproducible.
\end{itemize}

As a result, we developed a filter for artists as well as a method to numerically evaluate the model's proficiency in reproducing a particular style. With such a method in place, we can preemptively determine whether the model possesses mastery over a given style, and only then proceed to assess the influence of that style on the final generation. If a style is deemed unknown or poorly reproduced, we simply exclude it from the calculations.

\section{Experiments and Results}

\subsection{Evaluation of a single style}
\label{subsec:single_style}

Let us now turn to our primary task: the equitable distribution of rewards, which we will address using Shapley Values. Consider a simple example involving two participants: one responsible for style and the other for content. For style, we select as many available original works of the desired artist as possible. For content, we use Stable Diffusion generations based on a given prompt, without mentioning the artist's name or any other tags. The target image (or the result of their collaboration) is generated using a prompt that combines both content and style.

To identify the most relevant original works of an artist, we search for those with the closest cosine distance between embeddings obtained from the visual-transformer component of CLIP. For clarity, we select the top 9 closest works. Alternatively, we wait for a sharp decline in the similarity score among the works sorted in descending order and truncate everything below this threshold.

\begin{figure}[h]
    \centering
    
    \begin{subfigure}{\textwidth}
        \centering
        \includegraphics[width=\textwidth]{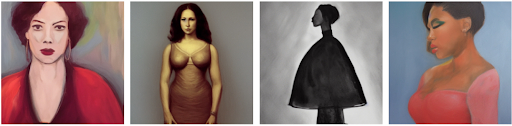}
        \caption{prompt: \textbf{'a woman'}}
        \label{fig5:sub1}
    \end{subfigure}
    \\
    
    \begin{subfigure}{\textwidth}
        \centering
        \includegraphics[width=\textwidth]{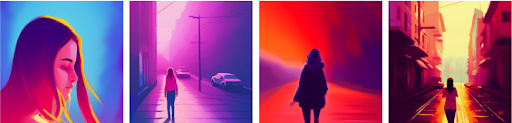}
        \caption{prompt: \textbf{'a woman, by Alena Aenami'}}
        \label{fig5:sub2}
    \end{subfigure}
    \\

    \begin{subfigure}{\textwidth}
        \centering
        \includegraphics[width=\textwidth]{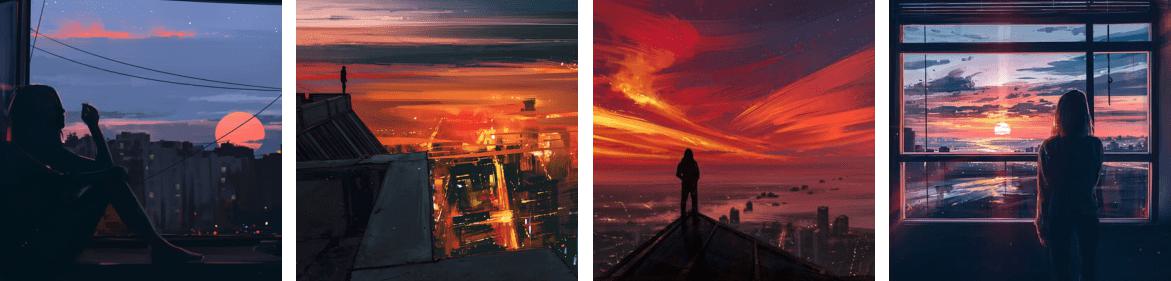}
        \caption{original artworks by \textit{Alena Aenami}}
        \label{fig5:sub3}
    \end{subfigure}    
    \caption{Participants (a and c) and the outcome (b) of their interaction for computing Shapley Values.}
    \label{fig5:stacked_images}
\end{figure}

\begin{figure}[h]
  \centering
  \begin{subfigure}[b]{0.6\textwidth}
    \includegraphics[width=\linewidth]{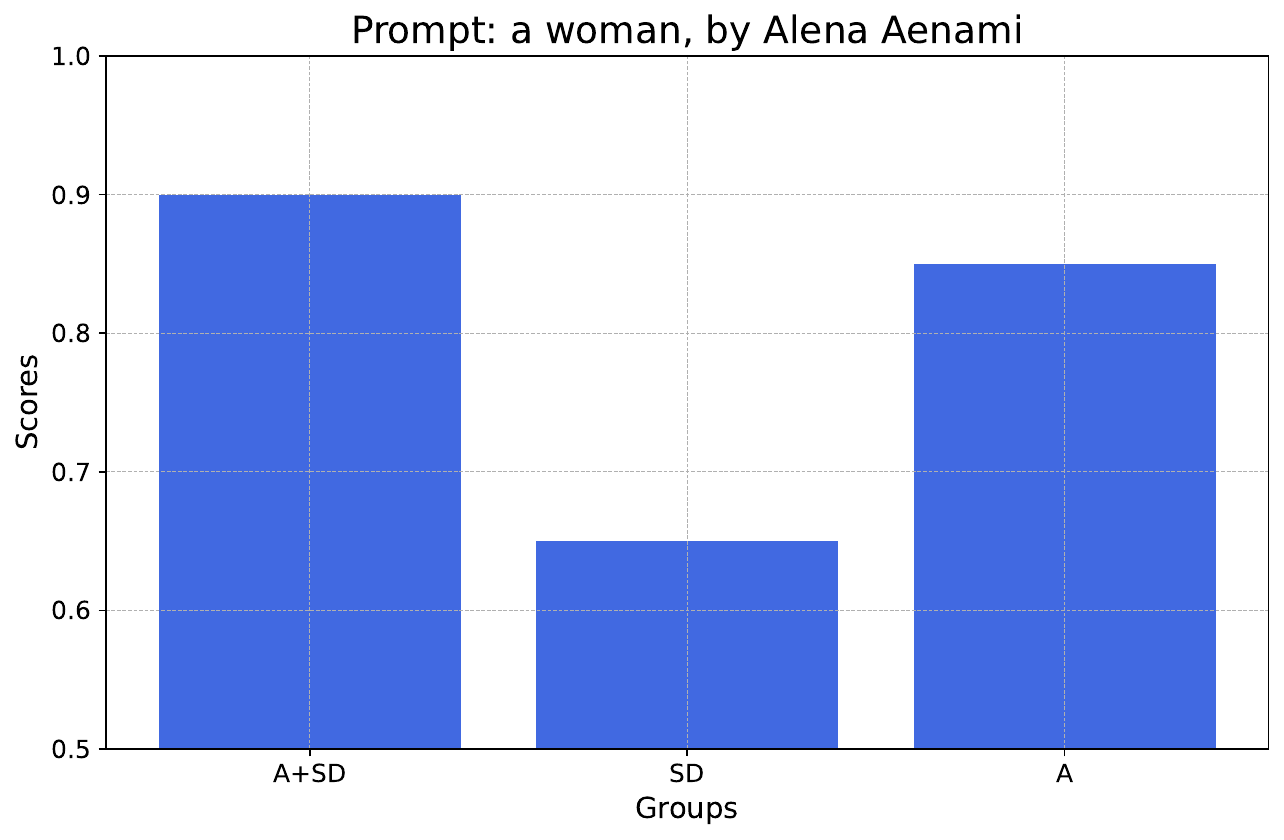}
    \caption{}
    \label{fig6:image1}
  \end{subfigure}%
  \begin{subfigure}[b]{0.4\textwidth}
    \includegraphics[width=\linewidth]{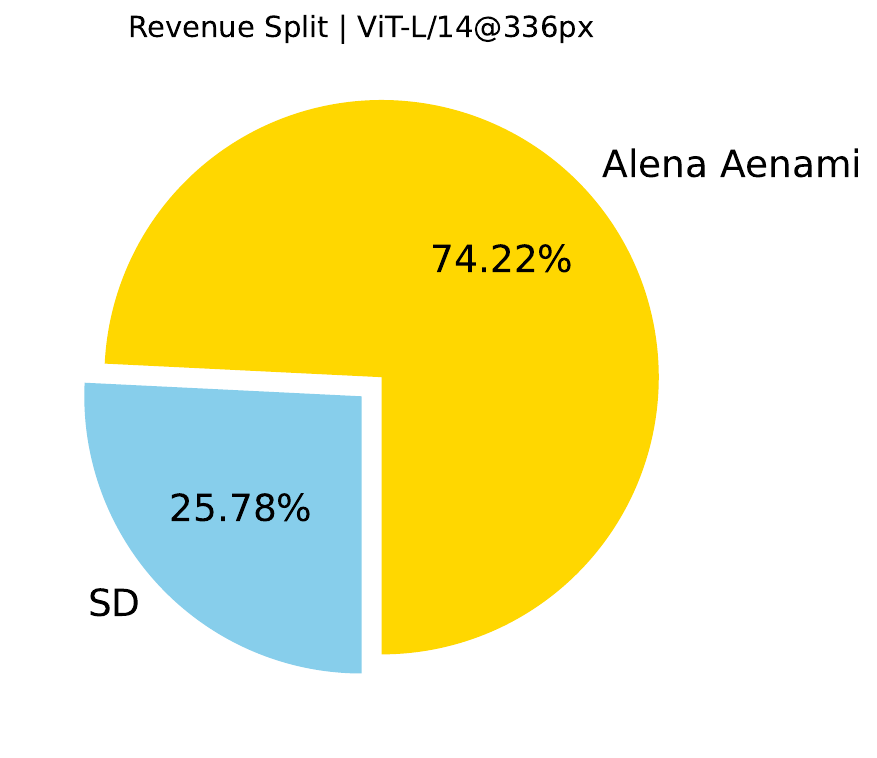}
    \caption{}
    \label{fig6:image2}
  \end{subfigure}
 \captionsetup{justification=centering}
  \caption{(a) - similarity score between generation, original artworks and pure SD model, (b) - Shapley Values derived from embeddings generated by the CLIP.}
  \label{fig6}
\end{figure}

Thus, we form a dataset as depicted in Figure~\ref{fig5:stacked_images}. We aim to assess the extent to which the images from the second and third columns influence the first one. As previously mentioned, we evaluate this influence by measuring the similarity score between the style-less generations, the original artworks, and the final result. This evaluation produces the graph shown in Figure~\ref{fig6:image1}. From this graph, it's evident that the original artworks by the artist bear a significant resemblance to the final generation. This observation can also be visually corroborated using Figure 3. Now, it is imperative to ascertain the correct methodology for employing the available data to compute the Shapley Values. The foundational equation is delineated as follows:

\begin{equation} \label{eq:eq1}
    \phi_i(v) = \sum_{S \subseteq N \setminus \{i\}} \frac{|S|! \cdot (|N| - |S| - 1)!}{|N|!} \left( v(S \cup \{i\}) - v(S) \right)
\end{equation}

where \(\phi_i(v)\) represents the Shapley value for player \(i\), \(N\) is the set of all players, \(S\) is a subset of \(N\) without \(i\), \(v(S)\) is the value function for coalition \(S\), \(|S|\) is the number of players in \(S\), \(|N|\) is the total number of players, and then we perform the sum over all \(S \subseteq N \setminus \{i\}\).

We then adapt this equation to accommodate our available data. Each player is represented as one of two categories: a group of original artworks (training data) or a group of generated images (generated data). We consider the maximum score achievable by such a collaboration as a constant, set to 1. A collaboration that attains a score of 1 is the group of images resulting from a combination of all available data and styles. Other coalitions receive lower scores, which are evaluated based on the embedding similarity between those coalitions and the target coalition, which represents the most rich and diverse amalgamation of available data and models.

For instance, when considering the combination of two entities — 'Alena Aenami,' representing an artist who supplies training data, and 'StableDiffusion-v1-5,' functioning as a generative model — we observe the emergence of three distinct coalitions. Each coalition is associated with specific prompts that activate its respective contribution (in cases where activation is required):

\textit{'Alena Aenami'} functions as the provider of training data.
\textit{'SD'} acts as the source of generated content, with the prompt: \textbf{'a woman.'}
\textit{'SD + Alena Aenami'} represents a synergy of the generative model with Alena Aenami's training data, utilizing the prompt: \textbf{'a woman, by Alena Aenami'}. 
(see Figure~\ref{fig6:image2})
Subsequently, we gather a selection of images from each collaborative effort, apply Equation ~\ref{eq:eq1} for analysis, and achieve the following outcomes: \texttt{\{'Alena Aenami':0.7422, 'Stable Diffusion':0.2578\}}. This suggests that in the event of a sale of an image generated by this model using the specified prompt, the proceeds should be apportioned between the developers of the generative model and the artist in such a ratio. This division is proposed to ensure that the collaborative effort is equitably recognized and that all contributing parties benefit commensurately.

\begin{figure}[h]
    \centering
    
    \begin{subfigure}{\textwidth}
        \centering
        \includegraphics[width=\textwidth]{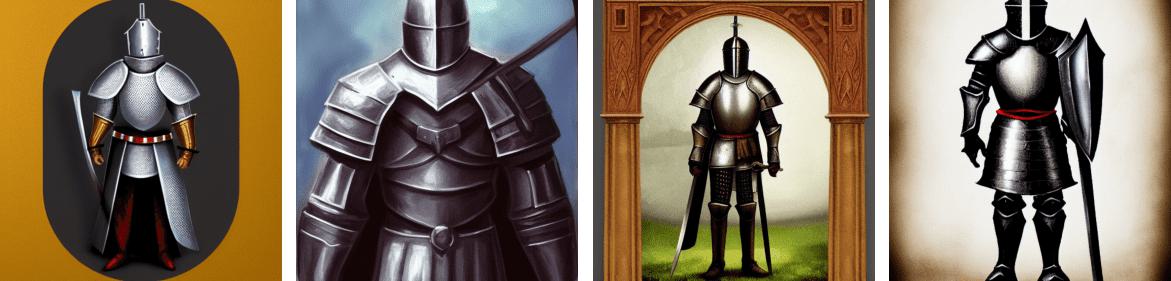}
        \caption{prompt: \textbf{a knight}}
        \label{fig7:sub1}
    \end{subfigure}
    \\
    
    \begin{subfigure}{\textwidth}
        \centering
        \includegraphics[width=\textwidth]{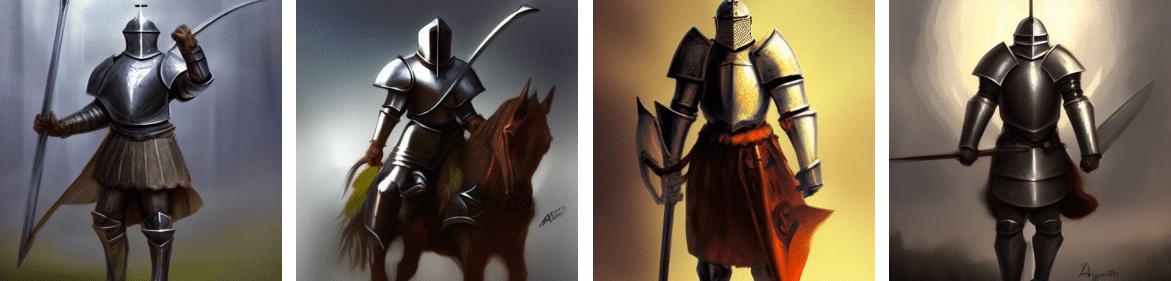}
        \caption{prompt: \textbf{a knight,  by Andrzej Dybowski}}
        \label{fig7:sub2}
    \end{subfigure}
    \\
   
    \begin{subfigure}{\textwidth}
        \centering
        \includegraphics[width=\textwidth]{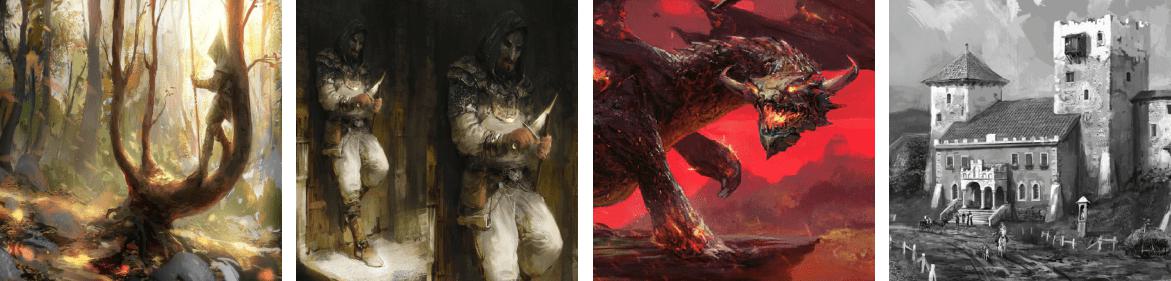}
        \caption{original artworks by \textit{Andrzej Dybowski}}
        \label{fig7:sub3}
    \end{subfigure}    

    \caption{Participants (a and c) and the outcome (b) of their interaction for computing Shapley Values.}
    \label{fig7:stacked_images}
\end{figure}

\begin{figure}[h]
  \centering
  \begin{subfigure}[b]{0.6\textwidth}
    \includegraphics[width=\linewidth]{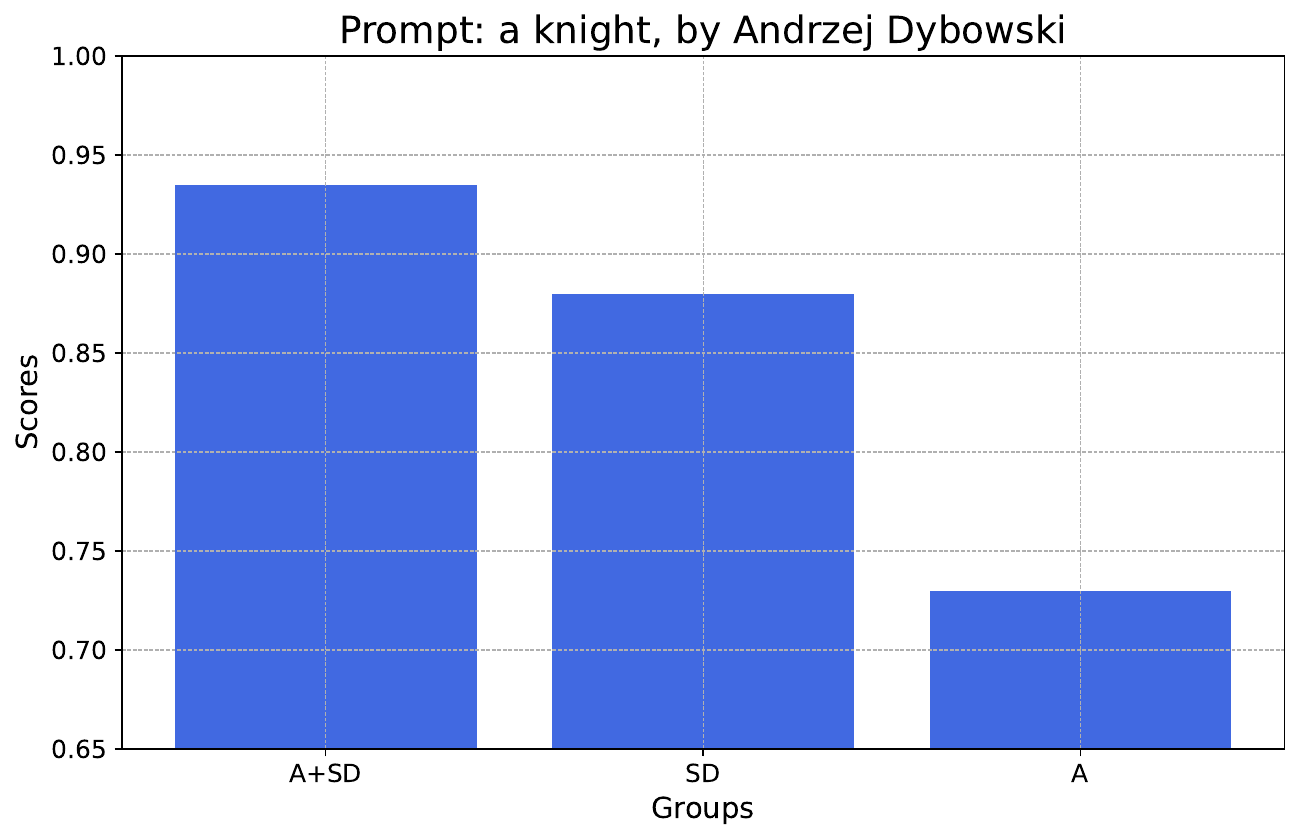}
    \caption{}
        \label{fig8:image1}
  \end{subfigure}%
  \begin{subfigure}[b]{0.4\textwidth}
    \includegraphics[width=\linewidth]{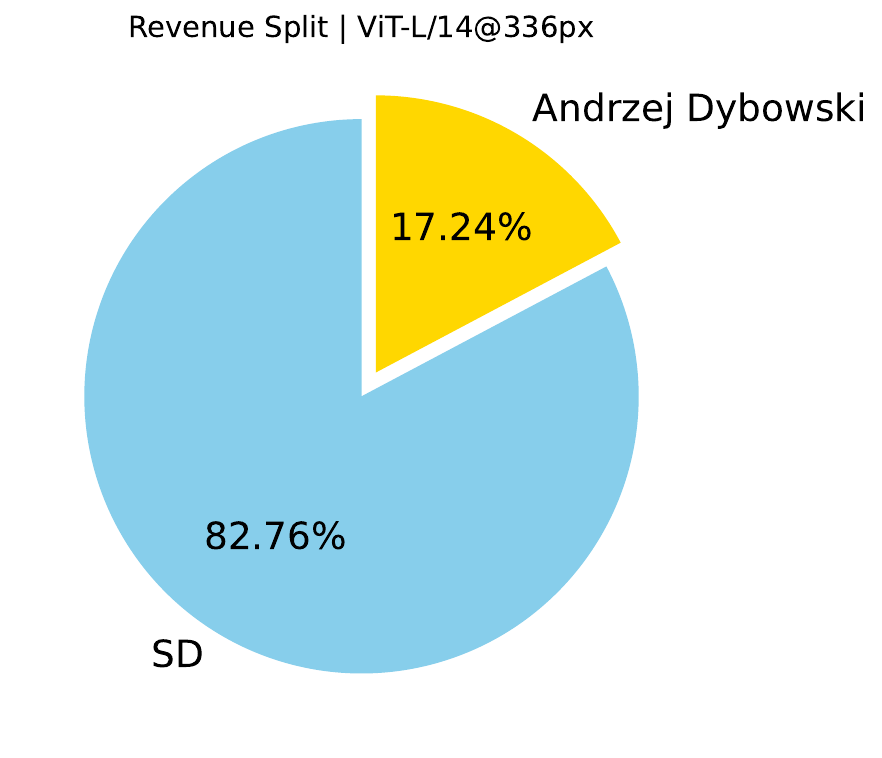}
    \caption{}
        \label{fig8:image2}
  \end{subfigure}
 \captionsetup{justification=centering}
  \caption{(a) - Similarity score between generation, original artworks and pure SD model, (b) - Shapley Values derived from embeddings generated by the CLIP}
  \label{fig8}
\end{figure}

Let us now examine how the algorithm behaves when the artist's name is unknown to the model. Consider this using the prompt: \textbf{'a knight, by Andrzej Dybowski'}. As observed in Figure~\ref{fig7:stacked_images}, even after adding the artist's name to the prompt, the generated images still closely resemble those produced by Stable Diffusion without any modifications. Furthermore, among the artist's works, we could not find any images that resembled the generated outputs. The generated images diverge both in style and content from what we obtained. Ideally, all rewards should have been attributed to the generative model. However, since the calculation of Shapley Values is based on embeddings, whose cosine similarity values are significantly greater than 0 (Figure~\ref{fig8:image1}), there emerges a bias, resulting in an artist with zero influence receiving a small share of the reward (Figure~\ref{fig8:image2}). To minimize the impact of this effect, we previously introduced a selection (filtering) phase for "recognizable" artists. Any artists who did not pass this initial stage automatically receive a zero share of the reward.

\begin{figure}[h]
    \centering
    
    \begin{subfigure}{\textwidth}
        \centering
        \includegraphics[width=\textwidth]{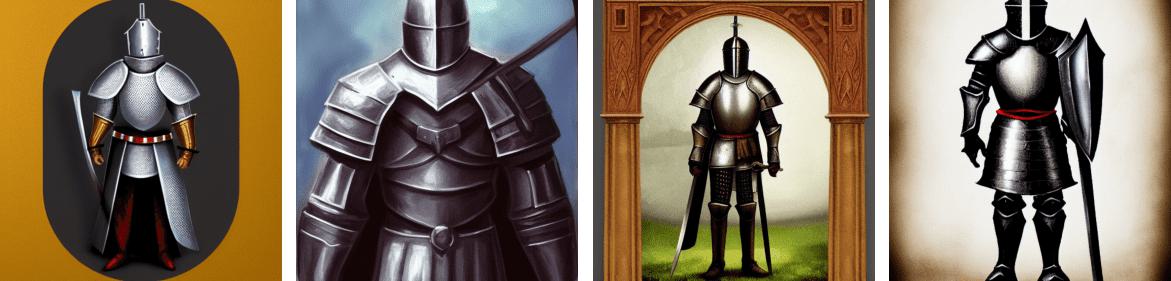}
        \caption{prompt: \textbf{a knight}}
        \label{fig10:sub1}
    \end{subfigure}
    \\
    
    \begin{subfigure}{\textwidth}
        \centering
        \includegraphics[width=\textwidth]{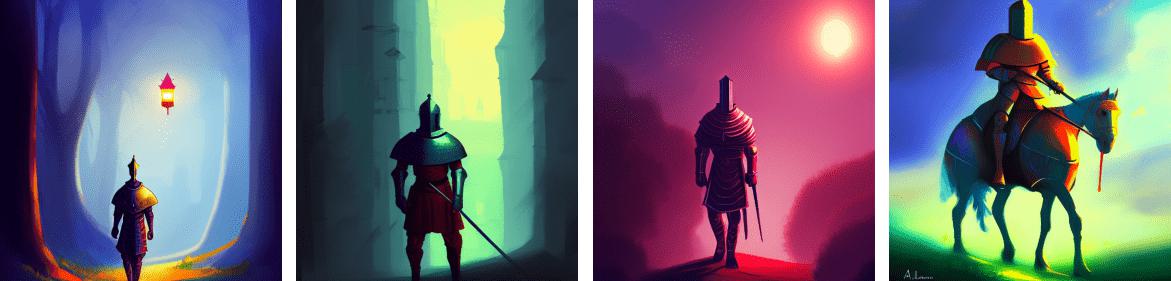}
        \caption{prompt: \textbf{a knight,  by Alena Aenami}}
        \label{fig10:sub2}
    \end{subfigure}
    \\
   
    \begin{subfigure}{\textwidth}
        \centering
        \includegraphics[width=\textwidth]{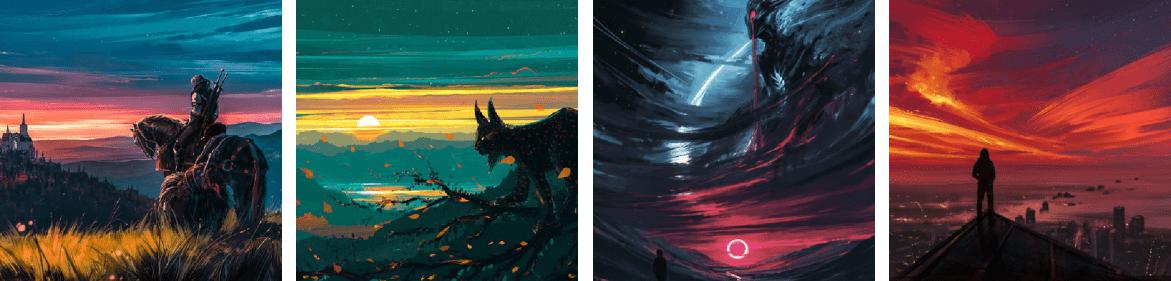}
        \caption{original artworks by \textit{Alena Aenami}}
        \label{fig10:sub3}
    \end{subfigure}    

    \caption{Participants (a and c) and the outcome (b) of their interaction for computing Shapley Values.}
    \label{fig10:stacked_images}
\end{figure}

\begin{figure}[h]
  \centering
  \begin{subfigure}[b]{0.6\textwidth}
    \includegraphics[width=\linewidth]{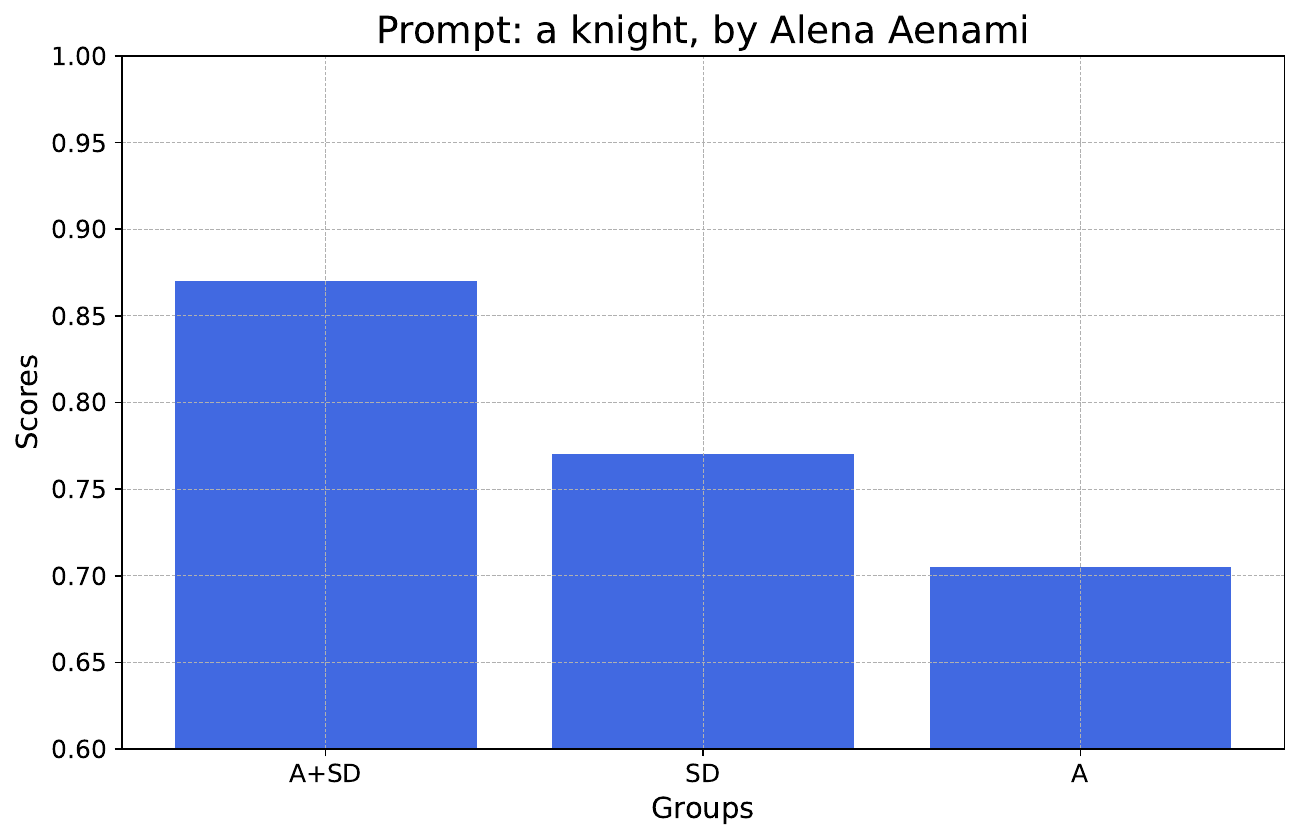}
    \caption{}
        \label{fig11:image1}
  \end{subfigure}%
  \begin{subfigure}[b]{0.4\textwidth}
    \includegraphics[width=\linewidth]{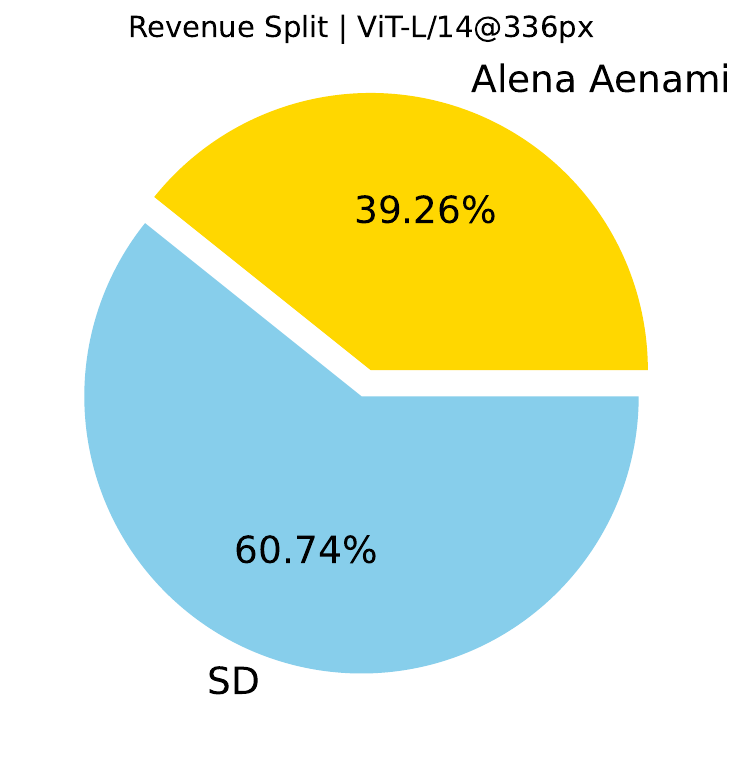}
    \caption{}
        \label{fig11:image2}
  \end{subfigure}
 \captionsetup{justification=centering}
  \caption{(a) - Similarity score between generation, original artworks and pure SD model, (b) - Shapley Values derived from embeddings generated by the CLIP}
  \label{fig11}
\end{figure}

Now, we propose to consider an example (Figure~\ref{fig10:stacked_images}) where Stable Diffusion received a greater reward (Figure~\ref{fig11}) because it was able to transfer the artist's style to a new context (object).
*It is worth noting that in our experiments, we encountered a tendency for CLIP embeddings to assign a higher similarity score to images with similar content, regardless of their style. Therefore, such behavior often skews Shapley Value estimates in favor of a larger reward for Stable Diffusion. However, we are actively addressing this issue.

\begin{figure}[h]
    \centering
    
    \begin{subfigure}{\textwidth}
        \centering
        \includegraphics[width=\textwidth]{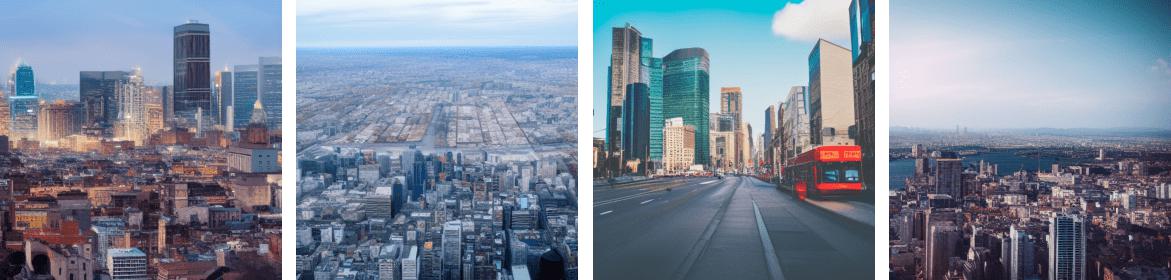}
        \caption{prompt: \textbf{a city}}
        \label{fig13:sub1}
    \end{subfigure}
    \\
    
    \begin{subfigure}{\textwidth}
        \centering
        \includegraphics[width=\textwidth]{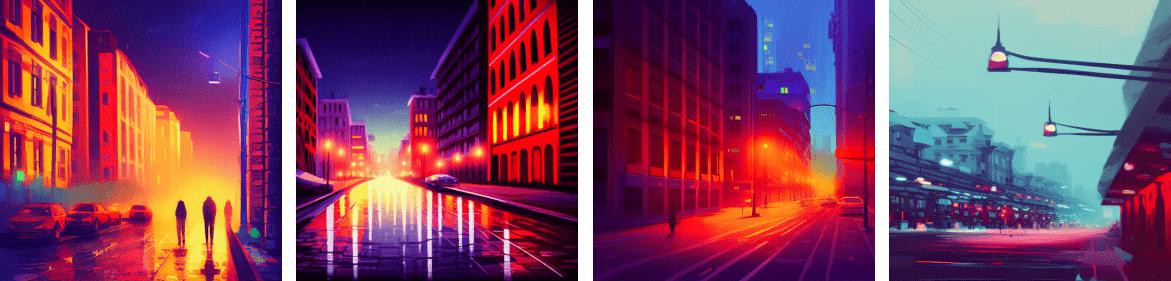}
        \caption{prompt: \textbf{a city, by Alena Aenami}}
        \label{fig13:sub2}
    \end{subfigure}
    \\
   
    \begin{subfigure}{\textwidth}
        \centering
        \includegraphics[width=\textwidth]{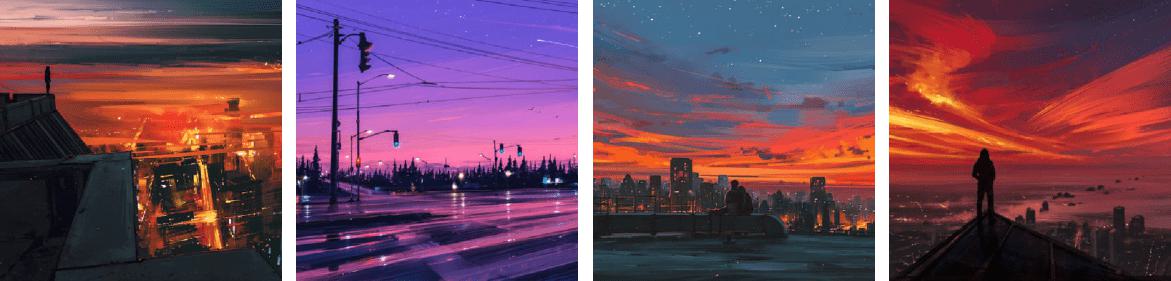}
        \caption{original artworks by \textit{Alena Aenami}}
        \label{fig13:sub3}
    \end{subfigure}    

    \caption{Participants (a and c) and the outcome (b) of their interaction for computing Shapley Values.}
    \label{fig13:stacked_images}
\end{figure}

\begin{figure}[h]
    \centering
    \begin{minipage}{0.55\textwidth}
        \centering
        \includegraphics[width=\linewidth]{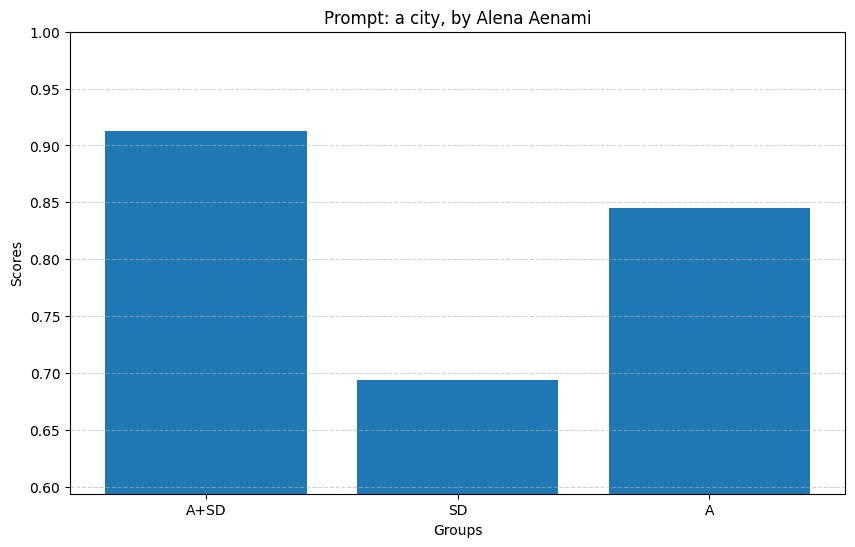}
        \caption{Similarity score between generation, original artworks and pure SD model}
        \label{fig14:image1}
    \end{minipage}\hfill
    \begin{minipage}{0.40\textwidth}
        \centering
        \includegraphics[width=\linewidth]{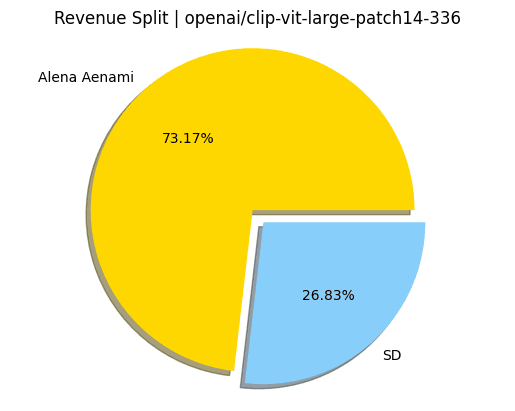}
        \caption{Shapley Values derived from embeddings generated by the CLIP}
        \label{fig15:image1}
    \end{minipage}
\end{figure}

Now, let's present an example in which the content of the generated image overlaps with content found in the original works of the artist, and, in addition to the content, the artist's style is also transferred to the generated image (Figure~\ref{fig13:stacked_images}). In such a scenario, it is justifiable for the artist to claim a significant portion of the reward (Figures~\ref{fig14:image1}-\ref{fig15:image1}).

\begin{figure}[h]
    \centering
    
    \begin{subfigure}{\textwidth}
        \centering
        \includegraphics[width=\textwidth]{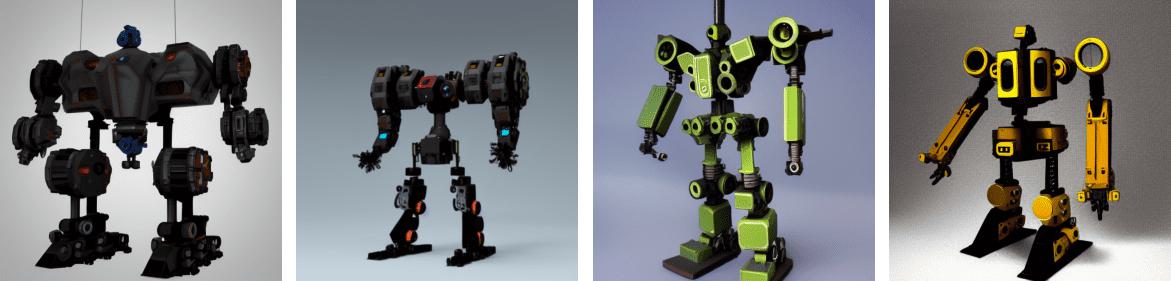}
        \caption{prompt: \textbf{a mech}}
        \label{fig16:sub1}
    \end{subfigure}
    \\
    
    \begin{subfigure}{\textwidth}
        \centering
        \includegraphics[width=\textwidth]{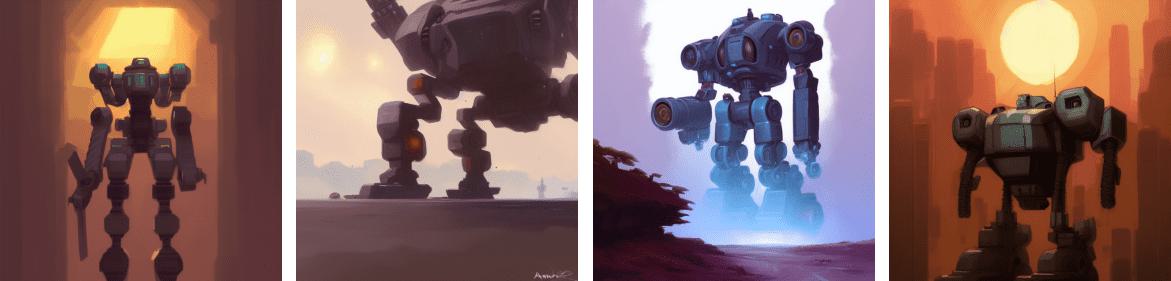}
        \caption{prompt: \textbf{a mech, by Alena Aenami}}
        \label{fig16:sub2}
    \end{subfigure}
    \\
   
    \begin{subfigure}{\textwidth}
        \centering
        \includegraphics[width=\textwidth]{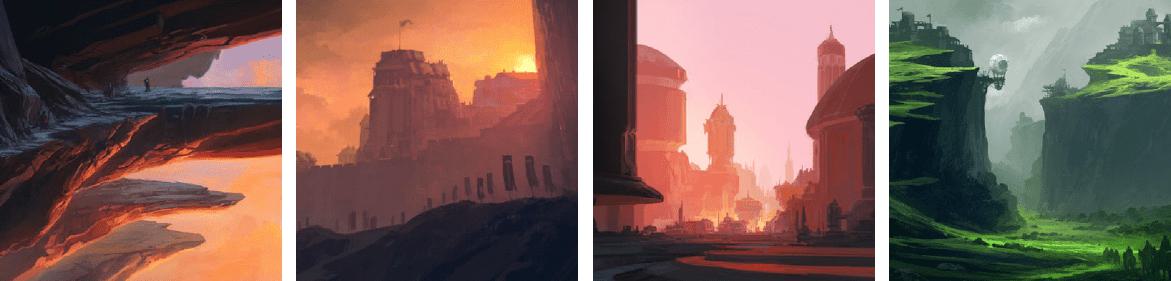}
        \caption{original artworks by \textit{Alena Aenami}}
        \label{fig16:sub3}
    \end{subfigure}    

    \caption{Participants (a and c) and the outcome (b) of their interaction for computing Shapley Values.}
    \label{fig16:stacked_images}
\end{figure}

\begin{figure}[h]
    \centering
    \begin{minipage}{0.55\textwidth}
        \centering
        \includegraphics[width=\linewidth]{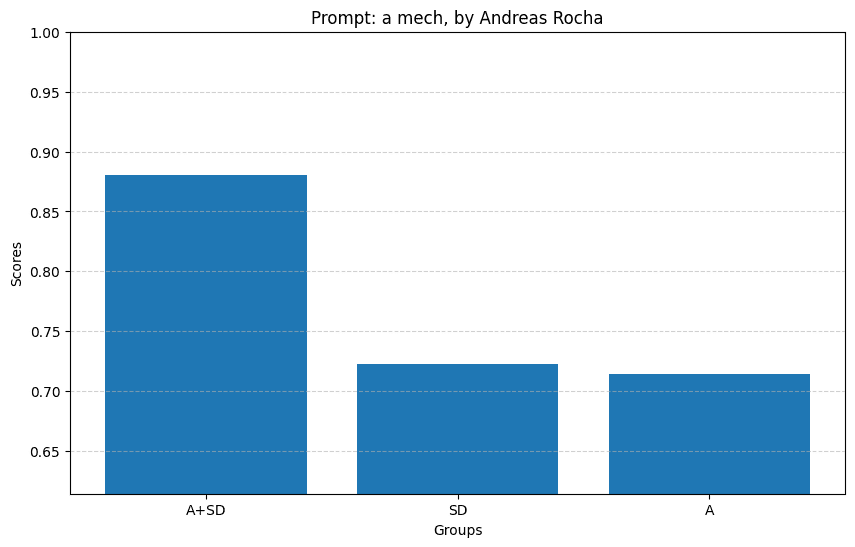}
        \caption{Similarity score between generation, original artworks and pure SD model}
        \label{fig17:image1}
    \end{minipage}\hfill
    \begin{minipage}{0.40\textwidth}
        \centering
        \includegraphics[width=\linewidth]{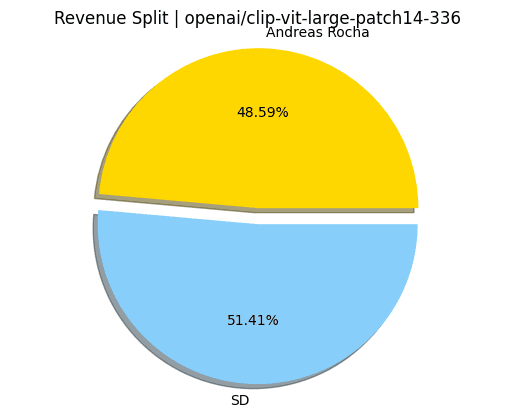}
        \caption{Shapley Values derived from embeddings generated by the CLIP}
        \label{fig18:image1}
    \end{minipage}
\end{figure}

Lastly, let's discuss the most intriguing example where the content is identical to that generated by Stable Diffusion, while the style is unmistakably attributed to the artist (Figure~\ref{fig16:stacked_images}). In such a scenario, the contributions of the participants should be nearly equivalent, which is indeed the case, as illustrated in Figures~\ref{fig17:image1}-\ref{fig18:image1}.

As can be observed, our approach effectively addresses instances where the style of a particular artist, on whose works the Stable Diffusion was trained, is used to impart a more artistic and aesthetically pleasing appearance to some content. In the following subsection, we will explore the more complex scenario involving the interplay of multiple artists.

\subsection{Evaluation of multiple styles}
\label{subsec:multiple_style}

Let's now delve into a more intricate scenario where two styles are incorporated within a single prompt. In the previous subsection, we had two participants and their collaboration (A - original artworks of artist A, SD - model generation without adding the artist's name to the prompt, A+SD - generation utilizing the artist's name).

Now, we have three participants:
\begin{itemize}[itemsep=0pt]
    \item A - Original artworks of artist "A"
    \item B - Original artworks of artist "B"
    \item SD - Model generation based on the provided prompt
\end{itemize}
And there are four collaborations stemming from these:
\begin{itemize}[itemsep=0pt]
    \item A + B - A collective pool of artworks from artists A and B
    \item A + SD - Generation based on the prompt and the name of artist A
    \item B + SD - Generation based on the prompt and the name of artist B
    \item A + B + SD - Generation based on the prompt, the name of artist A, and the name of artist B.
\end{itemize}

An intriguing implication of this method is that we can also evaluate the significance of the order in which an artist's name is inputted into the prompt and determine whether this can substantially influence the final result. For instance, we'll select two artists, A and B, and examine two prompts: “a content, by artist A, by artist B” and “a content, by artist B, by artist A”. We will then observe if the contribution of each style varies based on this order.

\begin{figure}[h]
    \centering

    \begin{subfigure}{0.8\textwidth}
        \centering
        \includegraphics[width=\textwidth]{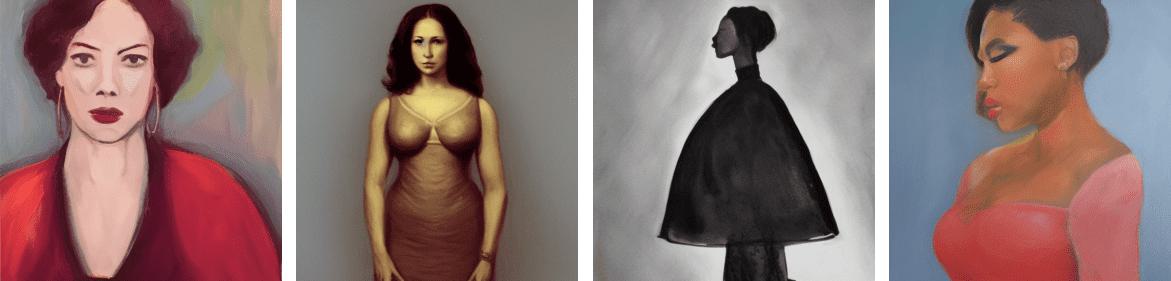}
        \caption{prompt: \textit{a woman}}
        \label{fig19:sub1}
    \end{subfigure}
    \\

    \begin{subfigure}{0.8\textwidth}
        \centering
        \includegraphics[width=\textwidth]{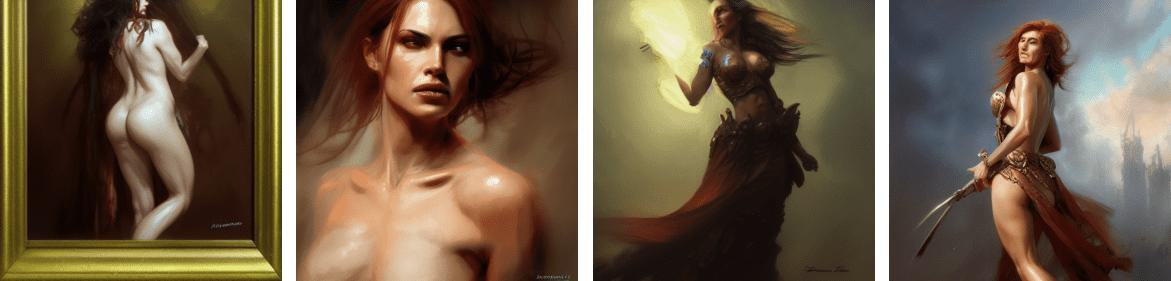}
        \caption{prompt: \textit{a woman, by Antonio J. Manzanedo}}
        \label{fig19:sub2}
    \end{subfigure}
    \\
    
    \begin{subfigure}{0.8\textwidth}
        \centering
        \includegraphics[width=\textwidth]{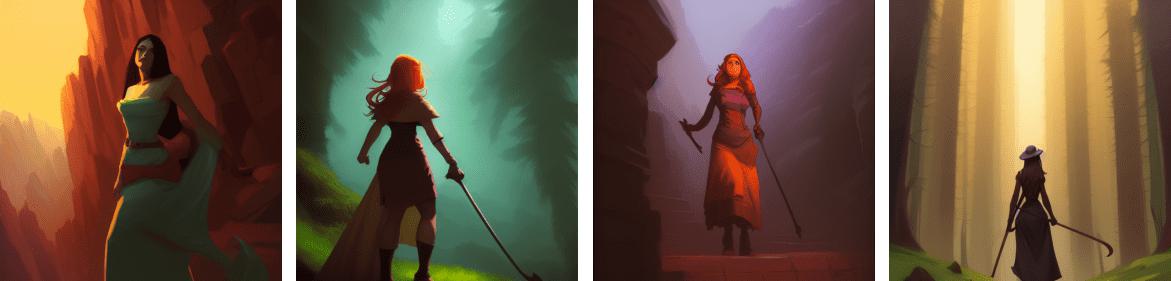}
        \caption{prompt: \textit{a woman, by Andreas Rocha}}
        \label{fig19:sub3}
    \end{subfigure}
    \\
    
    \begin{subfigure}{0.8\textwidth}
        \centering
        \includegraphics[width=\textwidth]{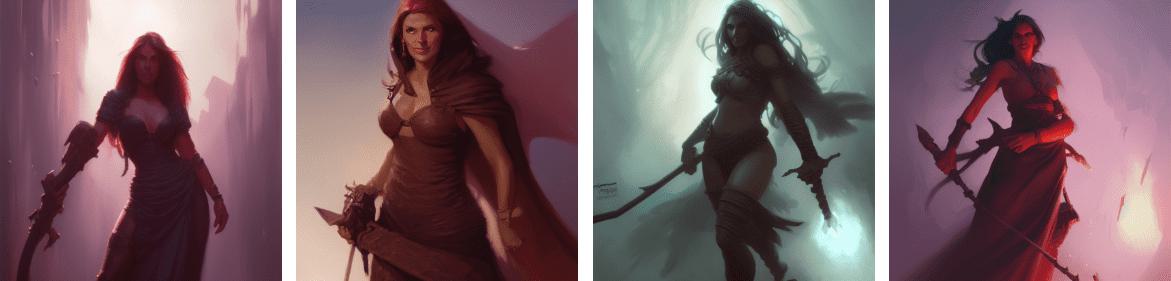}
        \caption{prompt: \textit{a woman, by Andreas Rocha, by Antonio J. Manzanedo}}
        \label{fig19:sub4}
    \end{subfigure}
    \\
   
    \begin{subfigure}{0.8\textwidth}
        \centering
        \includegraphics[width=\textwidth]{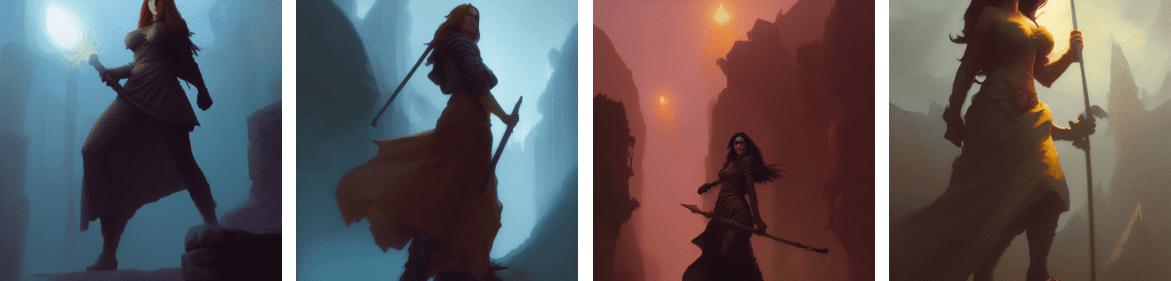}
        \caption{prompt: \textit{a woman, by Antonio J. Manzanedo, by Andreas Rocha}}
        \label{fig19:sub5}
    \end{subfigure}    

    \caption{Example of individual styles and their interactions.}
    \label{fig19:stacked_images}
\end{figure}

Visually (see Figure \ref{fig19:stacked_images}), it's evident that by swapping the order of the artists' names in the prompt, we alter the final generation. Furthermore, we observe that the artist's name listed first exerts a more significant influence than the one following it. We aim to substantiate this observation using Shapley Values.

\begin{figure}[h]
    \centering
    \begin{minipage}{0.45\textwidth}
        \centering
        \includegraphics[width=\linewidth]{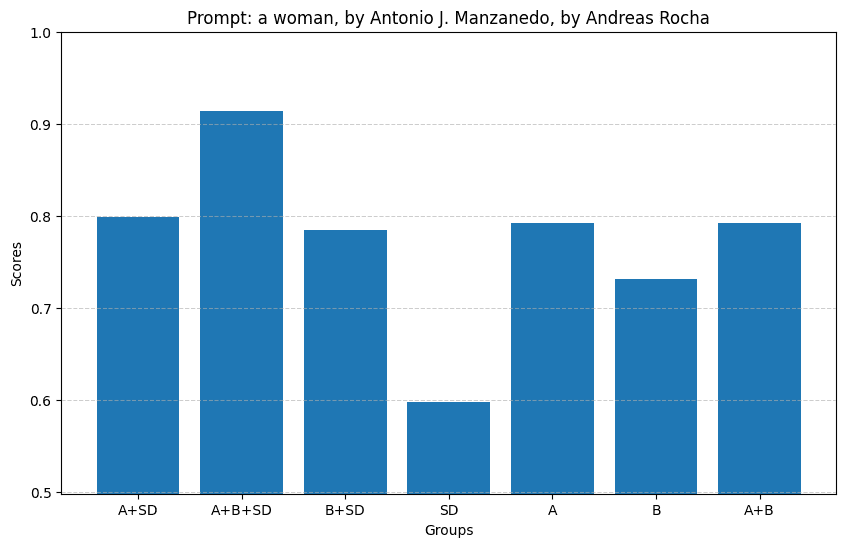}
        \caption{Similarity score between generation, original artworks and pure SD model}
        \label{fig20:image1}
    \end{minipage}\hfill
    \begin{minipage}{0.45\textwidth}
        \centering
        \includegraphics[width=\linewidth]{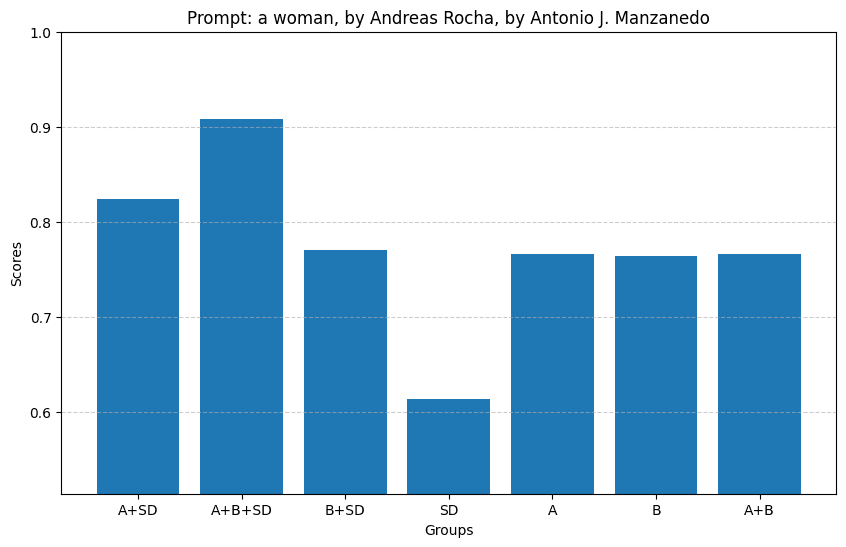}
        \caption{Similarity score between generation, original artworks and pure SD model}
        \label{fig21:image1}
    \end{minipage}
\end{figure}

\begin{figure}[h]
    \centering
    \begin{minipage}{0.45\textwidth}
        \centering
        \includegraphics[width=\linewidth]{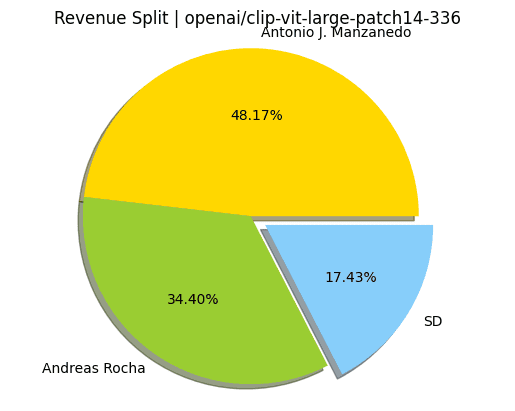}
        \caption{Shapley Values derived from embeddings generated by the CLIP}
        \label{fig22:image1}
    \end{minipage}\hfill
    \begin{minipage}{0.45\textwidth}
        \centering
        \includegraphics[width=\linewidth]{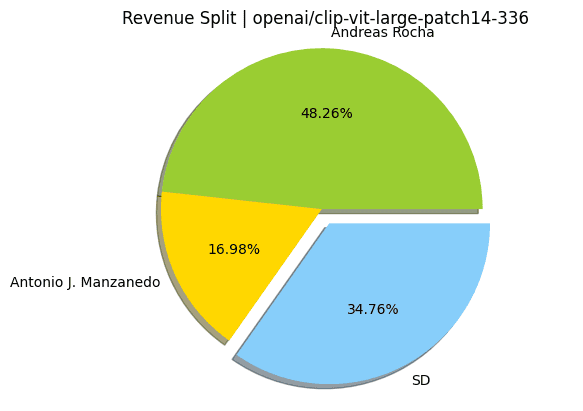}
        \caption{Shapley Values derived from embeddings generated by the CLIP}
        \label{fig23:image1}
    \end{minipage}
\end{figure}

The calculations depicted in the graphs (see Figure \ref{fig20:image1}-\ref{fig21:image1}) validate our visual hypothesis. After repositioning to the primary spot, the artist who previously claimed a smaller portion of the reward was able to amplify their contribution, thereby securing a larger share of the reward (see Figure \ref{fig22:image1}-\ref{fig23:image1}. This confirms the importance of the order in which the artist's name is introduced in the prompt.

\subsubsection{Dreambooth}

Our methodology is not only applicable to classical models but also to modified ones. For instance, we can learn the style of any desired artist using the Dreambooth technique.

Through this method, we can not only train the model to reproduce a specific style but also blend various styles in an arbitrary manner. This is achievable through the weighted summation of the weights of fine-tuned models. To facilitate this, we intentionally introduce an identical trigger-word for different styles during Dreambooth training, eliminating concerns about the word order in the prompt. Additionally, we now have verifiable knowledge about the images on which the model was fine-tuned and possess full access to them. This renders the outcomes of our experiment more transparent and allows us to identify the most pertinent images that exerted the most significant influence on the final generation. Here what we do next:
\begin{enumerate}[itemsep=0pt,topsep=5pt,partopsep=0pt]
    \item Process each original artwork created by artists to extract their respective embeddings.
    \item Compute the cosine similarity between the embeddings of generated images and those of the original artworks.
    \item Select the top9 images with the highest similarity or, alternatively, truncate the list when the drop in cosine similarity becomes too steep.
    \item Distribute the reward among the top \( n \) (where \( n \leq 9 \)) similar images, allocating the reward for each image based on its individual similarity score.
\end{enumerate}

\begin{figure}[h]
    \centering

    % First row of 4 images with "a)"
    \begin{minipage}{\textwidth}
        \centering
        \includegraphics[width=0.24\textwidth]{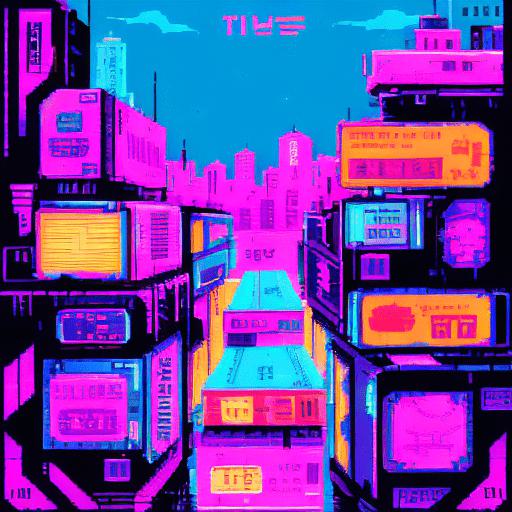}
        \includegraphics[width=0.24\textwidth]{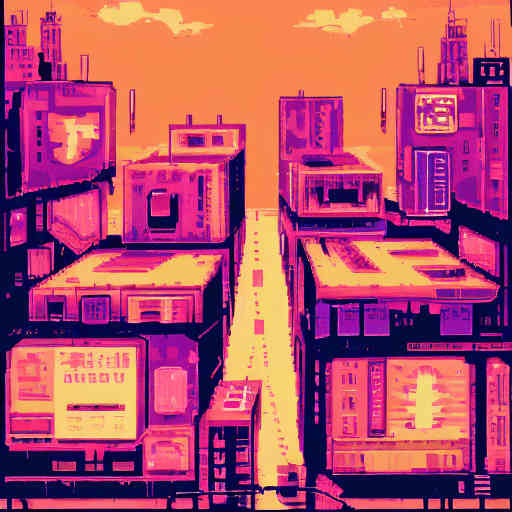}
        \includegraphics[width=0.24\textwidth]{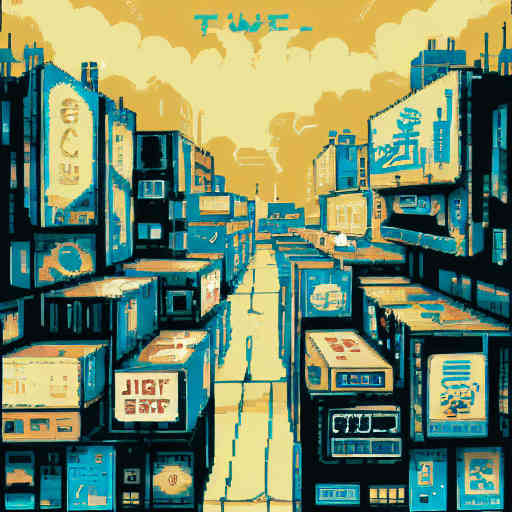}
        \includegraphics[width=0.24\textwidth]{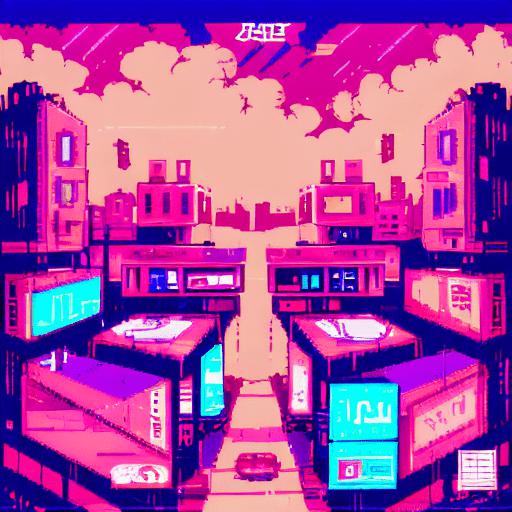}
        \subcaption{image produced by Dreambooth of \textit{Pixel Jeff}}
        \label{fig24:row_a}
    \end{minipage}

    % Second row of 4 images with "b)"
    \begin{minipage}{\textwidth}
        \centering
        \includegraphics[width=0.24\textwidth]{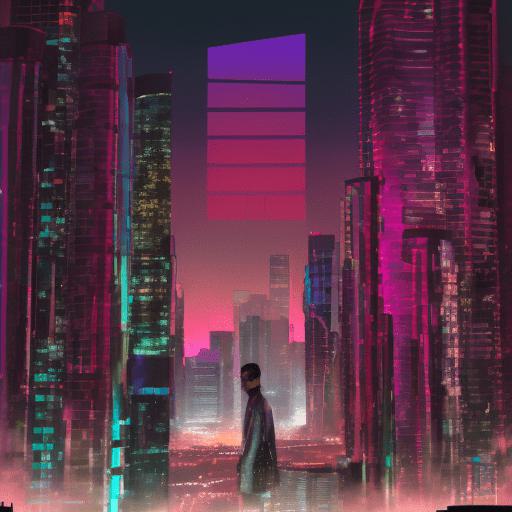}
        \includegraphics[width=0.24\textwidth]{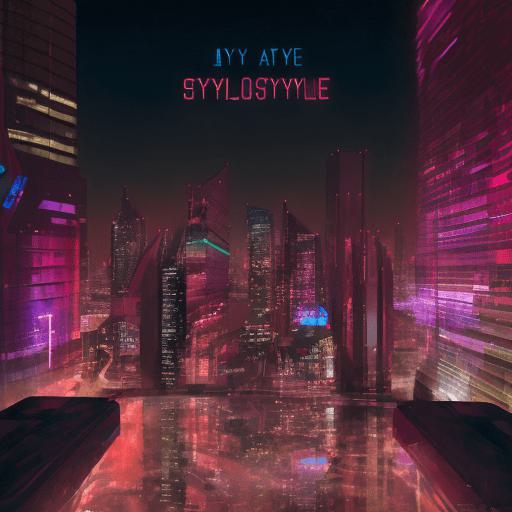}
        \includegraphics[width=0.24\textwidth]{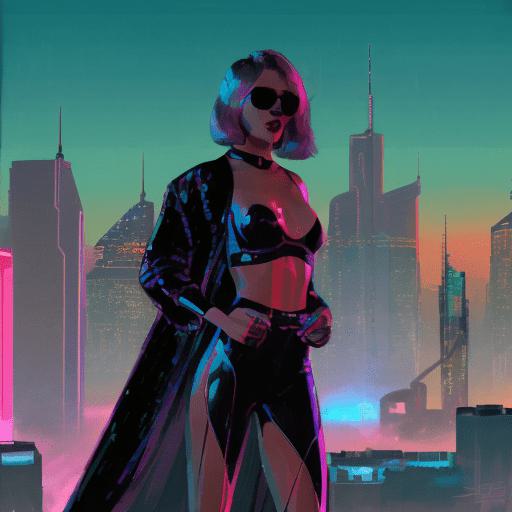}
        \includegraphics[width=0.24\textwidth]{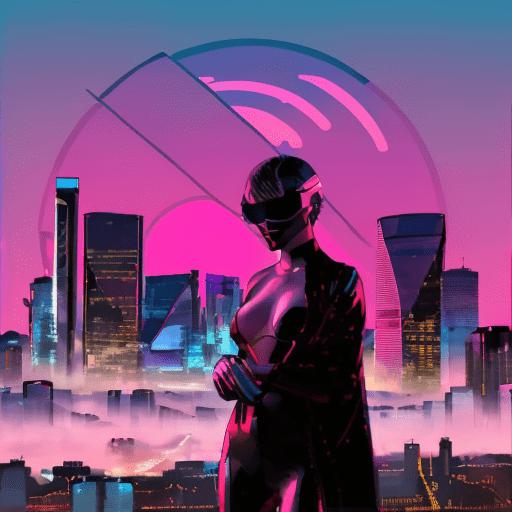}
        \subcaption{image produced by Dreambooth of \textit{Tony Skeor}}
        \label{fig24:row_b}
    \end{minipage}

    % Third row of 3 images with "c)"
    \begin{minipage}{\textwidth}
        \centering
        \includegraphics[width=0.32\textwidth]{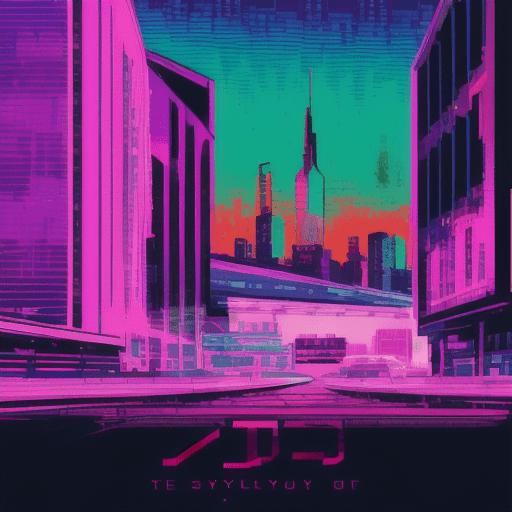}
        \includegraphics[width=0.32\textwidth]{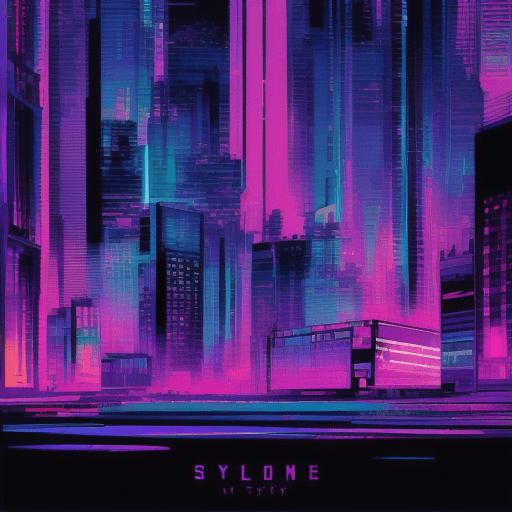}
        \includegraphics[width=0.32\textwidth]{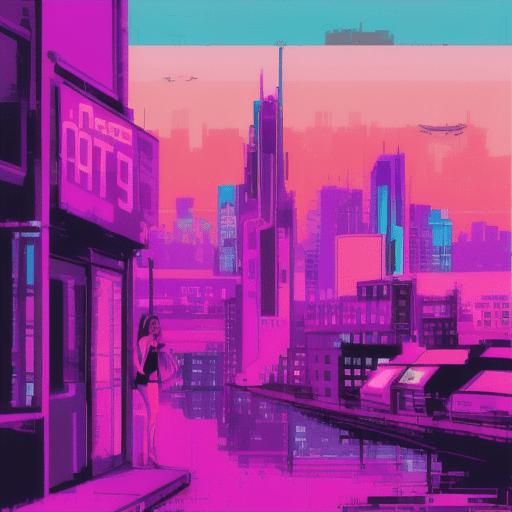}
        \subcaption{mix of \textit{Pixel Jeff} and \textit{Tony Skeor} models}
        \label{fig24:row_c}
    \end{minipage}

    % Fourth row of 3 images with "d)"
    \begin{minipage}{\textwidth}
        \centering
        \includegraphics[width=0.32\textwidth]{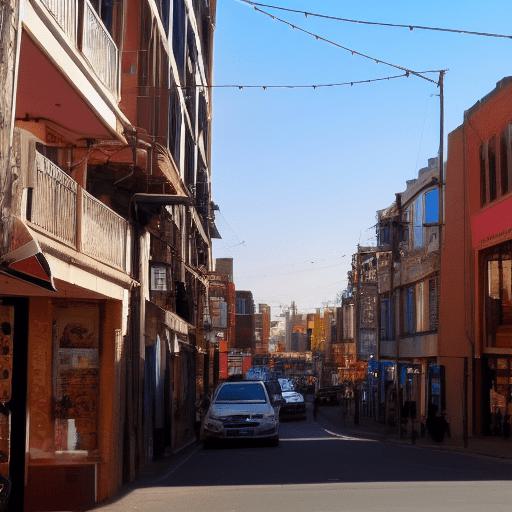}
        \includegraphics[width=0.32\textwidth]{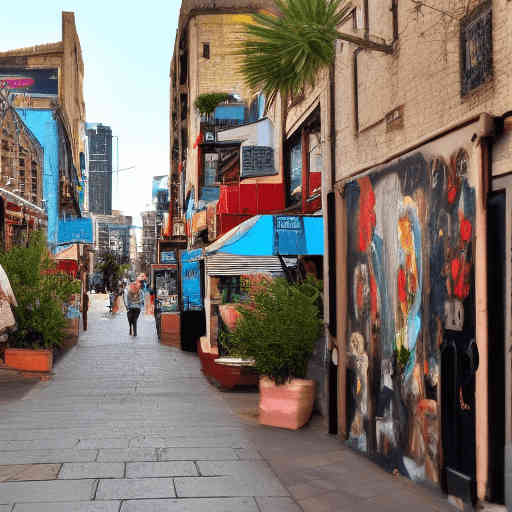}
        \includegraphics[width=0.32\textwidth]{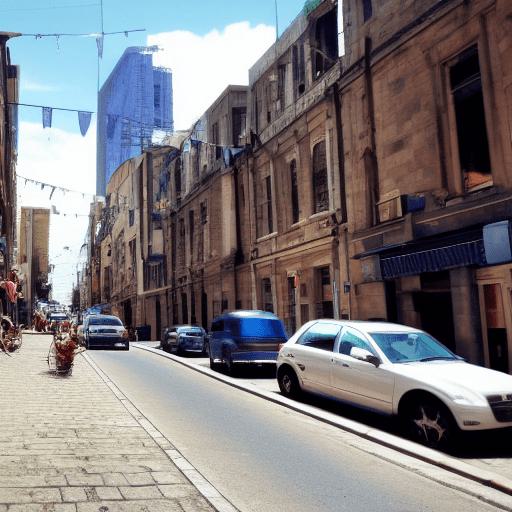}
        \subcaption{outputs from the SD1.5 model without any specific style applied.}
        \label{fig24:row_d}
    \end{minipage}

    \caption{Examples of generated outputs: a) \textbf{A+SD}, b) \textbf{B+SD}, c) \textbf{A+B+SD}, and d) \textbf{SD}.}
    \label{fig24:all_images}
\end{figure}

\begin{figure}[h]
    \centering
    \includegraphics[width=0.7\textwidth]{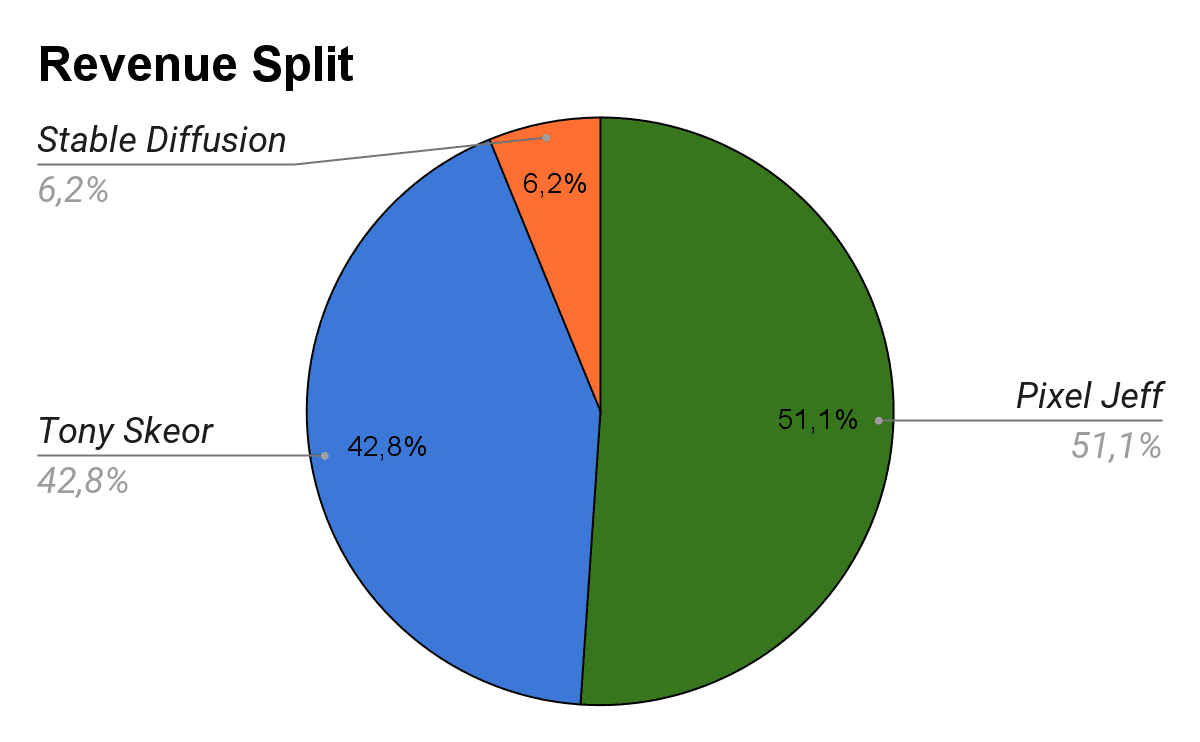}
    \caption{Distribution of revenue for the mix of \textit{Tony Skeor} and \textit{Pixel Jeff} styles.}
    \label{fig25:image1}
\end{figure}

Below is example (see Figure \ref{fig24:all_images}) of generation produced by mix between two dreambooth model, trained on Pixel Jeff(A) and  Tony Skeor(B). In order to achieve mix of two models we merge their weights as \(W = 0.5 * A + 0.5 * B\). After that we calculate Shapley Values based on embedding similarity score and achieve following results (see Figure \ref{fig25:image1}):
\begin{itemize}[itemsep=0pt,topsep=5pt,partopsep=0pt]
    \item Pixel Jeff - 51.1\%
    \item Tony Skeor - 42.8\%
    \item Stable Diffsuion - 6.2\%
\end{itemize}

\begin{figure}[h]
    \centering
    
    % First image
    \begin{subfigure}{0.48\textwidth}
        \centering
        \includegraphics[width=\linewidth]{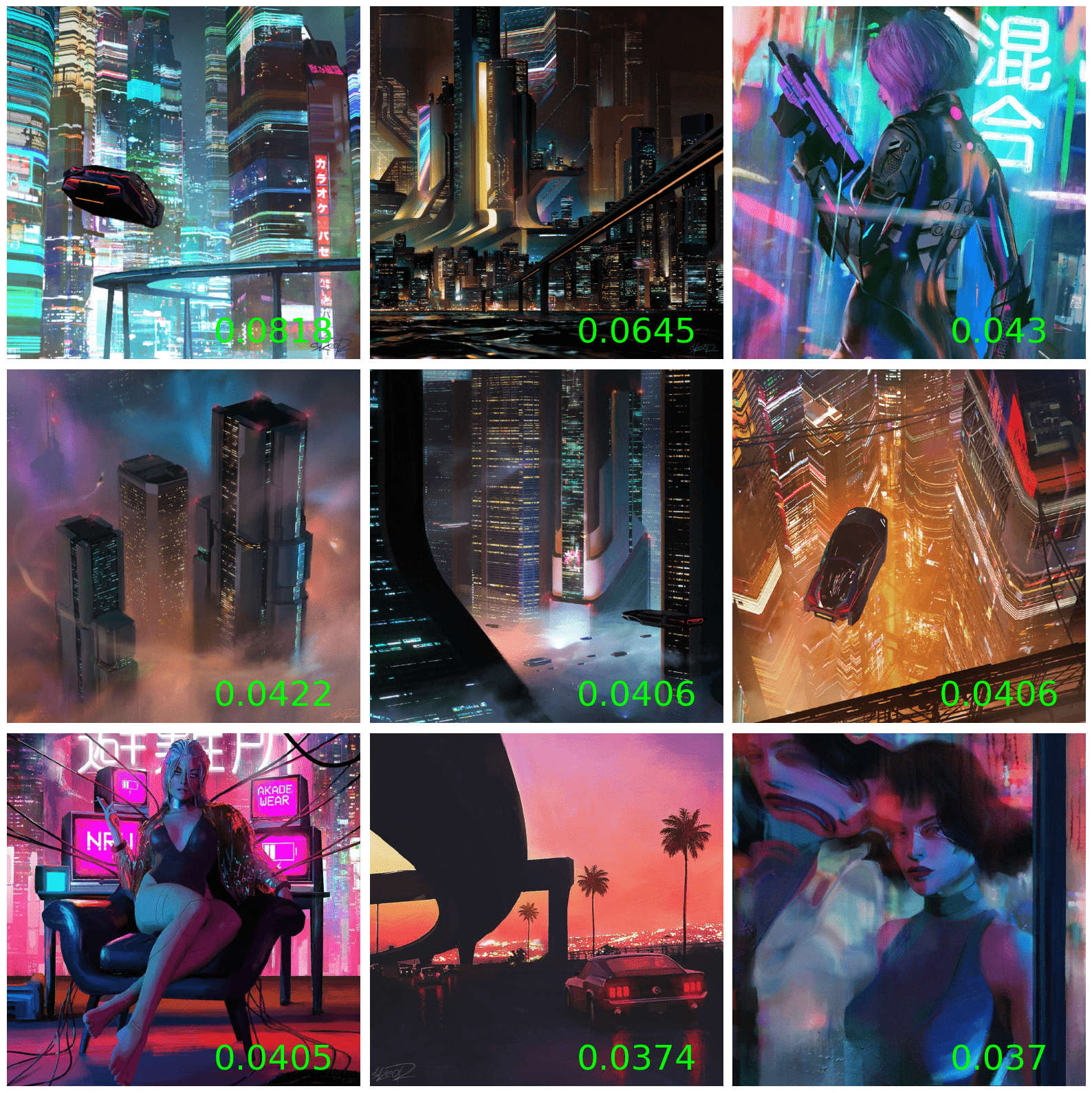}
        \caption{original artworks by \textit{Tony Skeor}}
        \label{fig26:first_image}
    \end{subfigure}
    % \hfill
    % Second image
    \begin{subfigure}{0.48\textwidth}
        \centering
        \includegraphics[width=\linewidth]{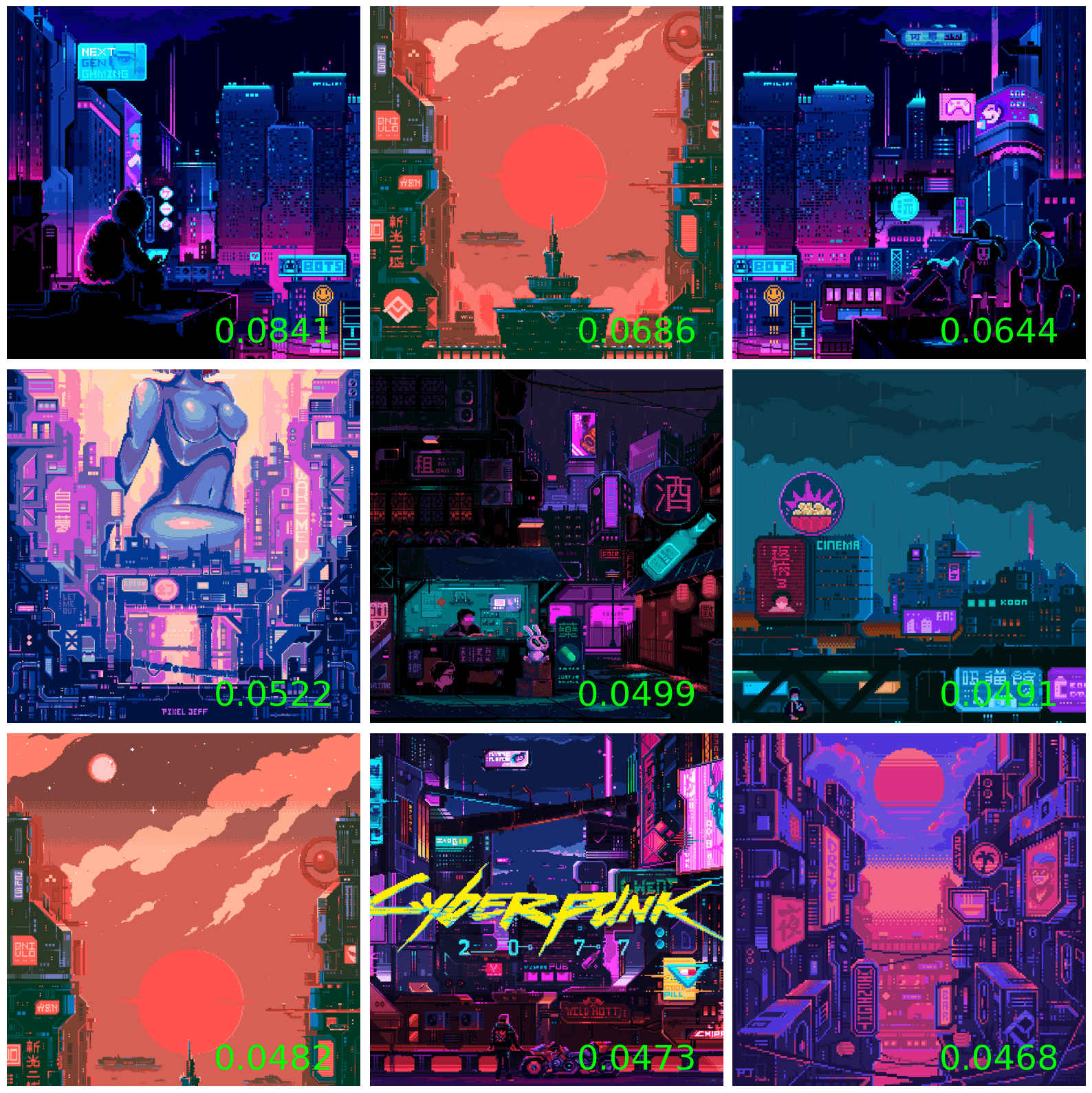}
        \caption{original artworks by \textit{Pixel Jeff}}
        \label{fig26:second_image}
    \end{subfigure}

    \caption{top closest images of real artists, scored by cosine-similarity}
    \label{fig26:all_images}    

\end{figure}

In this experiment, we see that among the works of real artists, there are examples very similar to the final generations (Figure \ref{fig26:all_images}). While the pure SD's generations based on identical prompt turn out to be significantly dissimilar to the targets. It is also apparent that the result incorporates features from both the first (A) and the second artist (B), but still turned out to be a bit closer to the works of the first one (since the model based on the second one was more inclined to generate humans). Therefore, the majority of the reward is divided among the artists, with a smaller portion going to the generative model.

In the subsequent experiment, we employed the Image Mixer \cite{pinkney2022imagemixer} model for the equal blending of two real images. 
Image Mixer is a deep learning model that allows for the creation of new images by combining the concepts, styles, and compositions from multiple images (and text prompts too). It was developed by Justin Pinkney at Lambda Labs. During training, up to 5 crops of the training images are taken and CLIP embeddings are extracted. These embeddings are concatenated and used as the conditioning for the model. At inference time, CLIP embeddings from multiple images can be used to generate images which are influenced by multiple inputs.

\begin{figure}[h]
    \centering
    \begin{subfigure}{0.24\textwidth}
        \centering
        \includegraphics[width=\linewidth]{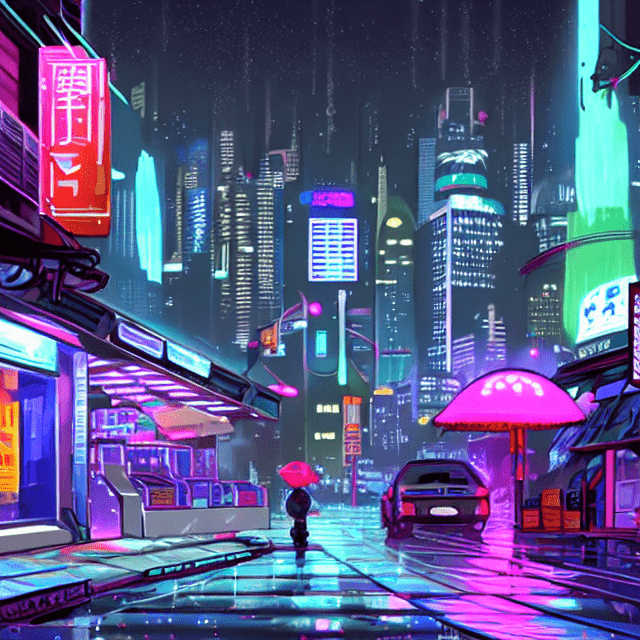}
    \end{subfigure}
    \begin{subfigure}{0.24\textwidth}
        \centering
        \includegraphics[width=\linewidth]{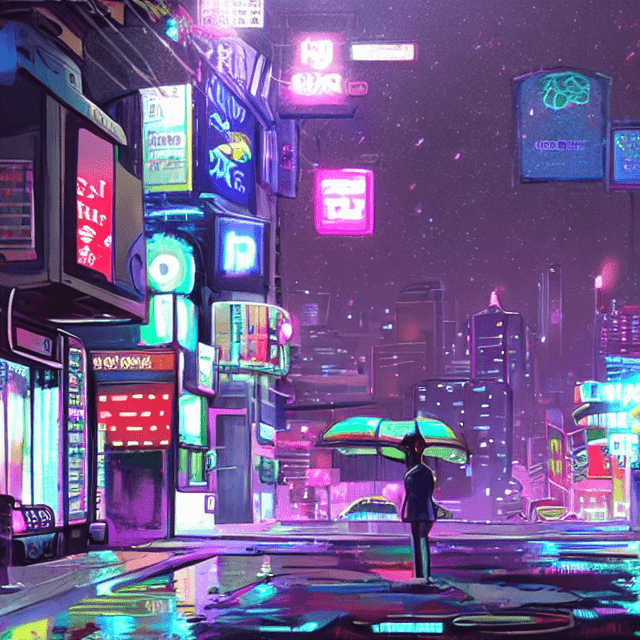}
    \end{subfigure}
    \begin{subfigure}{0.24\textwidth}
        \centering
        \includegraphics[width=\linewidth]{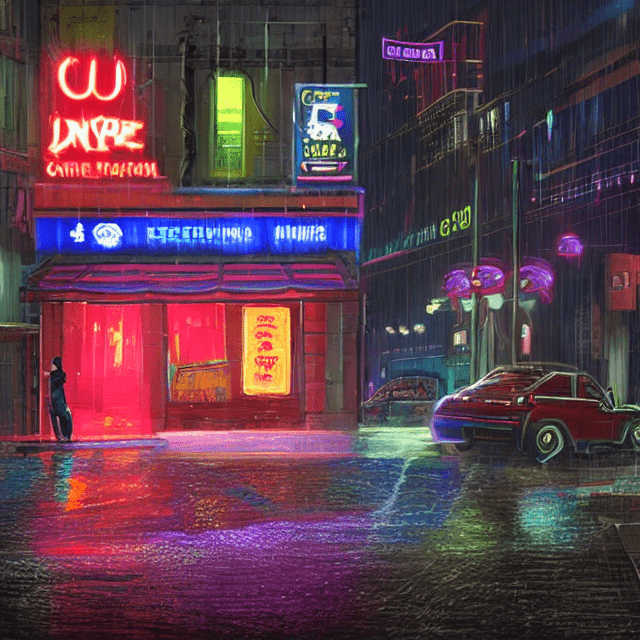}
    \end{subfigure}
    \begin{subfigure}{0.24\textwidth}
        \centering
        \includegraphics[width=\linewidth]{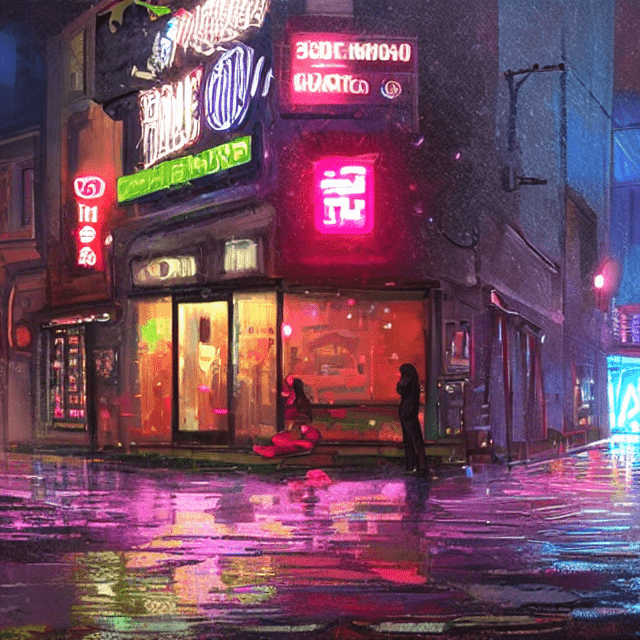}
    \end{subfigure}
    \caption{mix of \textit{Tony Skeor} and \textit{Pixel Jeff} produced by ImageMixer}
    \label{fig27:all_images}    
\end{figure}

\begin{figure}[h]
    \centering
    \includegraphics[width=0.7\textwidth]{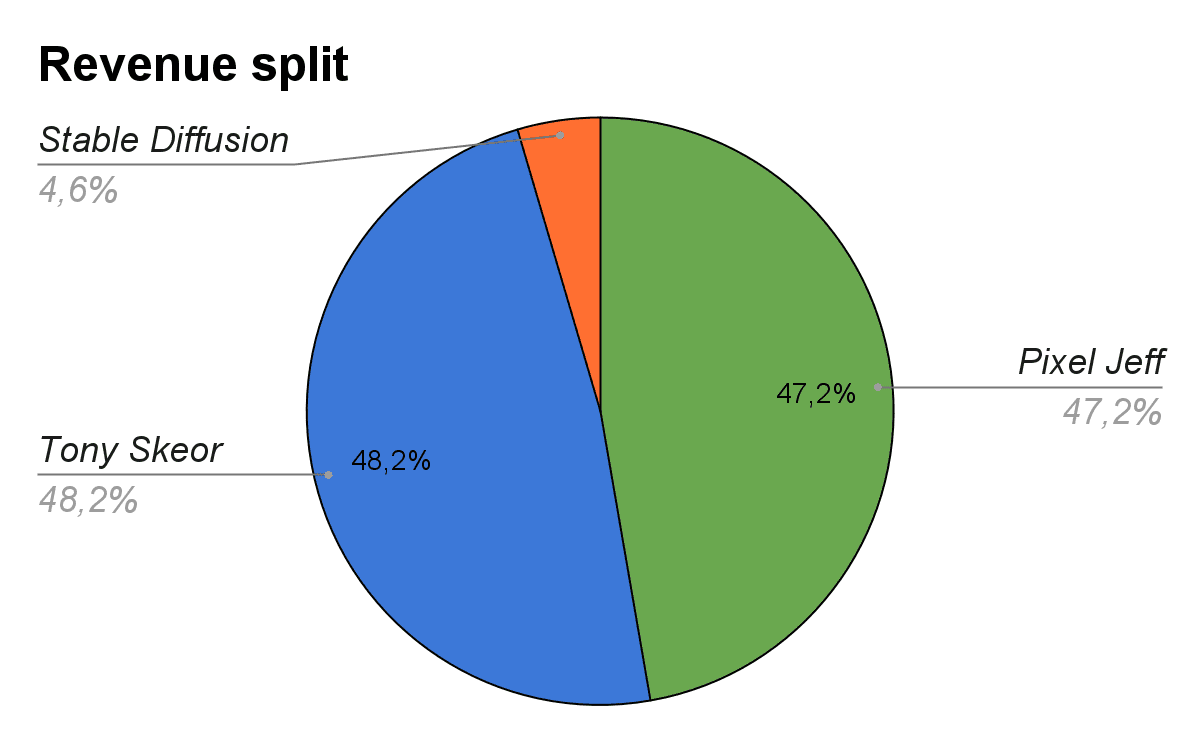}
    \caption{Contribution distribution among images generated by ImageMixer based on the works of \textit{Pixel Jeff} and \textit{Tony Skeor}}
    \label{fig28:image1}
\end{figure}

\begin{figure}[h]
    \centering
    \begin{subfigure}{0.48\textwidth}
        \centering
        \includegraphics[width=\linewidth]{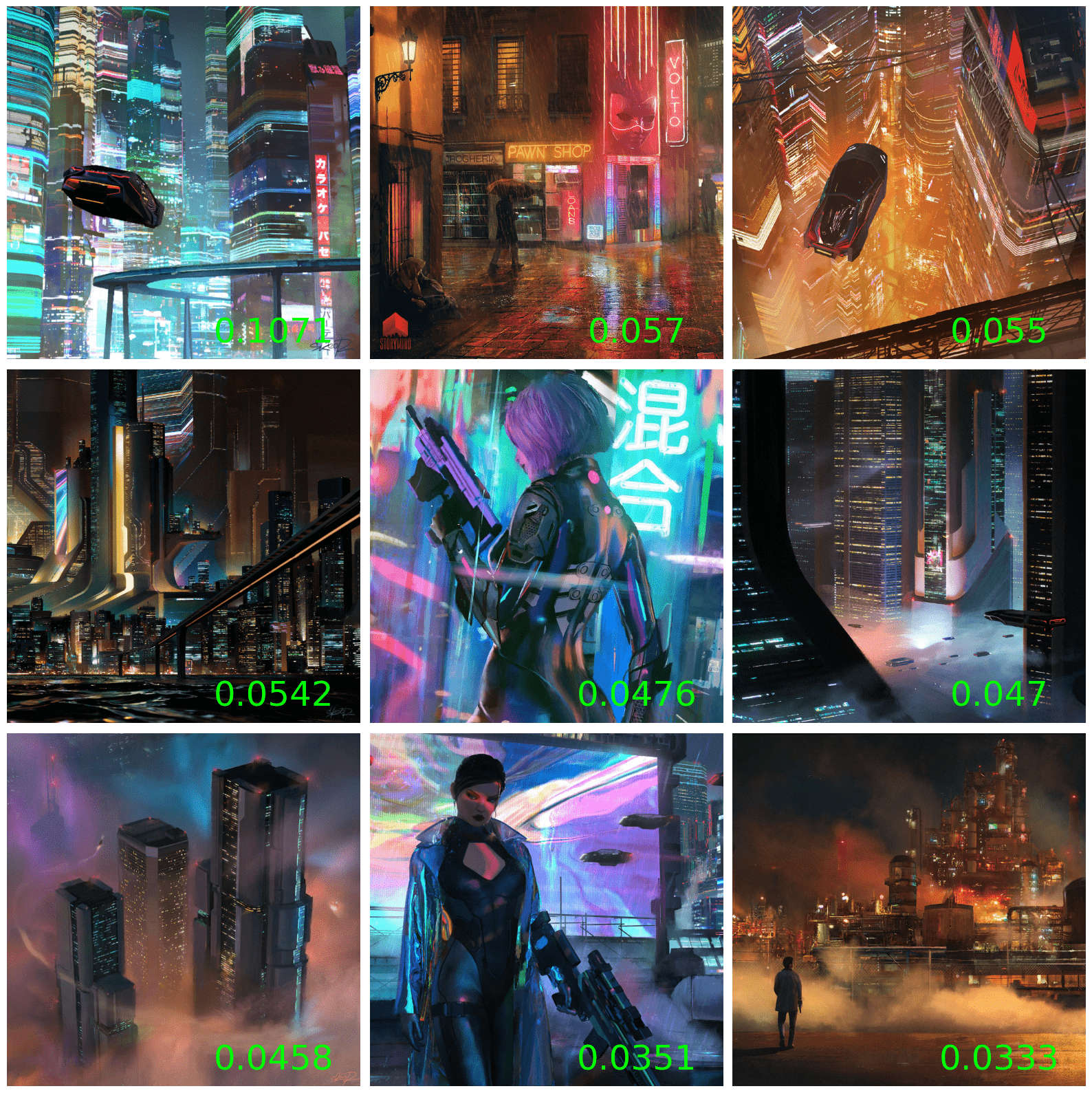}
        \caption{original artworks by \textit{Tony Skeor}}
        \label{fig29:first_image}
    \end{subfigure}
    \begin{subfigure}{0.48\textwidth}
        \centering
        \includegraphics[width=\linewidth]{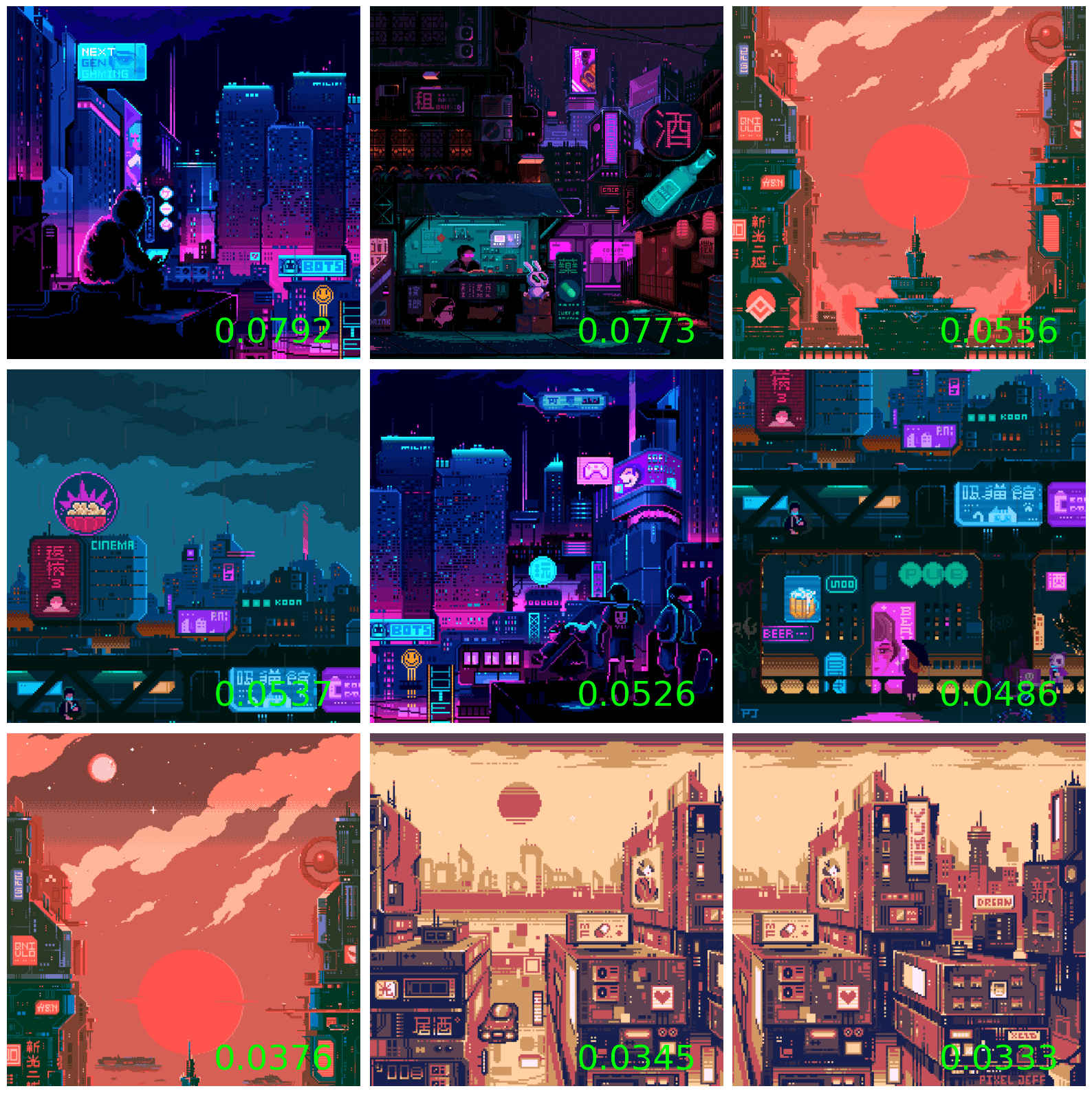}
        \caption{original artworks by \textit{Pixel Jeff}}
        \label{fig29:second_image}
    \end{subfigure}
    \caption{top closest images of real artists, scored by cosine-similarity}
    \label{fig29:all_images}    
\end{figure}

We used this mix (see Figure \ref{fig27:all_images}), provided by ImageMixer as the \texttt{A+B+SD} coalition \textit{(Pixel Jeff + Tony Skeor + SD)}, with the aim of achieving a maximally equivalent contribution for artist A and B. In modifying the previous experimental approach, we attained a more equitable distribution of rewards, as evidenced by the Shapley Values (see Figure~\ref{fig28:image1}). Significantly, our approach assigned higher scores to the images utilized for the image blending, as illustrated in Figure~\ref{fig29:all_images}.

\begin{figure}[h]
    \centering
    \begin{minipage}{\textwidth}
        \centering
        \includegraphics[width=0.24\textwidth]{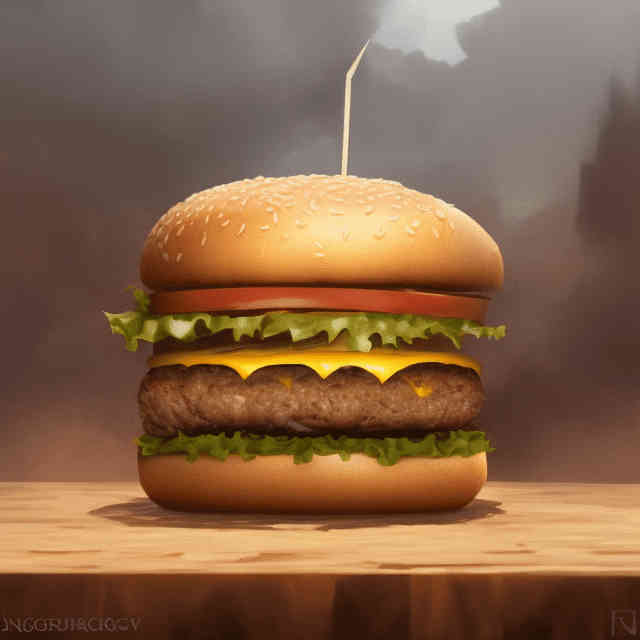}
        \includegraphics[width=0.24\textwidth]{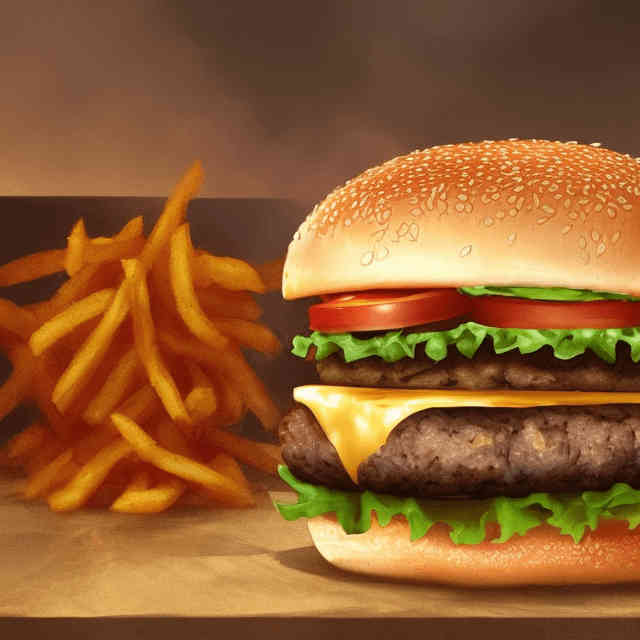}
        \includegraphics[width=0.24\textwidth]{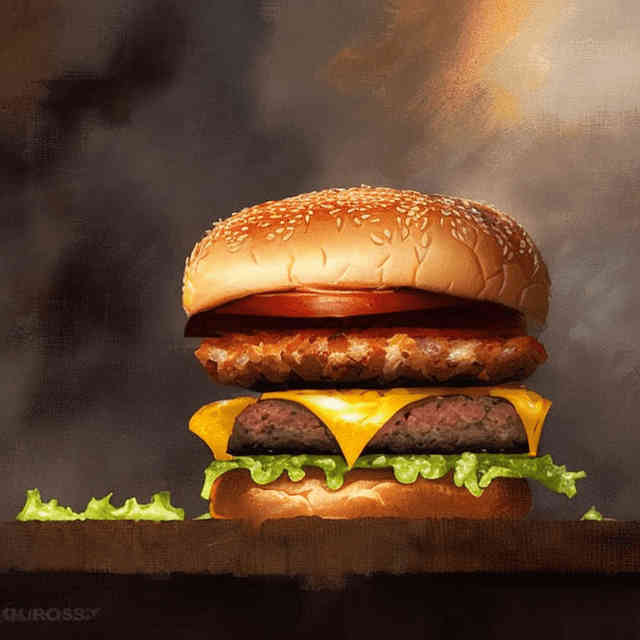}
        \includegraphics[width=0.24\textwidth]{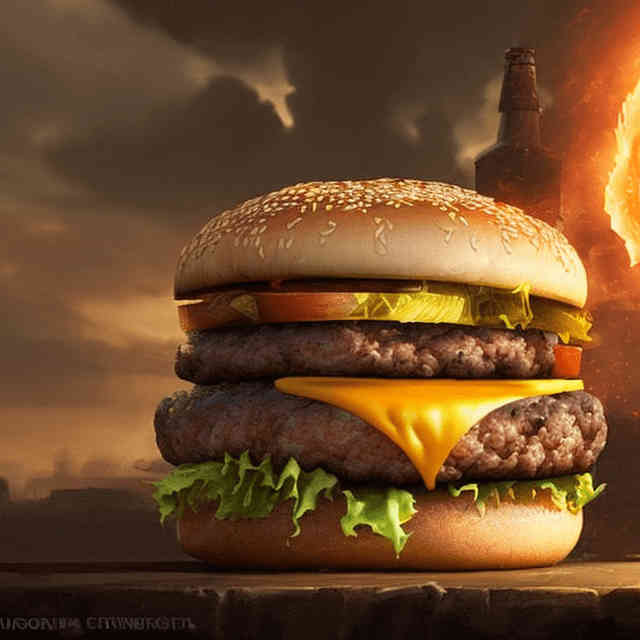}
        \subcaption{image produced by Dreambooth of \textit{Greg Rutkowski}}
        \label{fig30:row_a}
    \end{minipage}
    \begin{minipage}{\textwidth}
        \centering
        \includegraphics[width=0.24\textwidth]{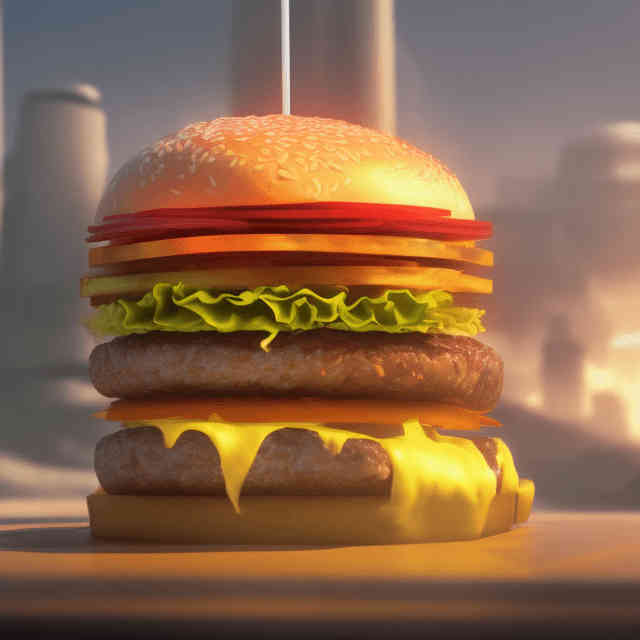}
        \includegraphics[width=0.24\textwidth]{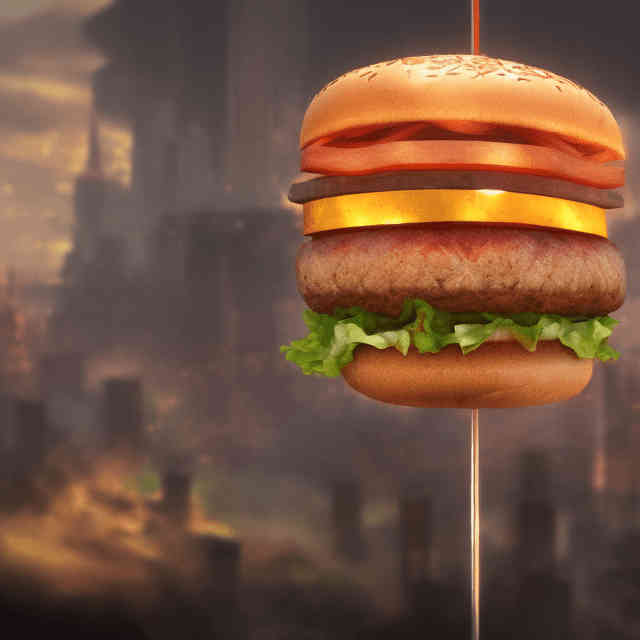}
        \includegraphics[width=0.24\textwidth]{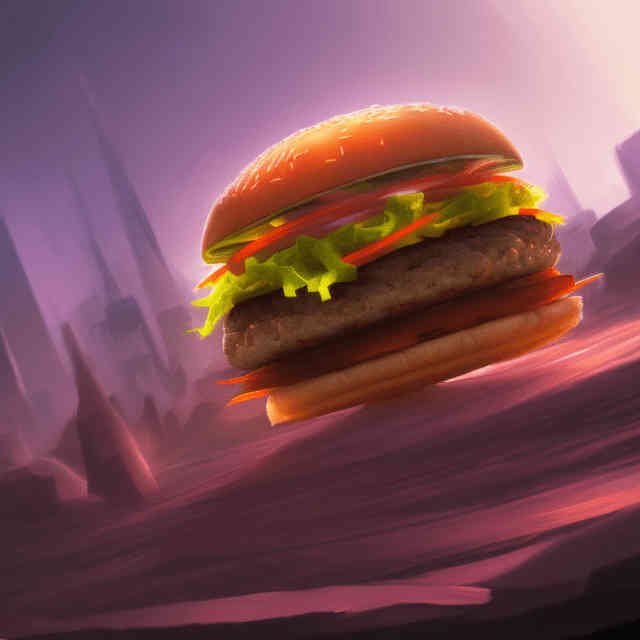}
        \includegraphics[width=0.24\textwidth]{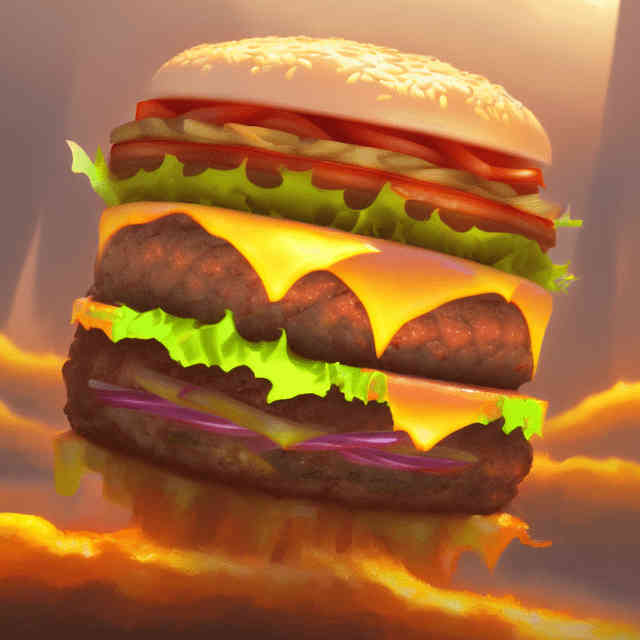}
        \subcaption{image produced by Dreambooth of \textit{Leon Tukker}}
        \label{fig30:row_b}
    \end{minipage}
    \begin{minipage}{\textwidth}
        \centering
        \includegraphics[width=0.32\textwidth]{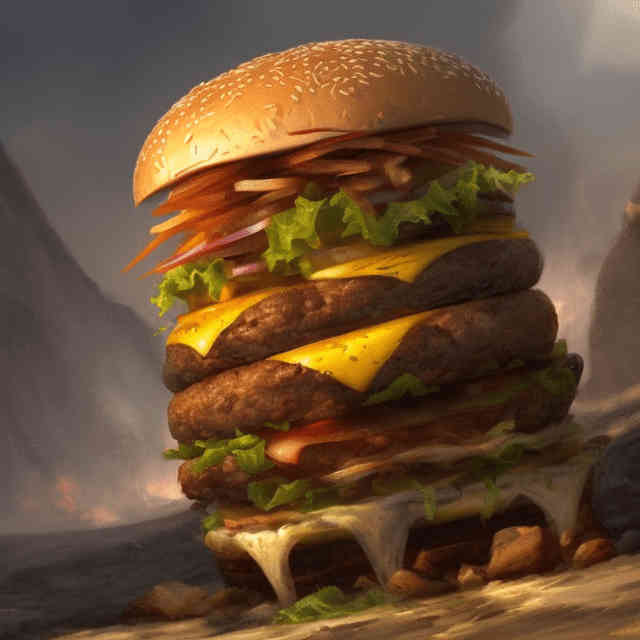}
        \includegraphics[width=0.32\textwidth]{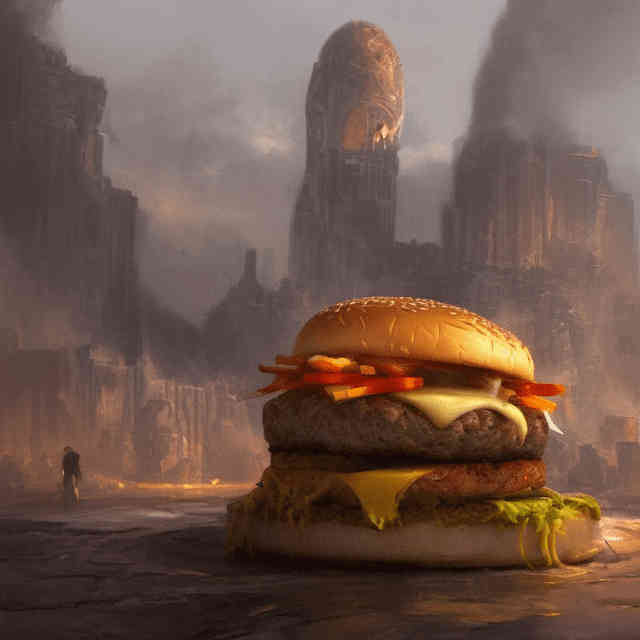}
        \includegraphics[width=0.32\textwidth]{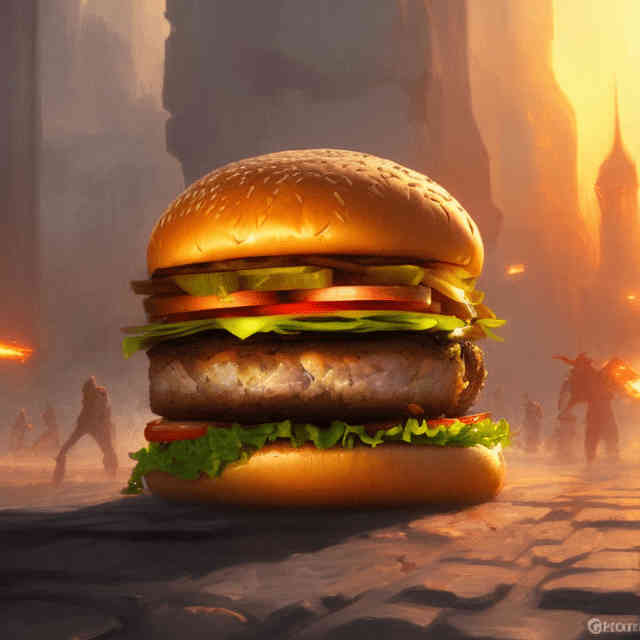}
        \subcaption{mix of \textit{Greg Rutkowski} and \textit{Leon Tukker} models}
        \label{fig30:row_c}
    \end{minipage}
    \begin{minipage}{\textwidth}
        \centering
        \includegraphics[width=0.32\textwidth]{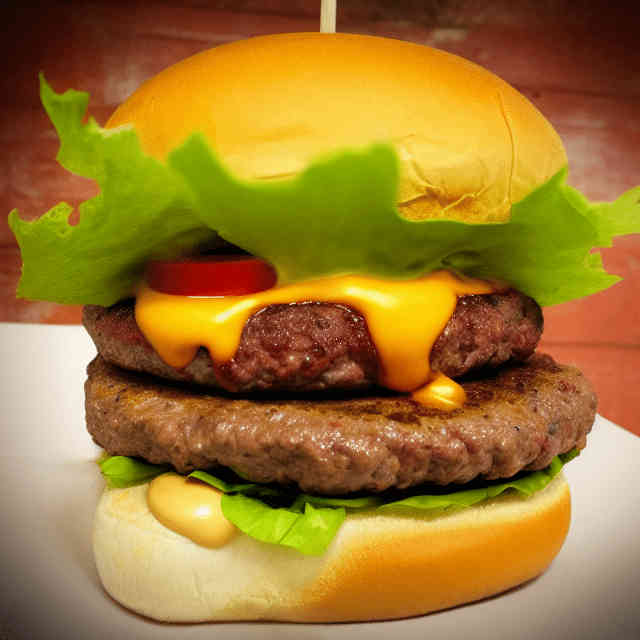}
        \includegraphics[width=0.32\textwidth]{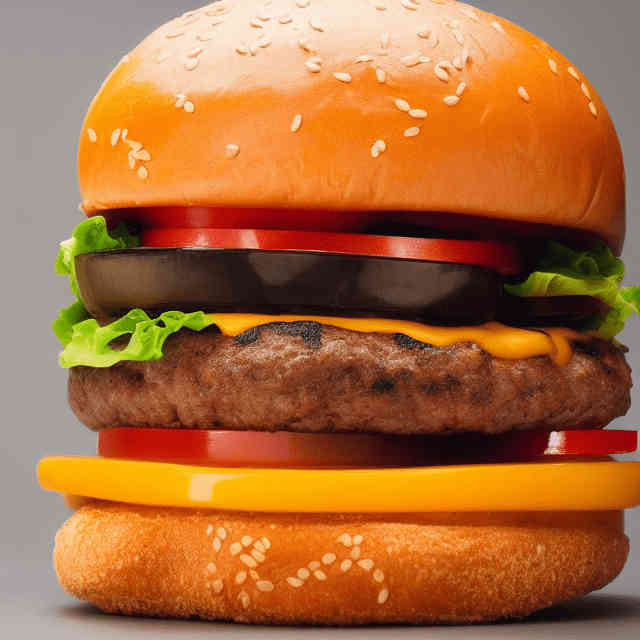}
        \includegraphics[width=0.32\textwidth]{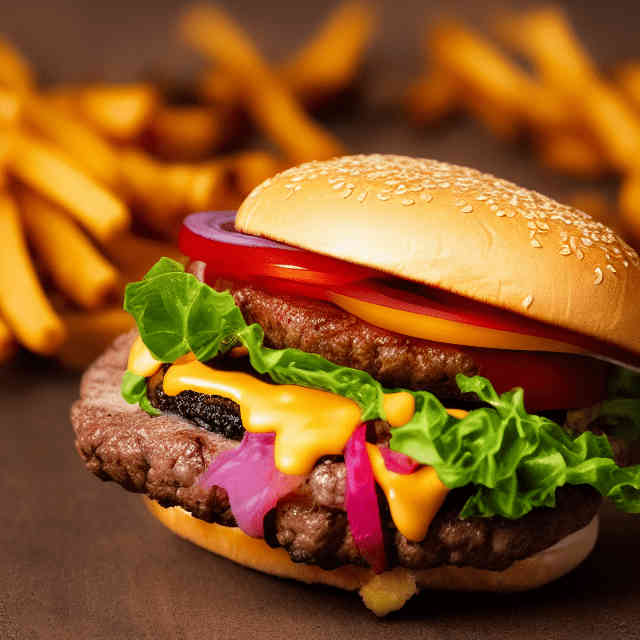}
        \subcaption{outputs from the SD1.5 model without any specific style applied.}
        \label{fig30:row_d}
    \end{minipage}
    \caption{Examples of generated outputs: a) \textbf{A+SD}, b) \textbf{B+SD}, c) \textbf{A+B+SD}, and d) \textbf{SD}.}
    \label{fig30:all_images}
\end{figure}

\begin{figure}[h]
    \centering
    \includegraphics[width=0.7\textwidth]{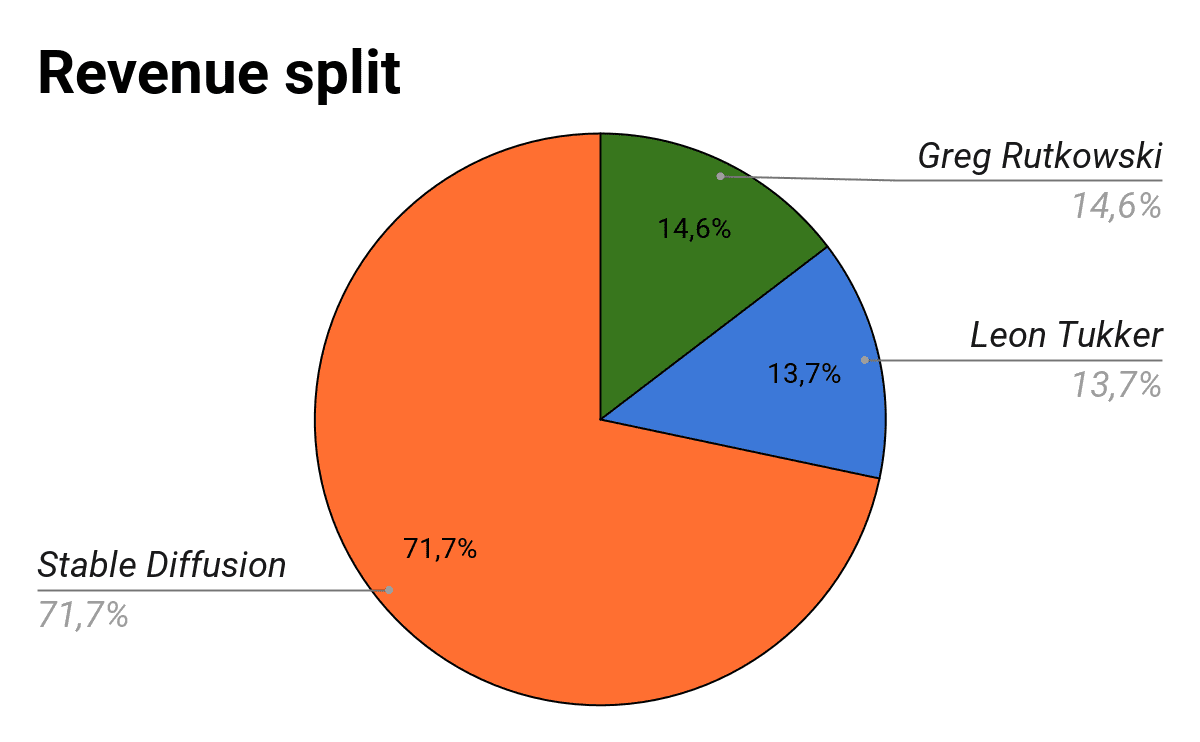}
    \caption{Distribution of revenue for the mix of \textit{Greg Rutkowski} and \textit{Leon Tukker} styles.}
    \label{fig31:image1}
\end{figure}

\begin{figure}[h]
    \centering
    \begin{subfigure}{0.48\textwidth}
        \centering
        \includegraphics[width=\linewidth]{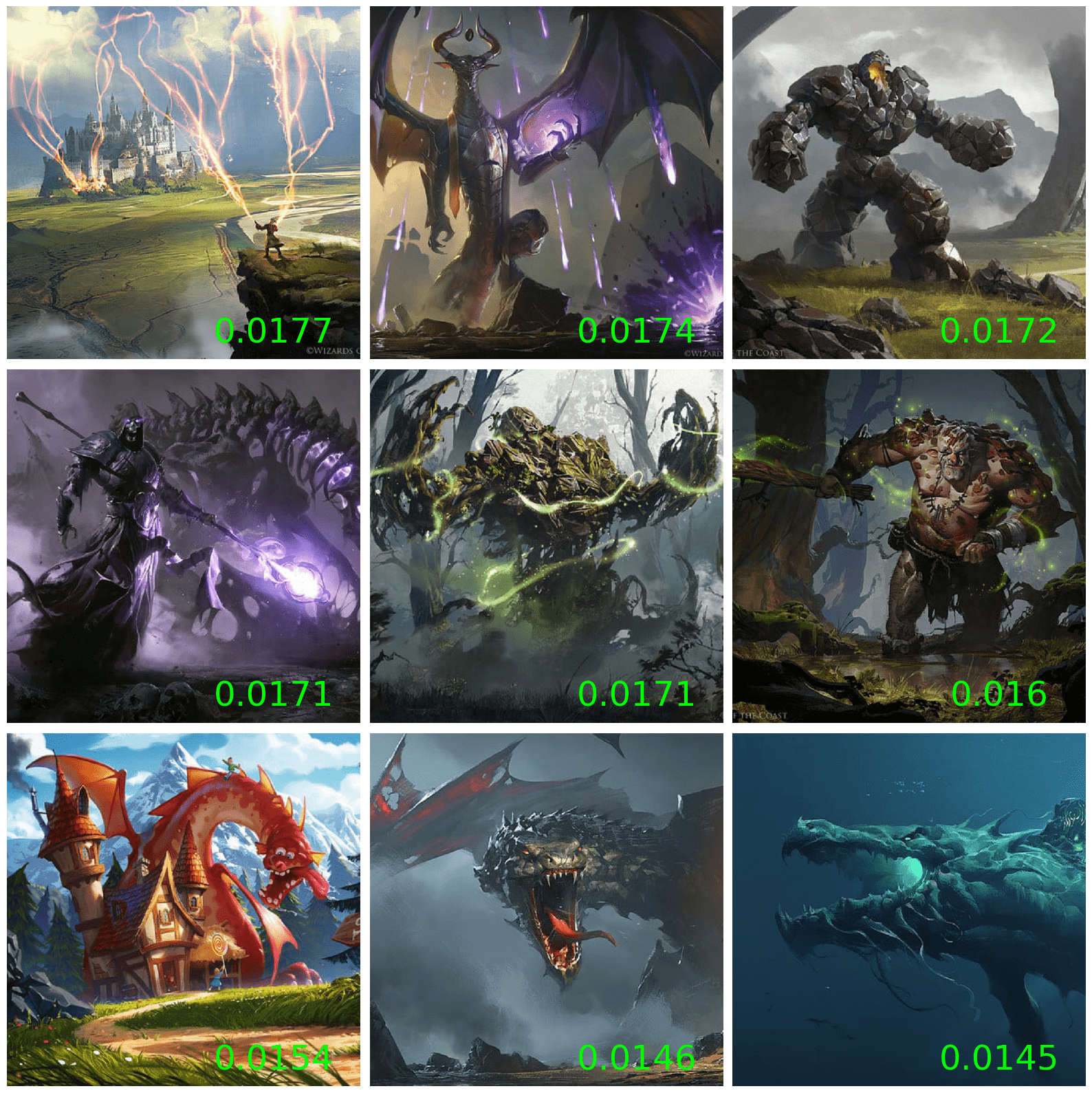}
        \caption{original artworks by \textit{Greg Rutkowski}}
        \label{fig32:first_image}
    \end{subfigure}
    \begin{subfigure}{0.48\textwidth}
        \centering
        \includegraphics[width=\linewidth]{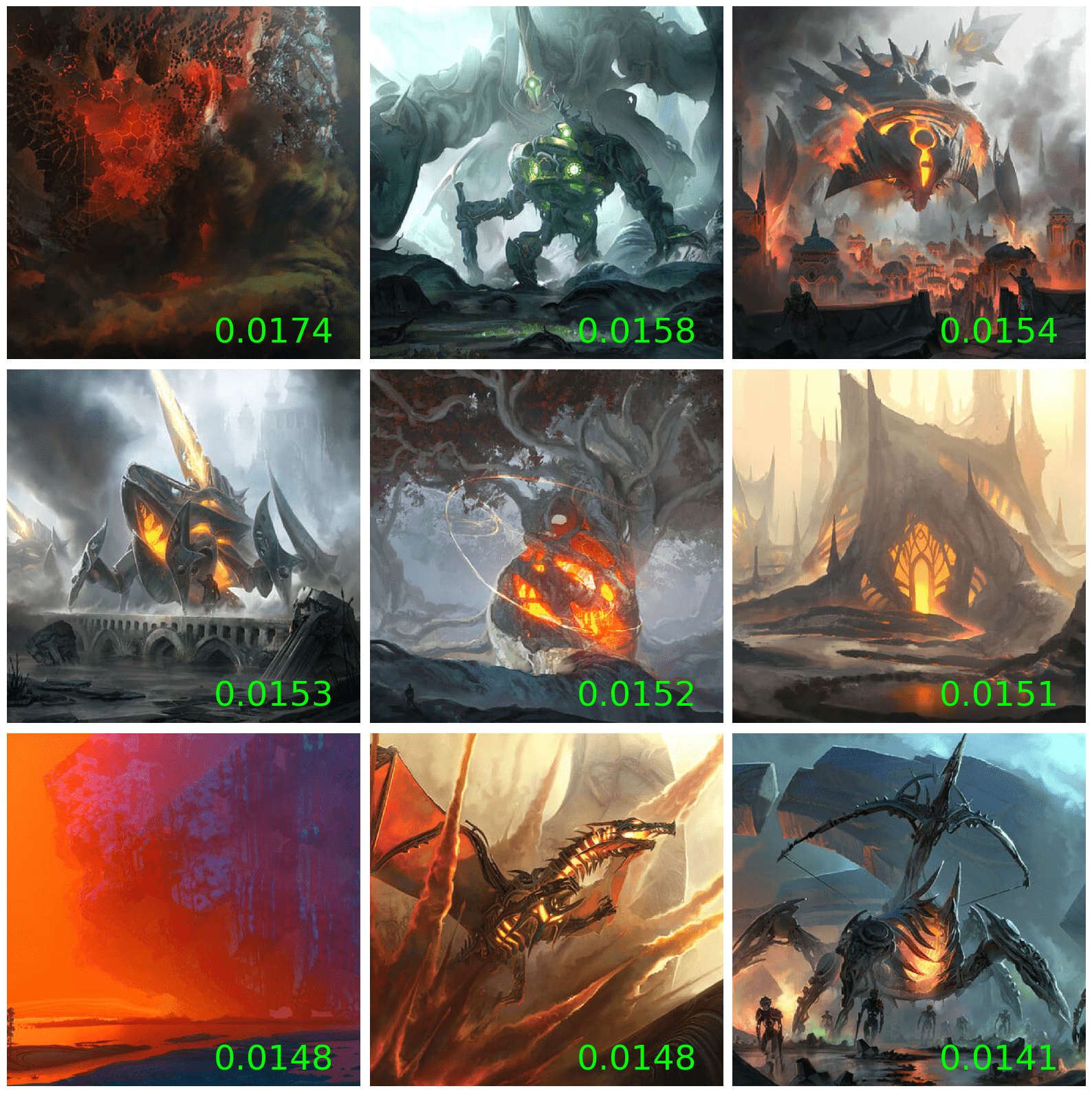}
        \caption{original artworks by \textit{Leon Tukker}}
        \label{fig32:second_image}
    \end{subfigure}
    \caption{top closest images of real artists, scored by cosine-similarity}
    \label{fig32:all_images}    
\end{figure}

In the following analysis, we will illustrate a scenario where the majority of the reward is allocated to Stable Diffusion due to its adeptness in seamlessly transferring style to a context not previously observed in the original artworks. To achieve this, we will endeavor to amalgamate the styles of artists “Greg Rutkowski” and “Tony Skeor” to generate “an image of a burger” (see Figure~\ref{fig30:all_images}). On the Figure~\ref{fig32:all_images} we can clearly see that in the works of the studied real artists, there is nothing resembling a "burger", hence the majority of the reward goes to the generative model (see Figure~\ref{fig31:image1}). However, we also observe that the images transitioned from realistic to drawn, and the background of the images significantly changed, indicating a contribution from the real artists. Therefore, they also receive a small portion of the reward.

\begin{figure}[h]
    \centering
    \begin{minipage}{\textwidth}
        \centering
        \includegraphics[width=0.24\textwidth]{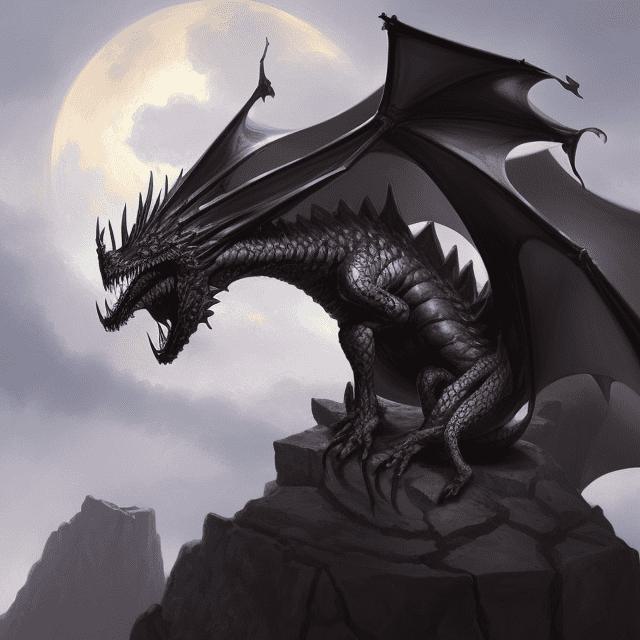}
        \includegraphics[width=0.24\textwidth]{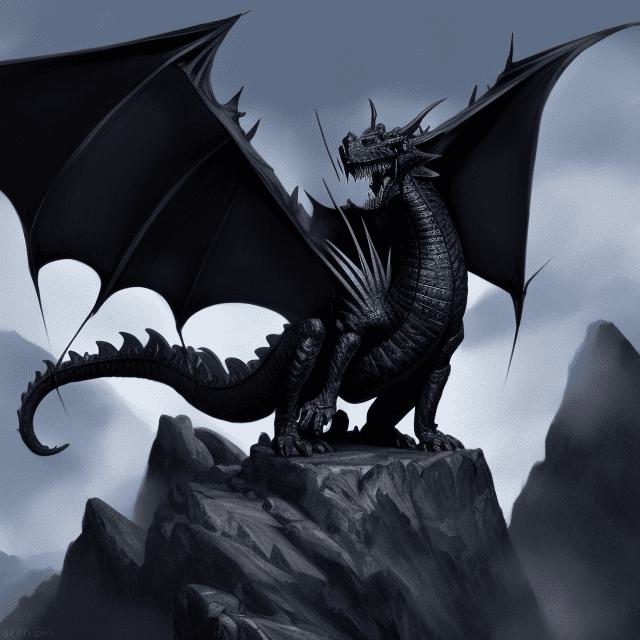}
        \includegraphics[width=0.24\textwidth]{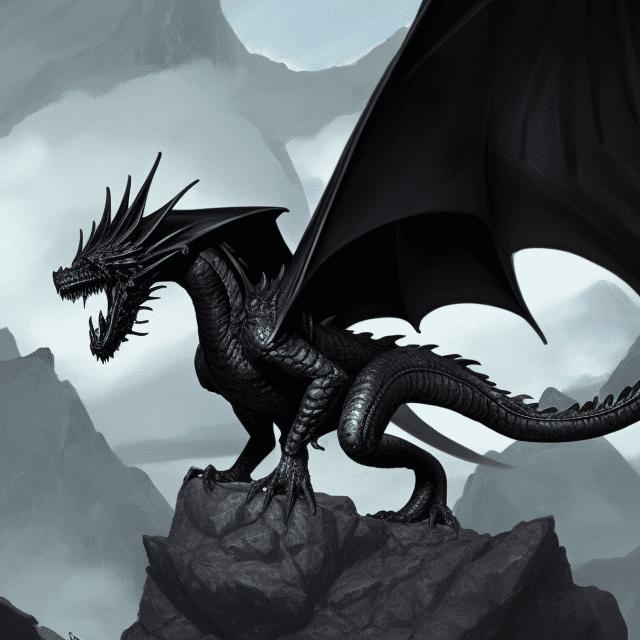}
        \includegraphics[width=0.24\textwidth]{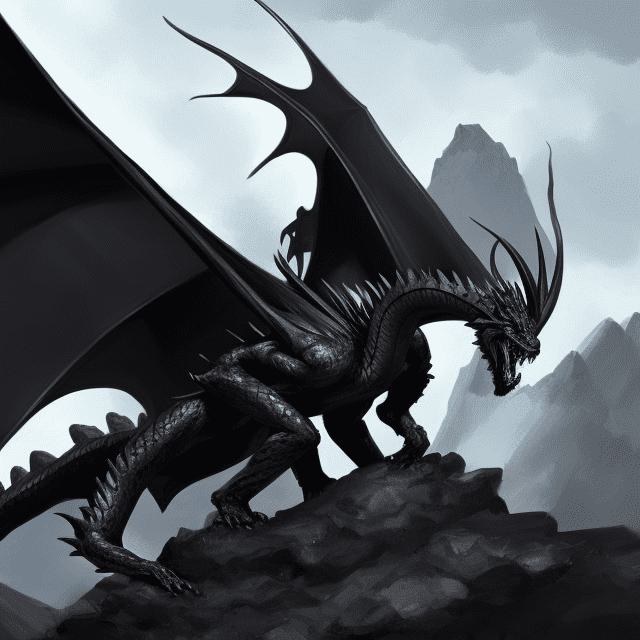}
        \subcaption{image produced by Dreambooth of \textit{Greg Rutkowski}}
        \label{fig33:row_a}
    \end{minipage}
    \begin{minipage}{\textwidth}
        \centering
        \includegraphics[width=0.24\textwidth]{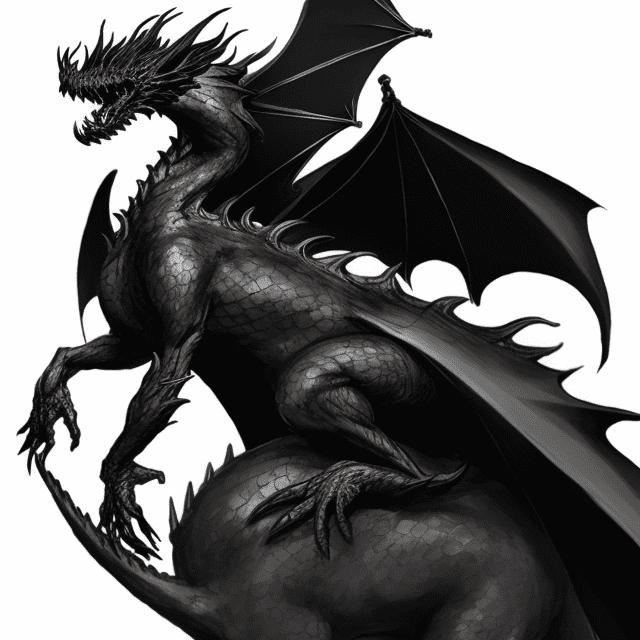}
        \includegraphics[width=0.24\textwidth]{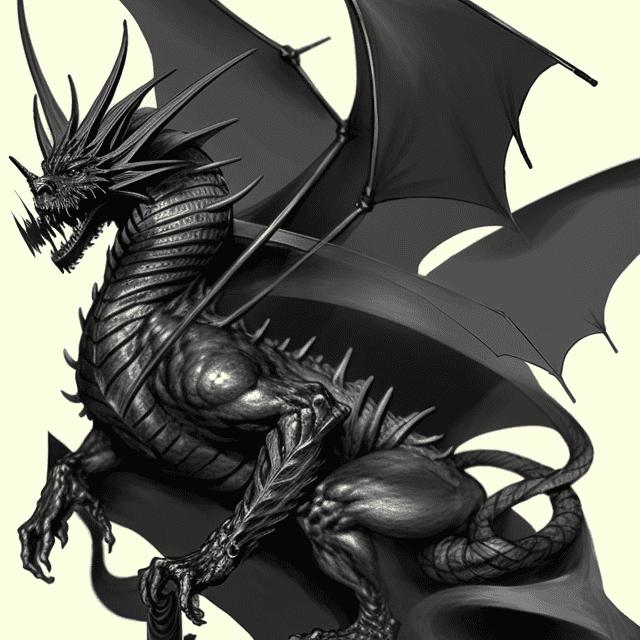}
        \includegraphics[width=0.24\textwidth]{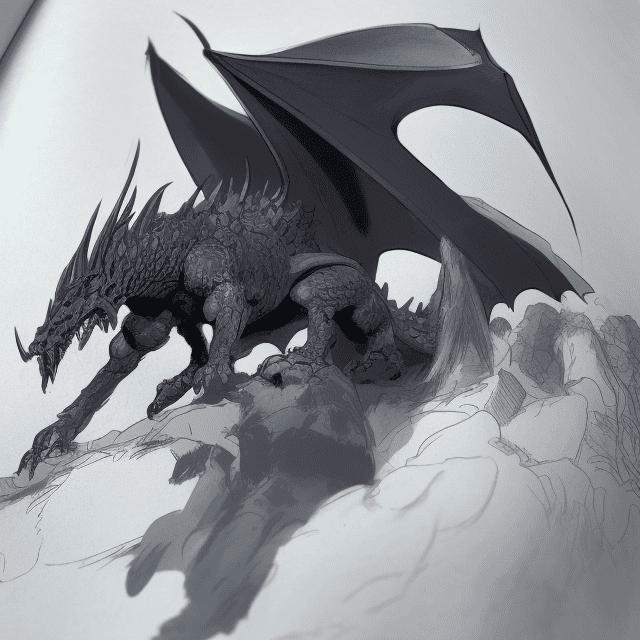}
        \includegraphics[width=0.24\textwidth]{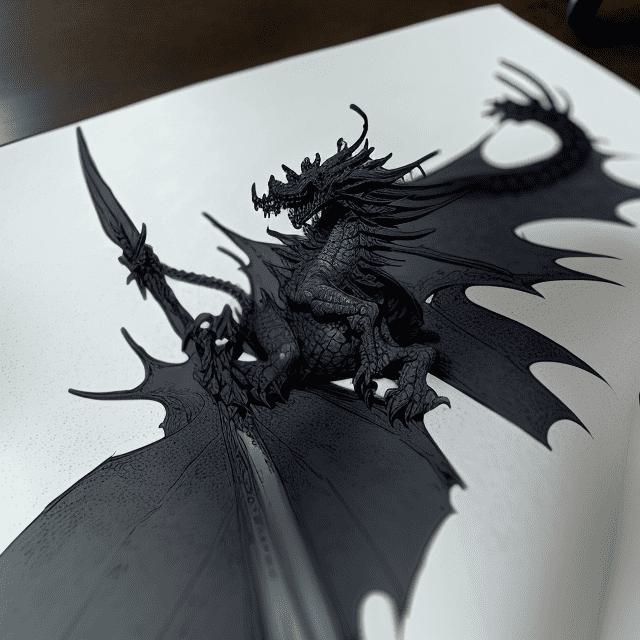}
        \subcaption{image produced by Dreambooth of \textit{Even Amundsen}}
        \label{fig33:row_b}
    \end{minipage}
    \begin{minipage}{\textwidth}
        \centering
        \includegraphics[width=0.24\textwidth]{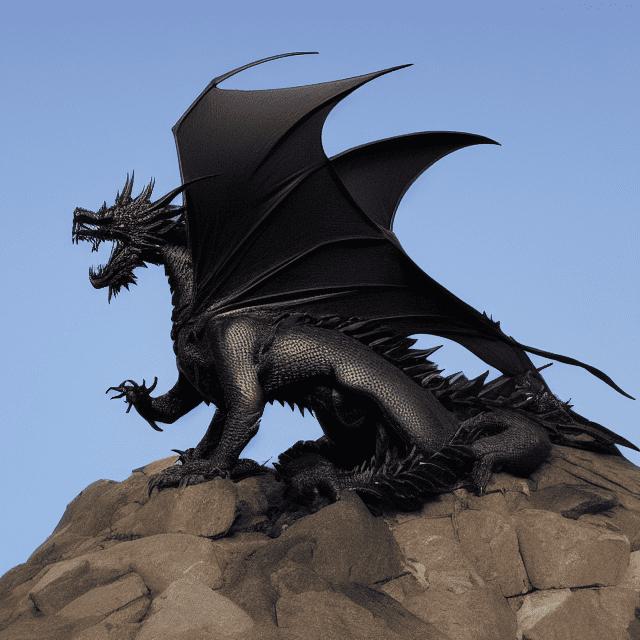}
        \includegraphics[width=0.24\textwidth]{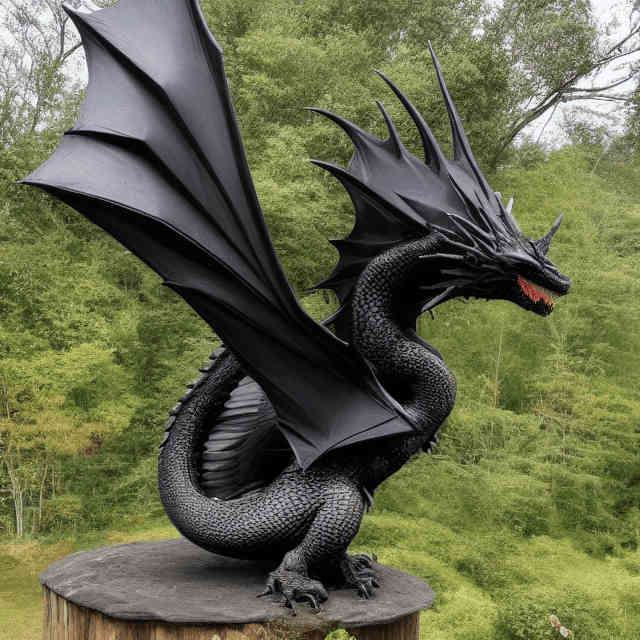}
        \includegraphics[width=0.24\textwidth]{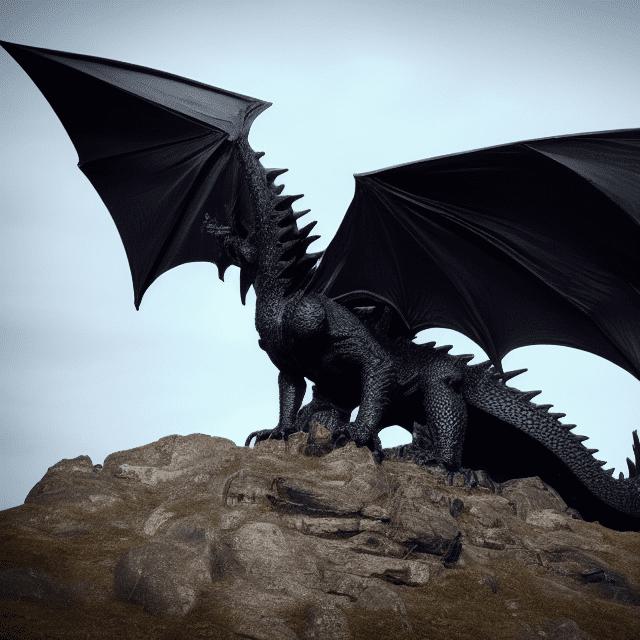}
        \includegraphics[width=0.24\textwidth]{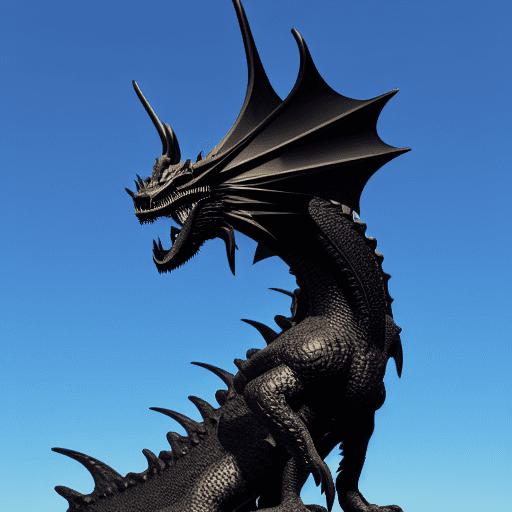}
        \subcaption{mix of \textit{Greg Rutkowski} and \textit{Even Amundsen} models}
        \label{fig33:row_c}
    \end{minipage}
    \caption{Examples of generated outputs: a) \textbf{A+SD}, b) \textbf{B+SD}, c) \textbf{A+B+SD}, and d) \textbf{SD}.}
    \label{fig33:all_images}
\end{figure}

\begin{figure}[h]
    \centering
    \includegraphics[width=0.5\textwidth]{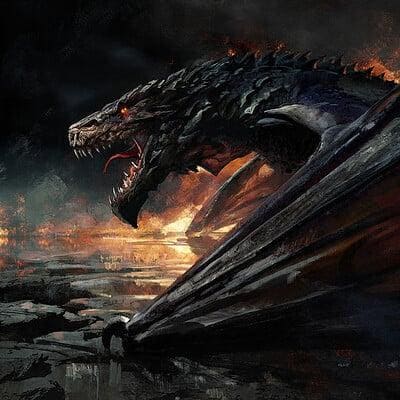}
    \caption{\textit{Greg Rutkowski's} artwork, treated as a generated blend of styles \textbf{(A + B +SD)}, with the objective of assessing each participant's contribution.}
    \label{fig34:second_image}
\end{figure}

And the last experiment (see. Figure~\ref{fig33:all_images}). We want to check whether our method is capable of recognizing the case, when we try to describe a real image instead of generated.
In this experiment, we consider a real image as a generated one, with the aim of determining whether our algorithm can identify real images and reward the artist accordingly. To do this, we substitute the synthetic images generated by the model with a real one (Figure~\ref{fig34:second_image}).

\begin{figure}[h]
    \centering
    \includegraphics[width=0.8\linewidth]{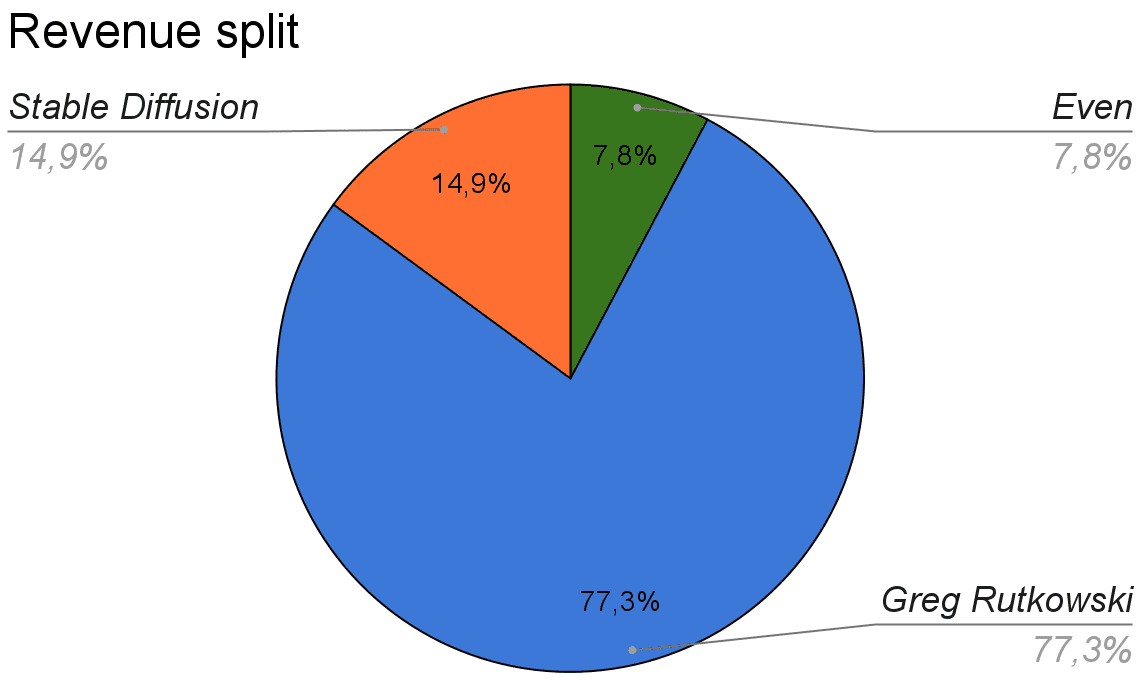}
    \caption{Distribution of revenue for the mix of \textit{Greg Rutkowski} and \textit{Even Amundsen} styles.}
    \label{fig34:first_image}
\end{figure}

This was quite a complex case. However, our algorithm handled it well and transferred the majority of the reward to the artist, the author of the original image (see Figure~\ref{fig34:first_image}). Yes, of course, not 100\% of the reward went to the original author, but this can be attributed to the peculiarities of calculating Shapley values and the fact that such a case is non-standard for our algorithm. We would also like to note that most of the remuneration went to the original work, which is why it was valued significantly more than the second closest image.

In this chapter, we examined scenarios where the generative model amalgamates multiple styles during image creation, significantly complicating the process of calculating the contribution of each participant. Nevertheless, our approach effectively addresses this complexity. Additionally, we tested the resilience of our model to unforeseen scenarios and extended our methodology to other generative models, namely ImageMixer and SDXL. All these findings underscore the robustness and reliability of our method, which is further corroborated by a human evaluation procedure of the obtained results.

\section{Conclusion}
In this study, we have proposed a methodology to allocate rewards for generated art between the StableDIffusion-v1.5 model and the artist(s) whose style was utilized during generation. Our approach is grounded in Shapley Values and embeddings from the CLIP model. We have introduced a method to assess the artist's contribution to the final generation, taking into account whether the model was trained on the works of a specific artist and how accurately it can reproduce their unique style. Additionally, we explore cases where the styles of multiple artists are blended in image generation and propose a method to calculate each artist's contribution.

The practical utility of our solution is multifaceted. It enables anyone who discerns their data being utilized during the model training process, and consequently realizes that the model can emulate the style of their works, to appraise the extent of their contribution to the final generation and demand due recompense from the model provider. Now, such a requirement will be fortified by a mathematical foundation and will be balanced to consider the contributions of all stakeholders, rewarding each in proportion to their input. Furthermore, all computations substantiating this requirement can be executed entirely locally, obviating the need to share private data with external entities.

We hold that this will facilitate a consensus on the use of data obtained through automated web scraping when training generative AI models, thereby averting potential social cataclysms. By adopting such an approach, a harmonious equilibrium between data utilization and fair compensation can be achieved, ensuring the protection of individual rights and fostering responsible advancement in the field of generative AI.

% Start appendices
\appendix
\section{List of prompts}
\label{app:example}

\begin{verbatim}
a castle, by [artist X]
a machine, by [artist X]
science fiction, by [artist X]
a portrait of a character in a scenic environment, by [artist X]
a building in a stunning landscape, by [artist X]
\end{verbatim}

\bibliographystyle{plain}
\bibliography{references}

\end{document}